%% file: main.tex
\PassOptionsToPackage{table}{xcolor}
\documentclass[fleqn,10pt]{wlscirep}
\usepackage[utf8]{inputenc}
\usepackage[T1]{fontenc}
\usepackage{booktabs}
\usepackage{multirow}
\usepackage{tabularx}
\usepackage{makecell}
\usepackage{array}
\usepackage{amsmath}
\usepackage{amssymb}
\usepackage{amsthm}
\usepackage{graphicx}
\usepackage{longtable}
\usepackage{array}
\usepackage{tikz}
\usepackage{caption}
\usepackage{placeins}   
\usepackage{algorithm}
\usepackage{algpseudocode}
\usetikzlibrary{arrows.meta}
\newtheorem{theorem}{Theorem}
\newtheorem{lemma}{Lemma}
\usepackage[table]{xcolor}
\usepackage[normalem]{ulem}
\usepackage{lineno}

\title{Causal graph fusion across incomplete cohorts and causal steering of tabular foundation models}

\author[1,2,*]{Yanke Li}
\author[1]{Dhiyaa Al Jorf}
\author[1,3]{Ayoube Ighissou}
\author[4]{Anke Scheel-Sailer}
\author[1,5,6]{Robert Riener}
\author[1,2,*]{Diego Paez-Granados}

\affil[1]{Spinal Cord Artificial Intelligence (SCAI) Lab, ETH Z\"urich, Zurich, Switzerland}
\affil[2]{Digital Healthcare and Rehabilitation, Swiss Paraplegic Research, Nottwil, Switzerland}
\affil[3]{Department of Physics, Universit\`a degli Studi di Milano-Bicocca, Milan, Italy}
\affil[4]{Swiss Paraplegic Center, Nottwil, Switzerland}
\affil[5]{Balgrist University Hospital, Z\"{u}rich, Switzerland}
\affil[6]{Graduate School of Engineering, Tohoku University, Sendai, Japan}

\affil[*]{Corresponding authors: yanke.li@hest.ethz.ch; diego.paez@hest.ethz.ch}

\begin{abstract}
Clinical data are split across incomplete cohorts with partially overlapping variables. Causal discovery under missing data and multi-source integration have developed separately, and tabular foundation models cannot take a causal graph without retraining. We make two contributions. First, causal graph fusion learns a partial ancestral graph on the shared variables, fuses it into each cohort's graph under temporal background knowledge and lets Markov blankets guide re-imputation, so missing data are handled inside discovery. Second, causal attention steering compiles the learned graph into the attention of a frozen tabular foundation model at inference. Against known ground truth, fusion dominates pooled and per-cohort baselines in every missingness and heterogeneity condition tested ($p<0.001$), and steering improves discrimination under distribution shift ($p<10^{-5}$), a gain that a random graph does not produce. In two spinal cord injury cohorts, fusion orients age, albumin and hemoglobin as ancestors of length of stay, a set no single cohort orients in full, and yields compact, temporally ordered predictor sets that match or exceed automatic feature selection on four of five outcomes.
\end{abstract}

\newif\ifSIinline \SIinlinetrue
\ifSIinline\else \usepackage{xr-hyper}\fi
\begin{document}
\flushbottom
\maketitle
\thispagestyle{empty}

\section*{Introduction}

Clinical knowledge is rarely held in one complete dataset. It accumulates across cohorts collected at different stages of care, each recording an overlapping but non-identical set of variables and each with substantial missingness. Learning how clinical variables depend on one another from such sources is a structure-learning problem that current methods address only in parts, and it bounds what medical AI can learn from real-world data, since causal discovery and tabular foundation models alike start from a single table. Secondary complications after spinal cord injury (SCI) are a representative case. Pressure injuries (PIs) and prolonged hospital stays arise from interacting neurological, nutritional, autonomic and care-process factors \cite{Mervis2023PressureInjuries,Sezer2015,Kirshblum2011,who2013,Gefen2014,Li2024MixedVariable}, and the data needed to study them are spread across registries that follow patients through different phases of care.

Causal graphical models are a natural tool for this problem. A \emph{causal graph} places an arrow from one variable to another when the first is a (possibly indirect) cause of the second, and \emph{causal discovery} infers such graphs from observational data by testing which variables remain associated once others are held fixed \cite{10.3389/fgene.2019.00524,Spirtes2000}. Because some causes are unmeasured, we target a \emph{partial ancestral graph} (PAG), which represents what the data can resolve under latent confounding and selection \cite{colombo2012learning,Zhang2008Orientation}. For each outcome, the \emph{Markov blanket}, the smallest set of variables that renders all others uninformative for predicting it, gives a minimal sufficient predictor set with a causal reading \cite{Tsamardinos2006MaxMin,Pellet2008MB,Aliferis2010LocalCausal,Yu2018MBNoCausalSufficiency}. Such a graph serves explanation and prediction at once. It shows where each predictor sits relative to the outcome, and because causal mechanisms are more stable than associations, it indicates which predictions should hold when the population shifts \cite{Peters2016ICP,Schoelkopf2021CausalRep}.

Two obstacles stand between these tools and fragmented clinical data. First, missing values distort the conditional-independence relations that discovery relies on, so naive deletion or off-the-shelf imputation changes the learned graph \cite{Tu2019MVPC,Gao2022MissDAG,Strobl2018TestWise}. Missing-data-aware methods now exist, including test-wise deletion for FCI-style search, Missing Value PC, local structure learning for incomplete data, MissDAG and optimal-transport approaches \cite{Strobl2018TestWise,Tu2019MVPC,Qiao2024ANMMissing,Gao2022MissDAG,Vo2024OTM}, but each operates on a single dataset, and the iterative coupling of Markov-blanket learning with imputation in MimMB targets one variable at a time \cite{Yu2022MimMB}. Second, integrative methods such as integrative causal analysis, IOD and its extension to tiered background knowledge, Joint Causal Inference and COmbINE show that overlapping sources support stronger structural conclusions than any one alone, including under latent confounding \cite{ion_paper,Tsamardinos2012Integrative,TillmanSpirtes2011IOD,BangDidelez2025tIOD,Mooij2020JCI,Triantafillou2015COmbINE}, but they take each source as already analysable. Two recent methods join the two lines. One pools conditional-independence evidence across multiply imputed sources with non-identical variable sets inside a PC-style search, which assumes no hidden confounders \cite{Woldemariam2026}, and another reconstructs missing variables across time series with partially overlapping variables \cite{Gomez2025CMCTS}. Cross-sectional clinical cohorts need all three ingredients at once: discovery that allows for hidden confounders, background knowledge about when along the care pathway each variable is recorded, and a treatment of missingness within each cohort. To our knowledge, no method yet combines them.

A second opportunity lies in how the learned structure is used. Tabular foundation models such as TabPFN predict accurately from small tables in a single forward pass \cite{Hollmann2025TabPFN}, and their frozen representations already encode recoverable causal structure \cite{Swelam2025TabPFNCausal}. A known causal graph, however, cannot be handed to them at inference, because current approaches that inject graph knowledge learn attention biases through training \cite{Reuter2026PartialGraphs}.

We make two methodological contributions and test them against known ground truth and in clinical data:
\begin{enumerate}
 \item \textbf{Causal graph fusion learns causal structure across incomplete, partially overlapping cohorts.} To our knowledge it is the first method to handle hidden confounders, temporal background knowledge and within-cohort missingness together. A PAG learned on the variables shared across cohorts is fused into each cohort's own PAG under temporal-tier background knowledge \cite{Meek1995CausalKnowledge}, and each variable's Markov blanket guides its re-imputation \cite{Liu2023MBMF,Yu2022MimMB}, so missingness is handled inside discovery instead of before it. For an idealized form of the fusion operator we prove consistency and finite-sample robustness, and give sufficient conditions for the imputation--discovery loop to stabilize.
 \item \textbf{Causal attention steering gives a frozen tabular foundation model a causal graph at inference, without training.} Steering compiles the learned PAG into a bias on the self-attention of the frozen model, instantiated on TabPFN\,v2~\cite{Hollmann2025TabPFN}, and a simple rule sets its target: the causal ancestors under mechanism shift and the Markov blanket under covariate shift.
 \item \textbf{Both methods are validated against known ground truth.} On synthetic benchmarks, fusion dominates pooled and per-cohort baselines on the precision/recall frontier in every missingness and heterogeneity condition tested ($p<0.001$), raises graph F1 from about $0.63$ to $0.71$, and imputes with a third less distributional distortion (KS distance $0.034$ versus $0.051$). Under distribution shift, with the steering strength tuned without any test-fold information, steering significantly improves discrimination ($+0.011$ AUROC, $p<10^{-5}$), rising to $+0.029$ when the validation shift matches the test shift. The gain is specific to the causal graph. A random graph yields none, and the gain grows as more of the true graph is revealed to the steerer. We also show why fusion imputes before testing. Conditional-independence tests that handle missing values natively reach at most 30\% power on discrete benchmark networks even without missingness, and on the clinical cohorts they leave length of stay nearly or entirely disconnected.
 \item \textbf{In spinal cord injury, fusion recovers relations that no single cohort resolves and returns compact, auditable predictor sets.} On two Swiss Paraplegic Center cohorts, hospital-acquired pressure injury (HAPI, $n=250$) and post-pressure-injury surgery (PPIS, 266 surgical episodes from 174 patients), the pooled graph orients age, albumin and hemoglobin as ancestors of length of stay, a set that each cohort alone orients only in part, and fusion places all three in each cohort's Markov blanket. The fused blankets match or exceed automatic feature selection on four of five outcomes while every predictor keeps its place on the care timeline and every cross-cohort edge carries bootstrap support. On secondary pressure-injury complication the blanket is the best feature set of all (AUROC $0.776$, versus $0.752$ with all 75 candidate variables), and on HAPI length of stay it is the simplest model within one standard error of the best ($R^2=0.75$ with 21 of 43 variables). Re-learning the graph inside every training fold keeps the blanket competitive, first of all feature sets on PPIS length of stay.
\end{enumerate}
These results make fragmentation across cohorts a source of causal evidence.

\section*{Results}

\subsection*{Markov-blanket graph learning through multi-cohort fusion}
\begin{figure}[ht]
\centering
\includegraphics[width=\linewidth]{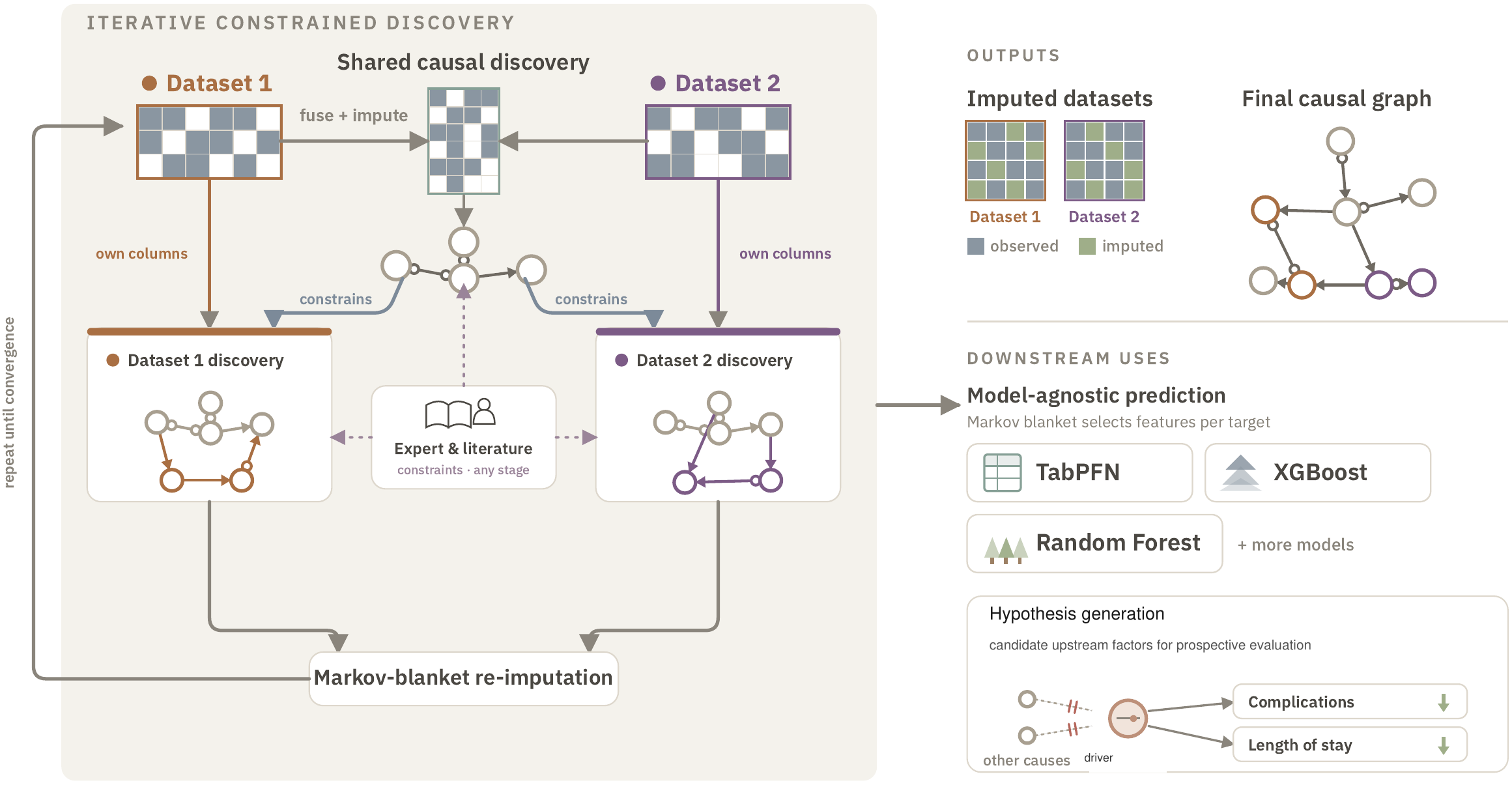}
\caption{\textbf{Causal graph fusion: from incomplete, partially overlapping cohorts to one causal graph and its downstream uses.} Individual incomplete datasets are aligned and pooled on shared variables. A pooled-shared partial ancestral graph yields cross-cohort causal evidence, which, together with temporal-tier expert knowledge, constrains per-cohort PAG learning, with Markov-blanket re-imputation iterated until convergence. The outputs (imputed per-cohort datasets and a final causal graph) feed downstream uses: model-agnostic prediction, where the target's Markov blanket selects features for predictors such as TabPFN, XGBoost, and random forests, and hypothesis generation for intervention planning over the learned graph.}
\label{fig:framework}
\end{figure}

The proposed workflow is shown in Figure~\ref{fig:framework} and summarized in Algorithm~\ref{alg:fusion}. After variable alignment across cohorts, missing entries are initialized with the median for continuous variables and the mode for categorical and ordinal-encoded variables. The pipeline then alternates between two ingredients. First, a pooled-shared PAG is estimated on the union of cohorts restricted to variables that appear in every cohort, optionally constrained by temporal-tier background knowledge \cite{Meek1995CausalKnowledge}. This small subgraph captures the cross-cohort causal evidence that survives pooling. Second, a per-cohort PAG is learned on each cohort using the union of (i) temporal-tier constraints and (ii) constraints derived from the pooled-shared subgraph, so that edges or orientations supported by cross-cohort evidence are required, and edges that are absent in the pooled-shared subgraph are forbidden. Iterative Markov-blanket-guided re-imputation is then applied within each cohort. The final per-cohort PAG, augmented by edges from the pooled-shared subgraph that involve shared variables, is used both for clinical interpretation and for target-specific Markov-blanket feature selection. Although the present implementation uses FCI, the framework is compatible with any causal discovery algorithm that supports enforcing or forbidding specific edges as background knowledge.

\subsection*{Fusion improves structural recovery and distributional fidelity}
\begin{figure}[h!]
\centering
\includegraphics[width=\linewidth]{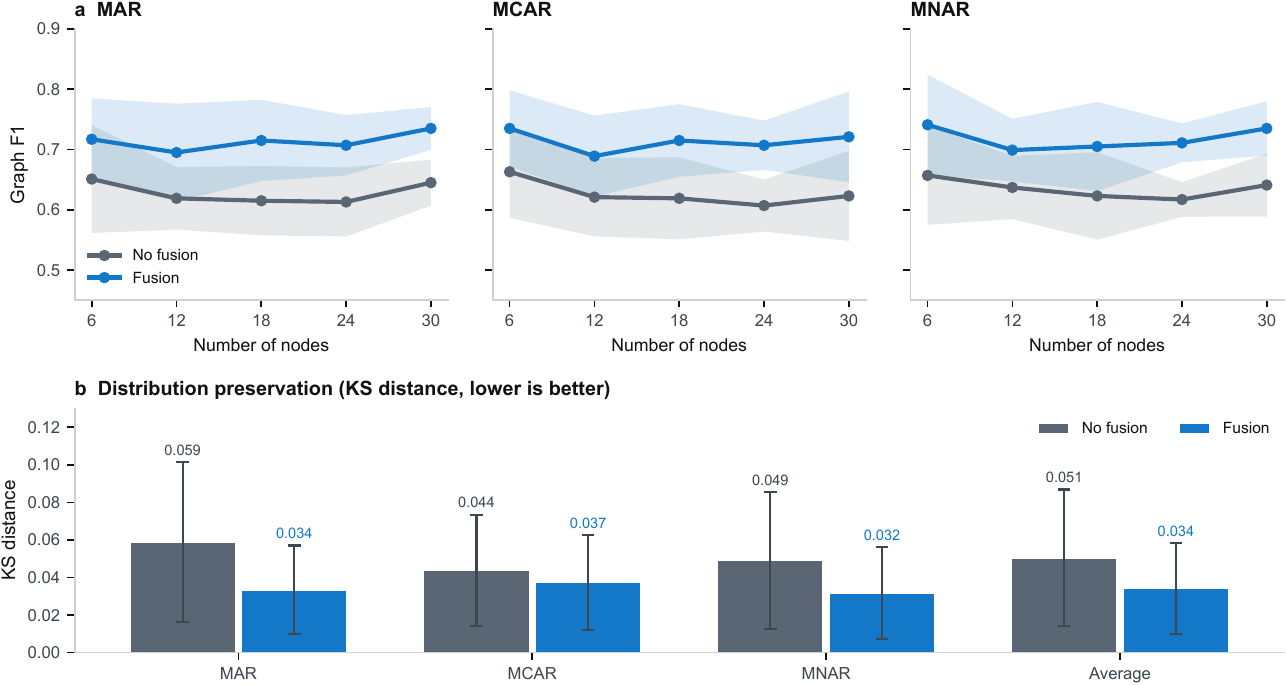}
\caption{\textbf{Fusion recovers more accurate graphs and better-preserved distributions than discovery without fusion, under every missingness mechanism.} \textbf{(a)}~Structural recovery: graph F1 of the learned structure against ground truth as a function of graph size, under MAR, MCAR, and MNAR (blue: fusion; grey: non-fusion; lines are means over sample sizes 200--1000, bands $\pm 1$ standard deviation). \textbf{(b)}~Distribution preservation: KS distance between original and imputed values (lower is better) by missingness mechanism and their average; bars are means with $\pm 1$ standard deviation across graph sizes. The fusion framework achieves higher F1 and lower KS across all mechanisms.}
\label{fig:synth}
\end{figure}

On synthetic data, the fusion-based approach consistently outperformed the non-fusion baseline in terms of structural recovery. Figure~\ref{fig:synth}a summarizes graph-F1 across graphs with 6--30 variables and sample sizes from 200 to 1000, under all three missingness mechanisms. The fused approach achieved higher F1 (around $0.71$ versus $0.63$ for the non-fusion baseline) across graph sizes and mechanisms, indicating that cross-dataset fusion is most valuable when the available information is limited or fragmented, and least so when it is abundant.

Distribution preservation was assessed using the Kolmogorov--Smirnov (KS) distance between the original and imputed values. Across missingness mechanisms the fusion approach maintained lower KS distances than the non-fusion baseline (Figure~\ref{fig:synth}b), indicating that borrowing information across overlapping datasets helped maintain the shape of the original empirical distribution in addition to improving pointwise accuracy.

A similar pattern was observed for raw imputation accuracy. In the reported experiments, both methods degraded with increasing missingness, but the fused approach remained consistently more accurate. The effect was again most pronounced in the high-missingness regime, where the gap between the two methods widened across MAR, MCAR, and MNAR settings (Figure~\ref{fig:rmse_panel}). The synthetic results therefore indicate that graph-aware multi-dataset fusion improves both value-level recovery and distributional fidelity under challenging missing-data conditions.

\subsection*{Why not skip imputation?}

We evaluate three native-NaN CI tests as drop-in replacements for Fisher-Z inside FCI: the Generalized Covariance Measure (GCM)~\cite{Shah2020GCM,Hollmann2025TabPFN}, CRT-TabPFN~\cite{Salem2026CRTTabPFN,Candes2018Panning,Berrett2020CRT}, and Missing-Values Fisher-Z (MV-FisherZ)~\cite{Zheng2024CausalLearn}.

All three tests are limited by effective-sample collapse.
Any CI test restricted to rows where all query variables $(X, Y, \mathbf{Z})$ are
simultaneously observed retains at most
\[
n_\mathrm{eff} = n\cdot(1-\pi)^{2+d_Z}
\]
rows under MCAR at rate $\pi$, where $d_Z = |\mathbf{Z}|$. At $n=250$, $\pi=0.30$,
$d_Z=3$: $n_\mathrm{eff}\approx 42$; at $d_Z=5$: $n_\mathrm{eff}\approx 21$.
This reduction applies directly to GCM (which discards rows where any conditioning
variable is missing) and MV-FisherZ (which uses test-wise deletion). At these
effective sample sizes, clinical data with mixed discrete-continuous variables
carries insufficient signal-to-noise to reliably distinguish conditional dependence
from independence. Under MNAR, the surviving complete cases additionally form a
biased subsample, distorting the estimated conditional dependencies in ways that likelihood methods cannot correct.

Two of the methods have additional failure modes.
For CRT-TabPFN under MCAR, the null draws
$Y^{(b)}\sim\hat{P}(Y\mid\mathbf{Z})$ are fully observed while the
observed $\widetilde{Y}$ contains missing entries. Because TabPFN's attention
mechanism treats missing and observed tokens differently, $T_\mathrm{obs}$ and
$T^{(b)}$ follow different distributions even under $H_0$, invalidating the
exchangeability on which the CRT's Type-I guarantee rests. Empirically, this
inflates Type-I error to 80\% on the linear null at any MCAR rate
(Table~\ref{tab:ci_test_summary}). MV-FisherZ fails for a reason unrelated to missingness. It assumes multivariate Gaussianity, so on nonlinear
data the linear partial correlation is nonzero under $H_0$ even on complete data,
producing 100\% Type-I error on the nonlinear null at every MCAR level (90\% under MNAR).

We evaluate all three methods on synthetic linear and nonlinear scenarios and
on real Bayesian networks from pgmpy (survey 6-node, cancer 5-node, child 20-node;
$n=250$, $B=99$, $\alpha=0.05$). Table~\ref{tab:ci_test_summary} summarizes the results. The primary finding is that all three methods achieve at most 30\% power on the real Bayesian networks, even without any missingness. Power failure, rather than Type-I miscalibration, is the dominant obstacle to using these tests inside FCI on clinical data.

\begin{table}[ht]
\centering
\caption{Rejection rates for the three CI tests across data types and missingness conditions ($n=250$, $B=99$, $\alpha=0.05$; values read from Figure~\ref{fig:negative_finding} in Supplementary). Bold: outside target range (Type-I $>15\%$; power $\le70\%$).}
\label{tab:ci_test_summary}
\small
\setlength{\tabcolsep}{4.5pt}
\begin{tabular}{llrrrrr}
\toprule
& & \multicolumn{5}{c}{Missingness level} \\
\cmidrule(l){3-7}
Data type & Method & None & MCAR 10\% & MCAR 20\% & MCAR 30\% & MNAR \\
\midrule
\multicolumn{7}{l}{\textit{$H_0$: Type-I error (\%; target $\approx$5\%)}} \\
\midrule
\multirow{3}{*}{Syn.\ linear}
  & GCM        & 2   & 0   & 10  & 0   & 0  \\
  & CRT-TabPFN & \textbf{80} & \textbf{80} & \textbf{80} & \textbf{80} & \textbf{50} \\
  & MV-FisherZ & 0   & 0   & 0   & 0   & 0  \\
\midrule
\multirow{3}{*}{Syn.\ nonlinear}
  & GCM        & 10  & 0   & 0   & 0   & \textbf{30} \\
  & CRT-TabPFN & \textbf{60} & \textbf{30} & \textbf{30} & \textbf{40} & \textbf{80} \\
  & MV-FisherZ & \textbf{100} & \textbf{100} & \textbf{100} & \textbf{100} & \textbf{90} \\
\midrule
\multirow{3}{*}{Real BNs}
  & GCM        & 15  & 10  & \textbf{20}  & 15  & 15 \\
  & CRT-TabPFN & 10  & 10  & 15  & 10  & 0  \\
  & MV-FisherZ & 15  & 10  & \textbf{20}  & 5   & 0  \\
\midrule
\multicolumn{7}{l}{\textit{$H_1$: Power (\%; target $>$70\%)}} \\
\midrule
\multirow{3}{*}{Syn.\ linear}
  & GCM        & 100 & 100 & 100 & 100 & 100 \\
  & CRT-TabPFN & 100 & 80  & 90  & 100 & \textbf{60} \\
  & MV-FisherZ & 100 & 100 & 100 & 100 & 100 \\
\midrule
\multirow{3}{*}{Syn.\ nonlinear}
  & GCM        & 100 & 100 & 100 & 90  & 100 \\
  & CRT-TabPFN & 100 & 100 & 100 & 90  & \textbf{70} \\
  & MV-FisherZ & 100 & 100 & 100 & 90  & 90  \\
\midrule
\multirow{3}{*}{Real BNs}
  & GCM        & \textbf{20} & \textbf{20} & \textbf{20} & \textbf{25} & \textbf{15} \\
  & CRT-TabPFN & \textbf{10} & \textbf{15} & \textbf{10} & \textbf{10} & \textbf{5}  \\
  & MV-FisherZ & \textbf{30} & \textbf{20} & \textbf{20} & \textbf{15} & \textbf{15} \\
\bottomrule
\end{tabular}
\end{table}

On clean continuous data such as the Sachs protein-signalling benchmark~\cite{Sachs2005} ($n=853$, 11 nodes, known ground-truth DAG), the same methods remain partially functional. The Erk--Akt edge survives in all methods at all MCAR levels, confirming that the loss of power on clinical data is driven by discrete variable types and plausibly non-random missingness rather than algorithmic limitations of FCI (full Sachs analysis in Supplementary).

\paragraph{HAPI and PPIS: structural collapse under real clinical conditions.}

We then applied GCM-TabPFN and CRT-TabPFN as CI oracles inside FCI to six real
SCI clinical cohorts from the Swiss Paraplegic Center: three HAPI cohorts (first
rehabilitation admission, non-first admission, and no-HAPI comparison; combined
$n=250$) and the PPIS cohort split into three admission-period strata for this experiment (2015--2018, 2019, 2020$+$; combined $n=266$).
These cohorts differ from Sachs in three ways. First, the missingness
mechanism is plausibly non-random (MNAR-like) at approximately 15--22\%. Values
appear more likely to be missing at extreme magnitudes, which would violate the MCAR
assumption that kept the Sachs experiments tractable, although MNAR cannot be proven from the
observed data alone. Second, the variables are discrete and mixed-type (ordinal, categorical, and binary) for which TabPFN residuals carry substantially lower
signal-to-noise than for continuous targets. Third, individual cohorts have
$n=43$--$132$ with up to 25 variables, creating high-dimensional regimes relative to
the sample size.

\begin{figure}[ht!]
\centering
\includegraphics[width=\linewidth]{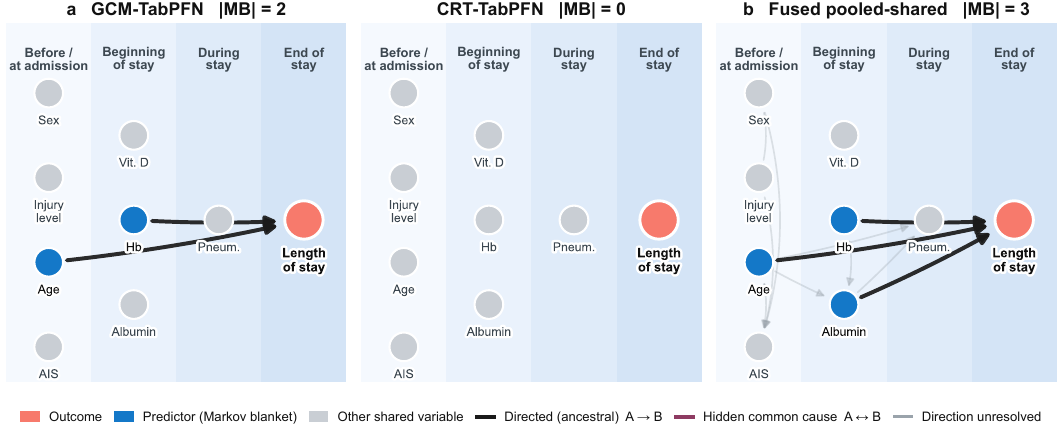}
\caption{\textbf{Imputation is a prerequisite for cross-cohort causal discovery: raw-data conditional-independence tests versus the fused pooled-shared graph for length of stay} ($n=516$; the union of HAPI and PPIS restricted to variables observed in both cohorts). All three sub-panels share one layout. Each variable is placed in the tier in which it is measured (admission $\rightarrow$ discharge), and node positions are identical throughout, so the panels can be read against one another. The outcome (coral) is length of stay, blue circles are its recovered Markov-blanket members, and grey circles are the other shared variables. \textbf{(a)}~Two sub-panels on raw (unimputed) data, GCM-TabPFN (left) and CRT-TabPFN (right). Native-NaN conditional-independence tests recover almost nothing. With GCM-TabPFN as the FCI oracle only two variables attach to length of stay (Age, Hemoglobin), and with CRT-TabPFN it is isolated entirely ($\lvert\mathrm{MB}\rvert=0$). Only edges incident to length of stay are drawn, and the GCM graph contains 3--6 edges in total. Per-cohort results show the same pattern across all six HAPI and PPIS cohorts. \textbf{(b)}~After imputation and fusion, the pooled-shared analysis orients three definite directed edges into length of stay, from Age, Albumin, and Hemoglobin, under the temporal-tier constraint (an admission-time variable cannot be caused by a discharge outcome). Each cohort alone orients only part of this set (Supplementary Table~\ref{tab:edge_stability}). Edge meanings (directed ancestral relation under PAG assumptions, shared hidden common cause, unresolved association) follow the plain-language legend, as in Figure~\ref{fig:mbgraph_comp}.}
\label{fig:pag_pooled}
\label{fig:ci_pooled}
\end{figure}

The results represent a qualitative break from the Sachs findings
(Figure~\ref{fig:ci_pooled}a). In the pooled-shared
analysis ($n\approx516$, 9 shared variables including Length of Stay), GCM finds only
3--6 edges; Length of Stay, the primary outcome, is barely connected to any other
variable. CRT isolates Length of Stay completely. Per-cohort analyses show the same
pattern. GCM finds 11--21 edges across 22--25 variables per cohort, but Length of Stay
remains weakly connected, and CRT isolates it in five of six cohorts. On clinical data at
comparable or lower missingness rates than Sachs, the same methods produce graphs in
which the primary outcome is functionally disconnected. The failure lies in the CI tests' inability to distinguish dependence
from independence under likely non-random missingness, discrete variables, and small cohort sizes, rather than in FCI's depth or orientation rules.
Any Markov blanket estimated from these near-empty graphs is too sparse to support
meaningful variable selection or prediction, confirming that imputation is a necessary
prerequisite for recovering clinical causal structure at the cohort sizes encountered
here.

\subsection*{Cross-cohort causal evidence: a pooled-shared partial ancestral graph}

We next applied the framework to two real-world SCI datasets from the Swiss Paraplegic Center: HAPI ($n=250$) and PPIS ($n=266$). Because no ground-truth causal graph is available in this setting, the evaluation focused on (i) the structural content of the pooled-shared subgraph (which can be interpreted as cross-cohort causal evidence), (ii) the evolution of the per-cohort graphs as expert and cross-cohort information is added, and (iii) the predictive utility of the resulting target-specific Markov blankets.

The pooled-shared subgraph was estimated on $n=516$ patients pooled from both cohorts, restricted to a backbone of eight shared predictors observed in every cohort (AIS, level of injury, age, sex, albumin, hemoglobin, vitamin D, pneumonia) together with the target length of stay (LoS), and constrained by temporal-tier knowledge \cite{Meek1995CausalKnowledge}. SCIM is harmonizable across cohorts but is recorded at discharge in PPIS, so the temporal-tier constraint excludes it from the LoS backbone (Supplementary Table~\ref{tab:shared_backbone}). In the pooled-shared analysis, three definite directed edges into LoS were oriented under the temporal-tier constraint: Age $\rightarrow$ LoS, Albumin $\rightarrow$ LoS, and Hemoglobin $\rightarrow$ LoS (Figure~\ref{fig:pag_pooled}b). Pooling brings the three together. On the shared-variable backbone, run with the same tier knowledge and threshold, each cohort alone orients only part of the set (HAPI the Age and Hemoglobin relations, PPIS the Albumin relation), whereas the pooled analysis orients all three in a single graph. In each cohort's full graph the three relations are screened off almost entirely (bootstrap inclusion $0.00$ to $0.17$), so fusing the pooled edges into the per-cohort graphs places them in the per-cohort Markov blankets (Supplementary Tables~\ref{tab:edge_stability} and~\ref{tab:constraint_ablation}). These three edges form the pooled-shared Markov blanket of LoS and act as a meta-prior on subsequent per-cohort searches.

\subsection*{Information-fusion trajectory: graphs and prediction co-evolve}

\begin{figure}[ht]
\centering
\includegraphics[width=\linewidth]{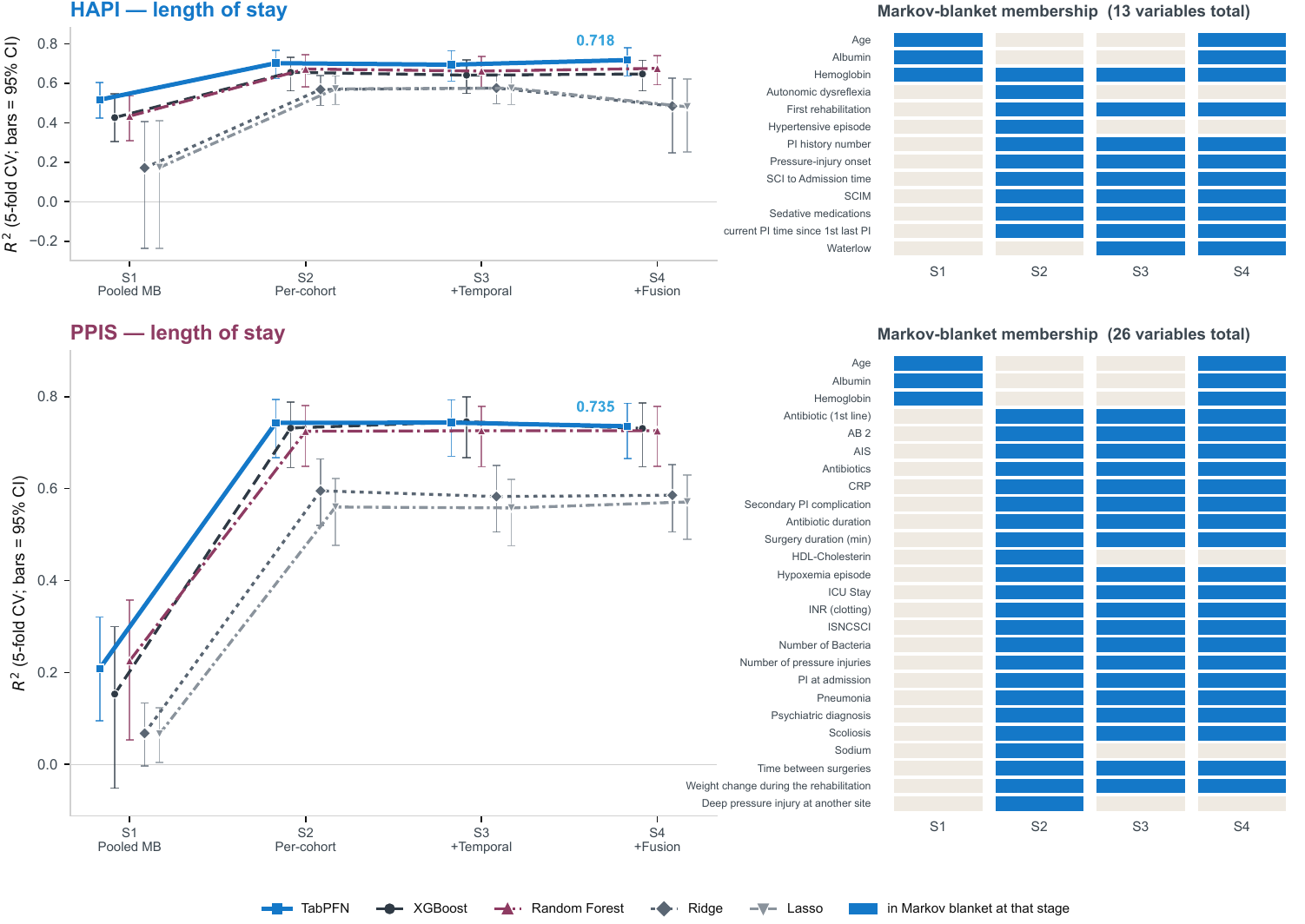}
\caption{\textbf{Information-fusion trajectory: each fusion stage builds the Markov blanket, and the predictive gain holds for every model family.} Rows: HAPI (top) and PPIS (bottom). \emph{Left}: length-of-stay prediction quality (5-fold cross-validated $R^2$) at each of the four fusion stages for five model families identified in the legend. Capped bars are 95\% bootstrap confidence intervals, with points dodged along the stage axis so overlapping intervals stay legible. \emph{Right}: the corresponding Markov-blanket composition, one row per variable and one column per stage; a filled cell marks membership at that stage, so growth and turnover are visible. Stage~1 uses only the three pooled-shared directed-ancestor variables of LoS (Age, Albumin, Hemoglobin); Stages~2--4 progressively add per-cohort discovery, temporal-tier constraints, and the pooled-shared constraints that complete the fusion at the deployable $\alpha=0.05$. The single largest predictive gain in both cohorts comes from Stage~1$\rightarrow$Stage~2 and holds for every model family ($+0.19$ to $+0.40$ $R^2$ on HAPI; $+0.49$ to $+0.58$ on PPIS), so the structural gain is not an artifact of any single learner. The compact cross-cohort claim becomes predictive only once embedded in a richer per-cohort blanket. At Stage~4 TabPFN leads on both cohorts ($R^2=0.718$ HAPI, $0.735$ PPIS; values printed in the panels). These per-stage blankets are the trajectory's own progression and differ from the deployed $\alpha=0.05$ benchmark blanket (HAPI 21, PPIS 34). Relaxing the threshold to $\alpha=0.10$ is model-dependent, and deployable attention steering adds no material gain on these tier-saturated final blankets; both are reported separately (Supplementary Tables~\ref{tab:alpha_trajectory},~\ref{tab:clinical_mode_grid},~\ref{tab:nested_gamma}).}
\label{fig:trajectory}
\end{figure}

To relate progressively richer information fusion to predictive performance, we ran the per-cohort pipeline in four stages, up to the deployable operating point, and at each stage evaluated length-of-stay prediction with five model families spanning distinct inductive biases: a tabular foundation model (TabPFN), boosted trees (XGBoost), bagged trees (random forest), and two penalized linear models, ridge and lasso (Figure~\ref{fig:trajectory}). Attention steering adds no material gain on these tier-saturated final blankets and is reported separately (Supplementary Tables~\ref{tab:clinical_mode_grid} and~\ref{tab:nested_gamma}). Stage~1 uses only the three pooled-shared directed-ancestor variables of LoS. Stage~2 adds per-cohort FCI with no constraints. Stage~3 adds temporal-tier constraints. Stage~4 adds the pooled-shared cross-cohort edges (full fusion, $\alpha=0.05$). Relaxing the threshold further to $\alpha=0.10$ helps the flexible learners but collapses the penalized linear models, so we report that additional stage separately in Supplementary Table~\ref{tab:alpha_trajectory}.

First, the single largest predictive gain in both cohorts comes from Stage~1$\rightarrow$Stage~2 and is model-agnostic. Adding cohort-specific predictors on top of the three-variable cross-cohort claim lifts every one of the five model families ($+0.49$ to $+0.58$ $R^2$ on PPIS, $+0.19$ to $+0.40$ on HAPI). The pooled-shared subgraph alone is a compact causal summary that is not competitive as a predictor set. Second, expert and cross-cohort constraints do not improve $R^2$ uniformly. Stage~3 (temporal-tier) prunes the PPIS Markov blanket from 23 to 20 features while preserving performance, and Stage~4 (pooled-shared constraints) restores the three shared directed-ancestor variables ($20\rightarrow23$), completing the fusion at the deployable $\alpha=0.05$. Relaxing the threshold beyond this point to $\alpha=0.10$ admits 12 further PPIS features ($23\rightarrow35$) and has a model-dependent effect, helping the flexible learners but collapsing the penalized linear models; we therefore report it separately (Supplementary Table~\ref{tab:alpha_trajectory}). Third, the ordering of the model families is stable across stages. TabPFN leads at every stage in both cohorts, the tree ensembles track it closely on PPIS after Stage~2, and the linear models trail throughout. The same trajectory shape for a tabular foundation model, boosted trees, bagged trees, and penalized linear regression confirms that the structural gain from fusion comes from the recovered graph and holds across learners. Deployable attention steering adds no material gain on these blankets. Steering within the tier-saturated final FCI Markov blanket at a fixed $\gamma=2.0$ leaves $R^2$ essentially unchanged (HAPI $0.745\rightarrow0.751$, PPIS $0.710\rightarrow0.710$), and leak-free per-fold $\gamma$ selection, which steers the full leakage-safe pool toward the blanket, likewise stays within noise of the baseline (Supplementary Table~\ref{tab:nested_gamma}).

\subsection*{From causal graph to attention mask: the steering mechanism}

\begin{figure}[ht]
\centering
\includegraphics[width=\linewidth]{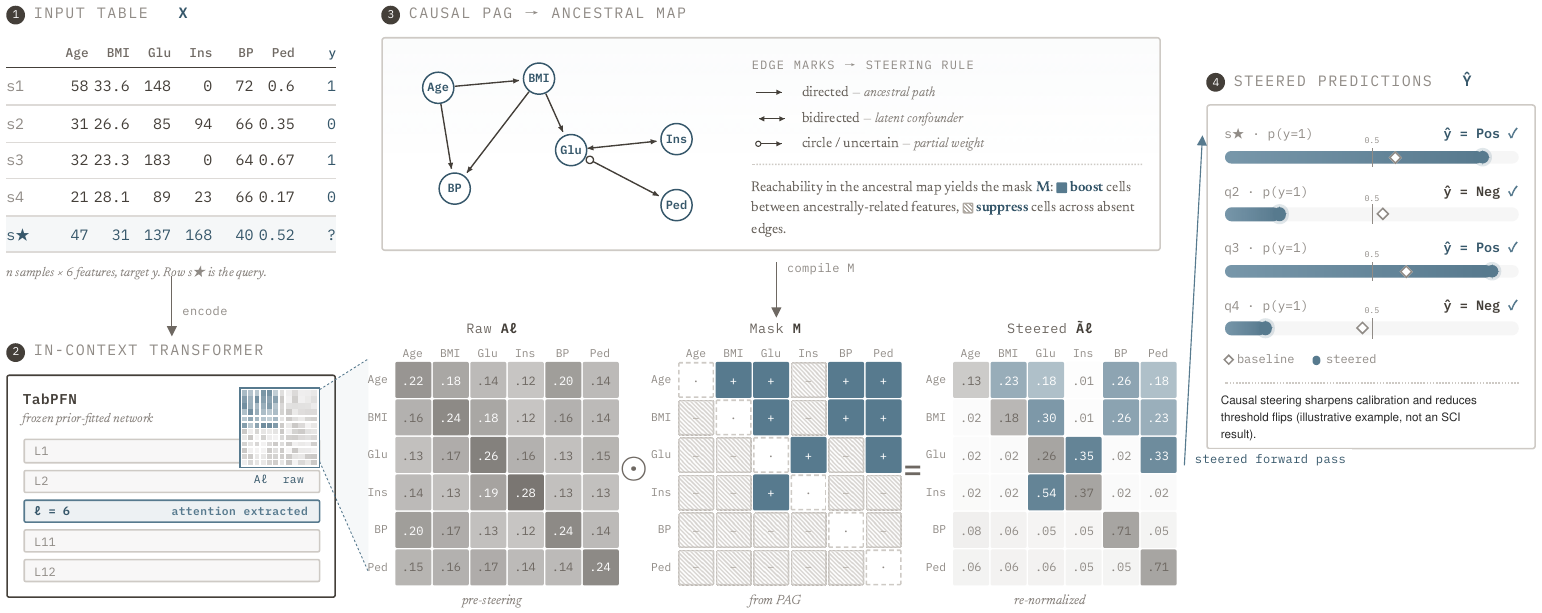}
\caption{\textbf{PAG-guided attention steering of TabPFN (schematic; illustrative feature set).} \textbf{(1)}~An input table is processed in context by \textbf{(2)}~a frozen, prior-fitted TabPFN network, from which the self-attention logits $S_\ell$ are read at an intermediate layer. \textbf{(3)}~The learned partial ancestral graph is compiled into a mask $M$. Reachability in its ancestral map marks attention entries to boost (between ancestrally related features) or suppress (across pairs with no causal path), and the directed, bidirected and circle edge marks map to the corresponding rule. The mask is added to the pre-softmax attention logits, $\tilde{A}_\ell = \mathrm{softmax}(S_\ell + \gamma M)$, and \textbf{(4)}~the steered forward pass re-weights which feature tokens the prediction reads from. The six-feature example (Age, BMI, Glucose, Insulin, blood pressure, pedigree) illustrates the mechanism generically and is not a spinal-cord-injury result. The quantitative clinical effects of steering are reported in Figures~\ref{fig:los_results_hapi}--\ref{fig:ppis_classification} and Supplementary Tables~\ref{tab:clinical_mode_grid} and~\ref{tab:nested_gamma}. Any pretrained model that exposes self-attention over feature and target tokens can be steered the same way; TabPFN\,v2 is the instance we evaluate here.}
\label{fig:steering_schematic}
\end{figure}

\begin{figure}[ht]
\centering
\includegraphics[width=\linewidth]{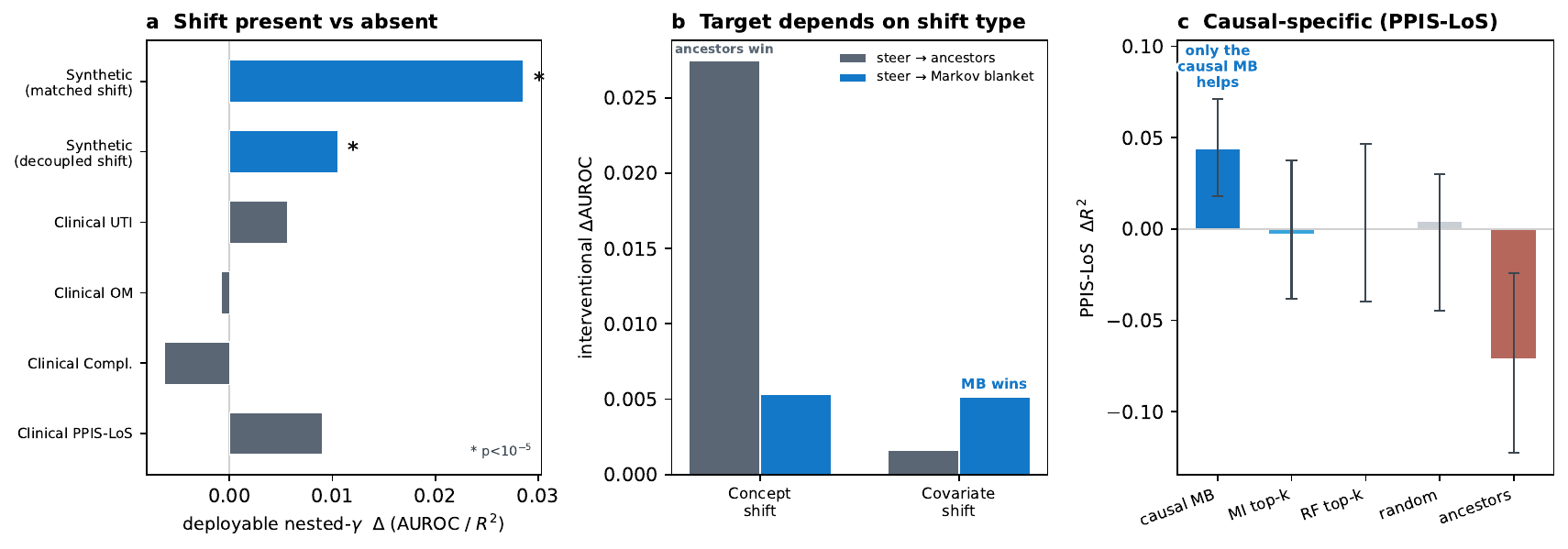}
\caption{\textbf{Attention steering as a deployable, shift-matched robustness tool.} \textbf{(a)}~Deployable per-fold ($\gamma$-selected) $y\!\to\!x$ steering significantly improves interventional accuracy on synthetic ground truth, where a distribution shift is present (144 structural causal models), but stays within noise on the observational clinical cohorts, where no explicit shift is available against which to select $\gamma$. The measured shift is covariate-only (domain-classifier AUROC $\approx0.68$ with stable $P(y\mid x)$). \textbf{(b)}~On synthetic ground truth, the better causal target depends on the shift type, with causal ancestors winning under concept (mechanism) shift and the Markov blanket under covariate shift. \textbf{(c)}~On PPIS length of stay the benefit is causally specific. Steering the full feature pool toward the causal Markov blanket helps, whereas equally-sized mutual-information, random-forest, or random subsets do not, and ancestor-steering is harmful (the covariate-shift regime, where the Markov blanket is the correct target). Panels correspond to the analyses in Supplementary Tables~\ref{tab:steer_significance} and~\ref{tab:clinical_mode_grid}.}
\label{fig:steering_synthesis}
\end{figure}

Attention steering turns the learned causal graph into a lightweight, inference-time bias for any attention-based tabular foundation model that ingests the feature columns and target as tokens, so that causal knowledge biases the model toward more distribution-shift-robust predictions without retraining (Figure~\ref{fig:steering_schematic}). We instantiate it on TabPFN\,v2~\cite{Hollmann2025TabPFN}, which processes an entire table in context. Each feature column and the target become tokens, and self-attention layers decide how information flows between them. Before a forward pass we add a scalar bias $\gamma$ to selected entries of the attention logits, nudging the model to attend more to causally relevant relationships and less to spurious ones. The bias is compiled from the constrained PAG and its temporal-tier ordering. Reachability in the graph's ancestral map determines which attention entries are boosted (between ancestrally related features, or from the target token to its ancestors) and which are suppressed (across pairs with no causal path). The effective mode is $y\!\to\!x$ (target-to-feature). It boosts the target token's attention toward a causal target set, the causal ancestors (\emph{Target-Boost}) or the full Markov blanket (\emph{MB-Boost}), and so gates which features the final prediction reads from. We find the alternative feature--feature ($x\!\to\!x$) mode, which biases attention between feature tokens, to be near-inert for prediction once verified against TabPFN's read-out, and do not use it as a predictive knob. The better $y\!\to\!x$ target depends on the distribution-shift type, with causal ancestors under mechanism (concept) shift and the full Markov blanket under covariate shift (Figure~\ref{fig:steering_synthesis}). A single strength $\gamma=2.0$, tuned only on independent synthetic and Sachs data, is used throughout the clinical evaluation. Our steering is training-free and driven by a fixed, externally estimated PAG. It complements concurrent work that injects partial causal-graph knowledge into tabular foundation models through learnable attention biases~\cite{Reuter2026PartialGraphs}, and is motivated by evidence that TabPFN's frozen representations already encode recoverable causal structure~\cite{Swelam2025TabPFNCausal}. Because the bias acts only on attention logits, the approach applies to any attention-based tabular foundation model that tokenizes features and target. We validate it on TabPFN\,v2 and examine in the Supplement the engineering step of adapting the injection to architecturally distinct successors such as TabPFN\,v2.5, whose additional between-items ``thinking-token'' pathway carries information that the feature and target attention rows do not.

\subsection*{Downstream prediction with PAG-guided attention steering: length of stay}

\begin{figure}[ht!]
\centering
\includegraphics[width=\linewidth]{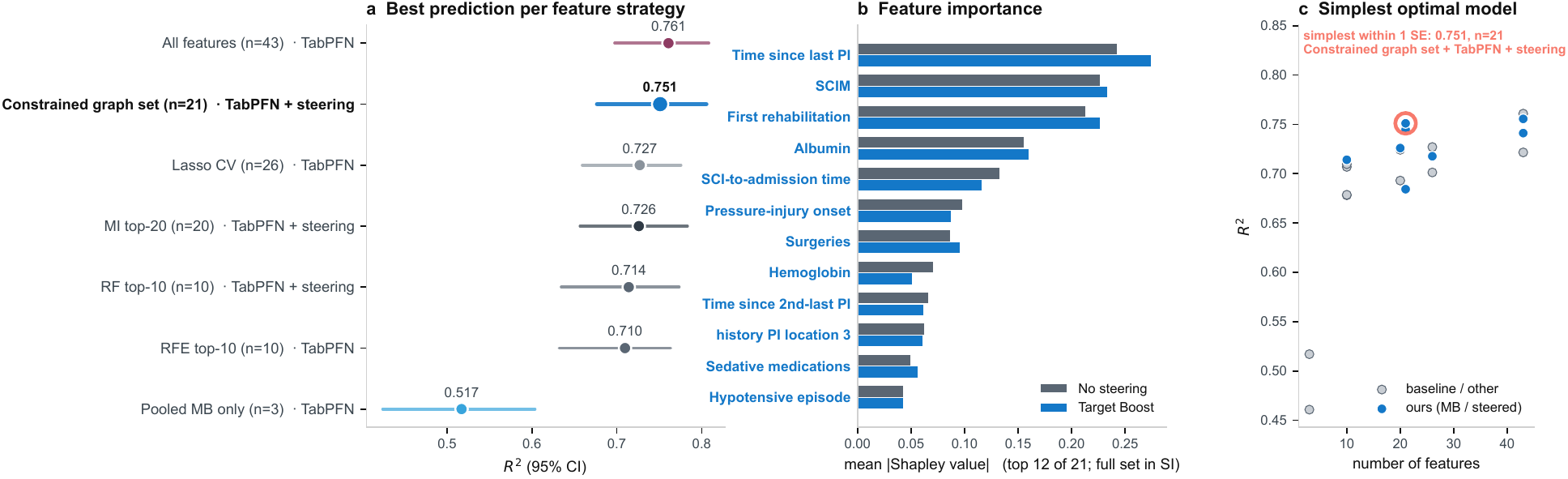}
\caption{\textbf{HAPI length of stay ($n=250$): the 21-variable constrained graph set scores higher than every automatic feature selector and is the simplest model within one standard error of the best.} Higher $R^2$ is better. \textbf{(a)}~Best out-of-fold performance achievable with each feature-selection strategy, our constrained FCI Markov blanket against the automatic baselines. Each row names the feature count $n$ and the model that achieved it, since a best-of-$N$ sweep is run per strategy and the winner is not always the same learner. Dots are point estimates, whiskers 95\% bootstrap confidence intervals (300 resamples); the causal Markov blanket is shown in bold. \textbf{(b)}~Marginal Shapley importance (mean $|\mathrm{SV}|$ under mean-masking) for the constrained FCI Markov blanket, without steering and with Target-Boost ($\gamma=2.0$); temporal causal ancestors of the target are in blue, and only the twelve strongest features are shown, with the full membership in Supplementary Table~\ref{tab:mb_features}. \textbf{(c)}~Out-of-fold $R^2$ against number of features for every evaluated method (blue: our Markov-blanket / steered variants; grey: baselines). The ring marks the \emph{simplest-optimal} model, the fewest-feature method whose performance is within one standard error of the best (one-SE rule), matching the ``simplest within 1 SE'' label printed in the panel. The SE is recovered from the 95\% bootstrap interval as $(\mathrm{hi}-\mathrm{lo})/(2\times1.96)$. Panels (a)--(c) carry the same meaning in Figure~\ref{fig:los_results}. All metrics are held-out cross-validated with the feature sets fixed; patient-grouped estimates, which reflect unseen patients, are in Supplementary Table~\ref{tab:groupcv}.}
\label{fig:los_results_hapi}
\end{figure}

\begin{figure}[ht!]
\centering
\includegraphics[width=\linewidth]{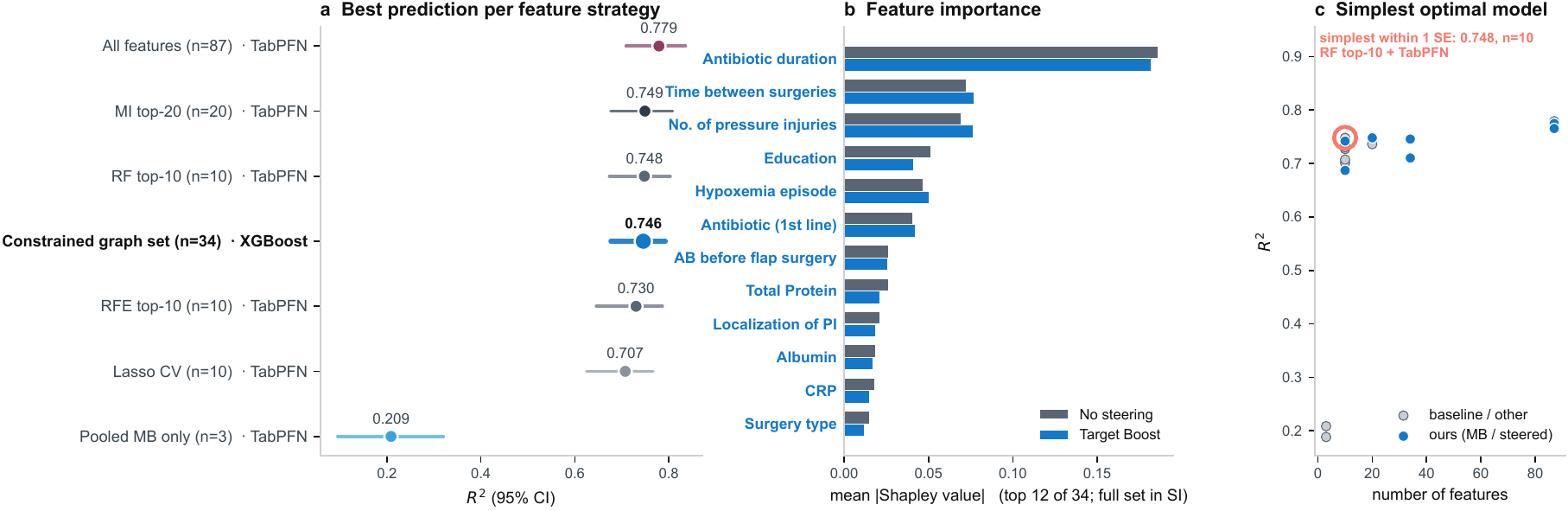}
\caption{\textbf{PPIS length of stay ($n=266$): the constrained graph set matches the best automatic feature selectors.} Higher $R^2$ is better; panels (a)--(c) as defined in Figure~\ref{fig:los_results_hapi}. The 34-variable graph set reaches $R^2=0.746$ (XGBoost; $0.710$ with TabPFN), level with mutual-information and random-forest selection ($0.749$, $0.748$) and within $0.033$ of all 87 variables ($0.779$). Because the blanket includes during-stay variables, this is an explanatory model of the realized stay; models restricted to admission or pre-surgery information are in Supplementary Table~\ref{tab:decision_time}. Every member is already a temporal ancestor of length of stay, so steering within the set leaves it unchanged. Steering the full feature pool toward the Markov blanket helps instead (Figure~\ref{fig:steering_synthesis}).}
\label{fig:los_results}
\end{figure}

We extended the benchmark to include PAG-guided attention steering as an additional method family applied to TabPFN~\cite{Hollmann2025TabPFN}. The effective mode is $y\!\to\!x$, which amplifies the $Y$-token's attention to a causal target set identified in the constrained PAG and so gates which features the prediction reads from. We use two target sets and keep them distinct throughout, boosting toward the target's graph ancestors (Target-Boost) or toward its Markov blanket (MB-Boost). Feature--Feature (FF) steering, which instead reshapes feature-to-feature routing, is included only as a control. Because it never touches the $Y$-token read-out, it is inert-to-harmful for prediction (Supplementary Table~\ref{tab:clinical_mode_grid}), and we do not use it as a predictive knob. The deployable steering variant uses fixed $\gamma=2.0$, chosen by evaluating performance across a range of $\gamma$ values on independent synthetic and Sachs data (not on HAPI or PPIS) and selecting the value that produced consistent gains with minimal observational cost. The baseline gradient-boosted ensemble (best of XGBoost/LightGBM/CatBoost/RF) and plain TabPFN are included as non-steered comparators. Three regression metrics ($R^2$, RMSE, MAE) on log-transformed LoS with 5-fold CV and 300-bootstrap 95\% confidence intervals are shown in Figures~\ref{fig:los_results_hapi} and~\ref{fig:los_results}.

Length-of-stay predictability depends on when the prediction is made and on the cohort. It is already strong at admission for HAPI ($R^2\approx0.69$) but weak at admission for PPIS ($R^2\approx0.31$--$0.34$). PPIS reaches the higher values reported below only once during-stay information is admitted, so the strong PPIS figures are retrospective/real-time rather than prospective. On PPIS, the all-feature model leads on length of stay with $R^2=0.779$ under leak-free evaluation, in which imputation and any data-driven selection are performed inside each fold (Methods). TabPFN on the constrained graph set is competitive at $R^2=0.710$ (34 features, $\alpha=0.05$), though slightly behind the gradient-boosted graph set ($R^2=0.746$). Steering within the graph-set vector is a no-op because all 34 members are temporal ancestors of length of stay. Redistributing attention among uniformly-relevant tokens cannot help (tier saturation), and a fixed strength only dilutes. TabPFN on the full feature pool does benefit from steering toward the causal Markov blanket. On this highest-dimensional outcome (87 candidate features), Markov-blanket steering yields a significant gain (fixed-strength $\Delta R^2=+0.029$ at $\gamma=2.0$, rising to a $+0.044$ plateau and significant across the whole strength range; Figure~\ref{fig:steering_synthesis}), consistent with a soft feature-selection effect that concentrates attention on the causal subset and away from the 53 noisier predictors. The gain is causally specific. Equally-sized mutual-information, random-forest, or random subsets do not reproduce it, and steering toward the ancestors alone is harmful, as expected in a covariate-shift regime, where the correct target is the Markov blanket. This $R^2$ remains a retrospective, real-time figure. The blanket includes during-stay variables (complication, ICU stay, antibiotic duration and time between surgeries) that are legitimately available for real-time but not admission-time prediction, where performance falls to $R^2\approx0.31$--$0.34$ (Supplementary Table~\ref{tab:decision_time}). On this small cohort, the deployable per-fold $\gamma$ selection does not reliably capture the $+0.029$ and under-selects to a non-significant $+0.009$ (Supplementary Table~\ref{tab:nested_gamma}), so we present it as a promising but preliminary observational signal awaiting larger cohorts.

On HAPI, TabPFN with broad feature sets (All features, $R^2=0.761$) modestly outperforms the constrained FCI MB with gradient-boosted models ($R^2=0.684$) and the TabPFN FCI MB ($R^2=0.745$, 21 features). Deployable fixed-$\gamma=2.0$ steering adds no gain on HAPI either. Steering the full feature pool toward the blanket costs $-0.020$ in $R^2$ (Supplementary Table~\ref{tab:nested_gamma}), and steering within the 21-feature tier-saturated blanket is within noise ($0.745\rightarrow0.751$ Target-Boost, $-0.006$ FF), consistent with HAPI's tier structure, which places all features as temporal ancestors of length of stay and leaves no scope for beneficial attention redistribution. On its own, the pooled-shared Markov blanket (three features) was, as expected, a poor predictor ($R^2=0.52$ on HAPI, $R^2=0.21$ on PPIS), confirming that the cross-cohort causal claim must be embedded into a richer per-cohort search to become predictively useful.

\textbf{Decision-time-stratified length of stay.} Because the strong PPIS length-of-stay result draws on variables observed only as the stay unfolds, we stratified each LoS feature set by temporal tier into admission-only, pre-surgery (adding beginning-of-stay laboratory values), and retrospective (the full blanket, adding during-stay variables), and re-evaluated each (Supplementary Table~\ref{tab:decision_time}). The two cohorts behave very differently. On HAPI, admission-only features already reach $R^2=0.69$ (TabPFN), rising only modestly to $0.72$ (pre-surgery) and $0.75$ (full blanket). HAPI length of stay is largely set by baseline severity and function and is predictable at admission. On PPIS, admission-only and pre-surgery features reach only $R^2\approx0.31$--$0.34$ with TabPFN ($0.31$--$0.36$ across learners; Supplementary Table~\ref{tab:decision_time}), whereas the full retrospective blanket reaches $0.71$. Almost all of the PPIS signal comes from during-stay events (complication, ICU stay, immobilization and antibiotic duration). The headline PPIS LoS figure is therefore an explanatory model of realized stay, and only the admission/pre-surgery models ($R^2\approx0.31$--$0.34$) are prospectively deployable. We further probed the robustness of the reported results with patient-grouped and fully nested cross-validation, prediction calibration and average precision, steering negative controls, imputation-method and age-harmonization sensitivity, and graph-stability analyses (Supplementary Tables~\ref{tab:groupcv}--\ref{tab:stability}). Fully nested cross-validation, which re-derives the causal graph and every method's feature set inside each training fold, shows that the discovery procedure itself generalizes. The re-discovered blanket stays competitive (first on PPIS length of stay, second on UTI), indicating that the fusion recovers predictive causal structure from new data without overfitting the development cohort. Absolute numbers fall for all methods once features are selected without seeing the held-out fold, and the constrained FCI Markov blanket's PPIS length-of-stay $R^2$ settles at $0.674$ (Supplementary Table~\ref{tab:nestedcv}). An age-harmonization sensitivity analysis further shows the Age--LoS pooled edge is stable once age is harmonized consistently across cohorts. We also evaluated a HAPI pressure-injury classification endpoint ($\mathrm{PI\_onset}=[\text{PI stage}\ge 1]$) and rejected it as confounded. Even a Markov blanket that correctly excludes its definitional source (PI stage) attains a near-perfect AUROC ($\approx0.95$). The clinical predictors' univariate discrimination is only modest ($0.58$--$0.81$); the AUROC is driven by differential missingness between the injury and no-injury sub-cohorts, so the model detects measurement regime rather than risk. We therefore do not report it. The case shows that leakage auditing must extend to the missingness pattern itself as well as to feature values.

\subsection*{PAG-guided attention steering for the PPIS binary post-surgical outcomes}

\begin{figure}[p]
\centering
\includegraphics[width=\linewidth]{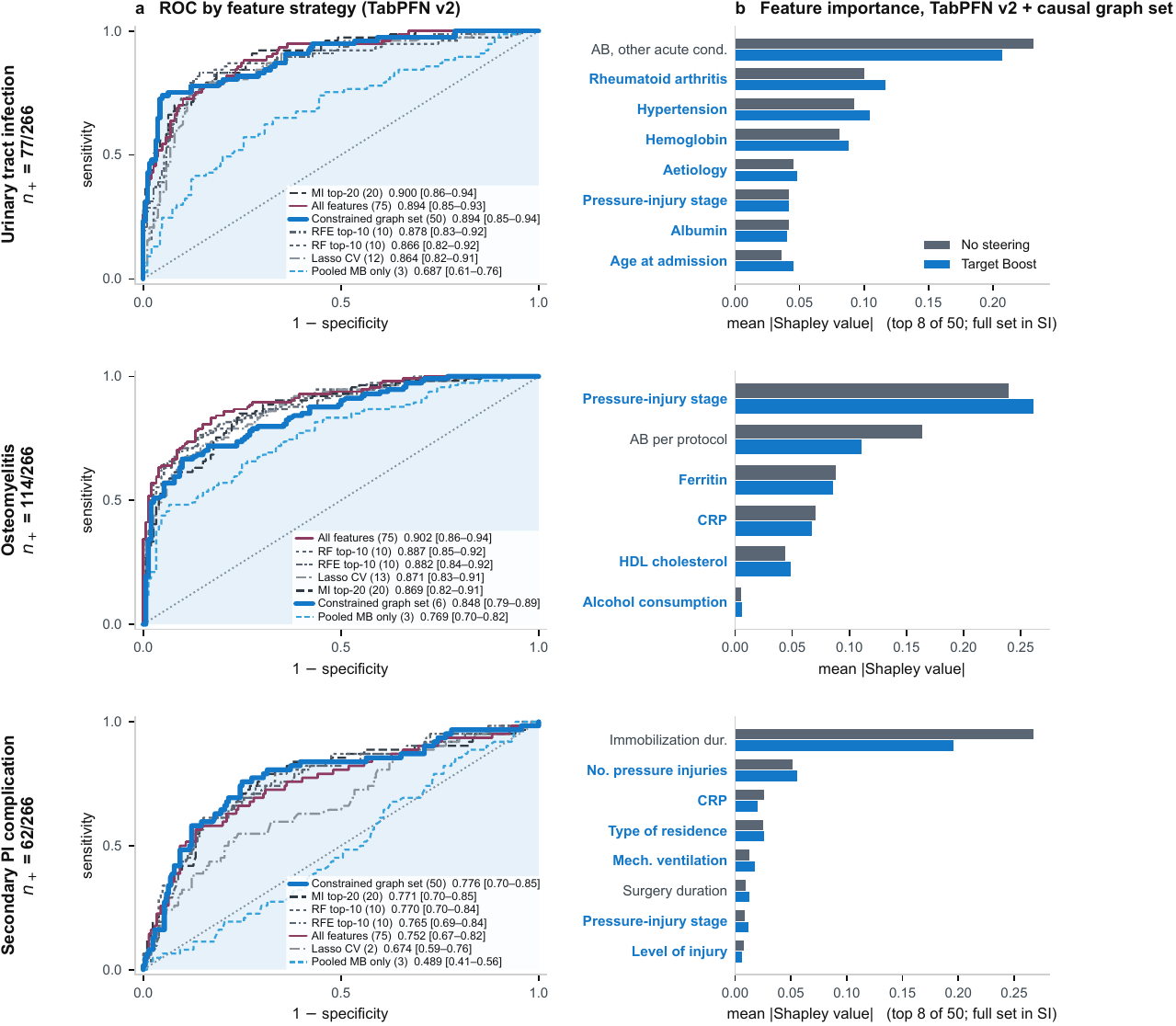}
\caption{\textbf{PPIS post-surgical outcomes: the constrained graph set is the best feature set for secondary PI complication and ties the best for urinary tract infection.} One row per outcome: urinary tract infection during stay ($n_{+}=77/266$), osteomyelitis ($n_{+}=114/266$), and secondary PI complication ($n_{+}=62/266$). \textbf{(a)}~Out-of-fold ROC curve for each feature-selection strategy, with the constrained graph set highlighted; the legend ranks the strategies and gives AUROC with its 95\% bootstrap confidence interval, and the dotted diagonal is chance. All curves share one model (TabPFN\,v2), so the panel isolates the feature set from model choice. The data-driven selectors are re-fitted within each training fold, so their feature counts are per-fold medians. \textbf{(b)}~Mean $|\mathrm{SV}|$ Shapley importance for that model on the constrained graph set, without steering (grey bars) and with Target-Boost (blue bars); feature names printed in blue are temporal (tier-earlier) causal ancestors of the target. Only the eight strongest features are shown; the blankets hold 50 features for UTI and secondary PI complication and six for osteomyelitis, with the full membership in Supplementary Table~\ref{tab:mb_features}. Osteomyelitis is the one outcome where automatic selection leads, its dominant predictors being non-ancestors of the target.}
\label{fig:ppis_uti}
\label{fig:ppis_om}
\label{fig:ppis_classification}
\end{figure}

We evaluated PAG-guided attention steering on three binary post-surgical outcomes in PPIS: urinary tract infection during stay (UTI, $n_{+}=77/266$), osteomyelitis (OM, $n_{+}=114/266$), and secondary PI complication ($n_{+}=62/266$). The fusion pipeline was re-run with each binary outcome as the target. Constrained Markov blankets and feature-selection baselines were evaluated with gradient-boosted ensembles and TabPFN, the latter with and without attention steering. AUROC, AUPRC, and F1 at threshold 0.5 were computed on stacked out-of-fold predictions with 300-bootstrap 95\% confidence intervals (Supplementary Table~\ref{tab:binary_metrics}). Figure~\ref{fig:ppis_uti} holds the model fixed (TabPFN\,v2) so that panel~(a) isolates the feature-set effect.

\begin{figure}[!tbp]
\centering
\includegraphics[width=\linewidth,height=0.9\textheight,keepaspectratio]{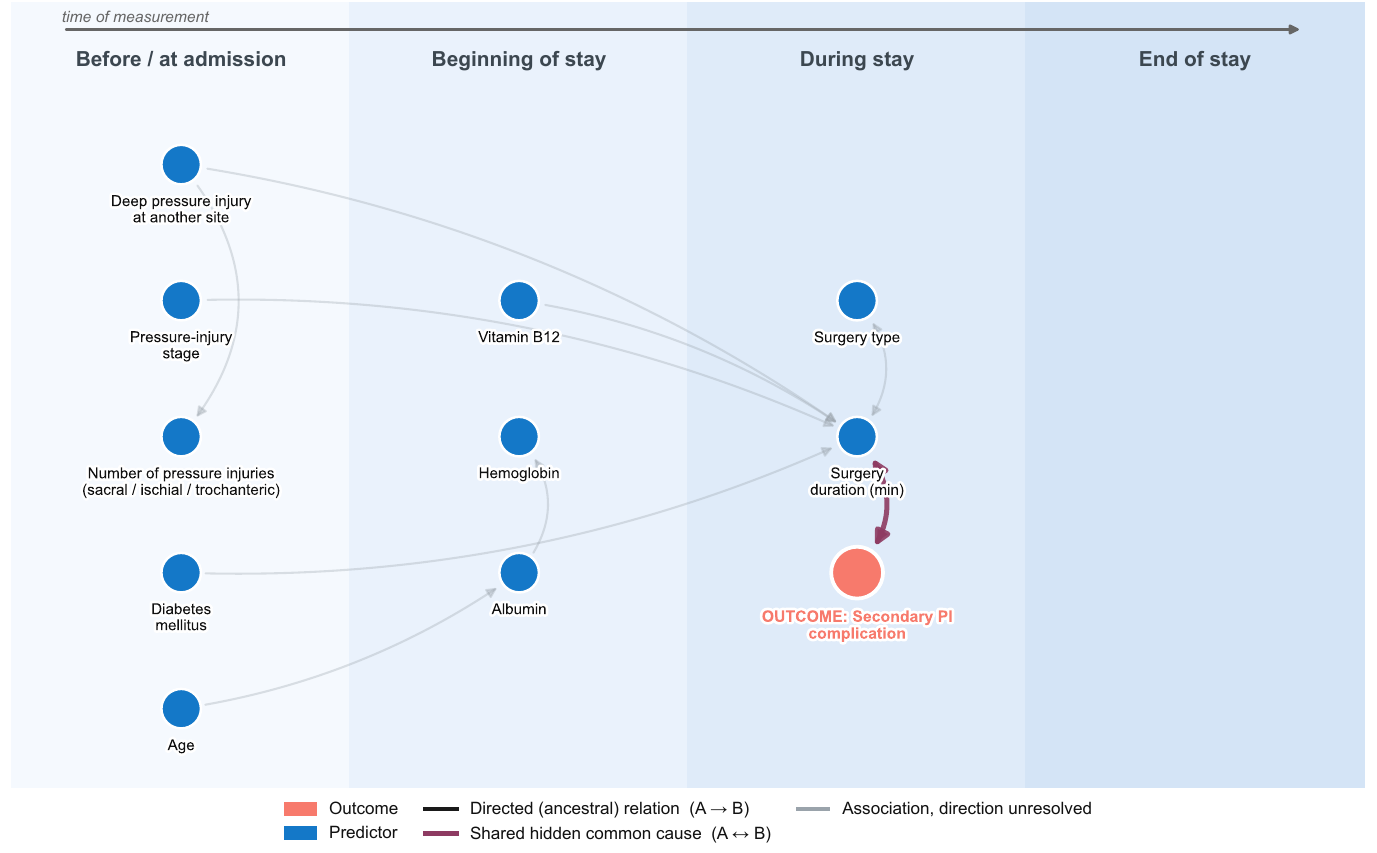}
\caption{\textbf{Temporally-organized causal graph for secondary PI complication (PPIS; depth-2 neighborhood of the outcome, $\lvert\mathcal{N}\rvert=10$ of a 50-member Markov blanket).} The learned graph reads left to right along the clinical timeline. Each circle is a clinical variable, placed in the column for when it is measured (before/at admission $\rightarrow$ beginning of stay $\rightarrow$ during stay $\rightarrow$ end of stay). The blue circles are the outcome's depth-2 temporally-available causal neighborhood (Methods). The prediction model itself conditions on the full 50-feature Markov blanket of this outcome (its total-conditioning, temporally-restricted m-separating set in the constrained $\alpha=0.05$ fused PAG). To keep the graph legible we draw only the 10 blanket members lying within two steps of the outcome, so every node shown is one of the 50 blanket features; the complete 50-member list is in Supplementary Table~\ref{tab:mb_features}. The coral circle is the outcome. Edges follow partial-ancestral-graph semantics. A solid dark arrow $A\rightarrow B$ denotes a directed (ancestral) relation under FCI/PAG assumptions, with $A$ a possibly indirect cause of $B$. A purple double-headed arrow $A\leftrightarrow B$ denotes latent confounding by a shared hidden common cause, and a light grey line denotes an association whose direction the data could not resolve. Edges incident to the outcome are drawn boldly; predictor--predictor links are faded to background context. This figure is the reading key for the per-target causal-neighborhood graphs of the other outcomes (Supplementary Figures~\ref{fig:mbgraph_hapi_los}--\ref{fig:mbgraph_ppis_om}).}
\label{fig:mbgraph_comp}
\end{figure}

For the binary outcomes we take the constrained graph set to be the total-conditioning Markov blanket of the target in the constrained ($\alpha=0.05$) fused PAG, restricted by m-separation to variables in temporal tiers no later than the target's. This restriction removes clear post-target descendants (for example, a raw Markov blanket can contain treatments administered because the outcome occurred) and thereby reduces obvious future-information leakage. It does not, however, establish prospective validity, because outcome-onset dates are not recorded in these cohorts, and several retained predictors (antibiotic duration, number of antibiotics, administration of antibiotics, duration of immobilization) accumulate over the stay and may in part follow outcome onset. We therefore describe these models as \emph{temporally-constrained retrospective classifiers} rather than prospective predictors. They use only variables in tiers up to the target's, but a prospective, index-time evaluation (a fixed cut such as pre-flap-surgery, immediately post-surgery, or day~7) would require onset timestamps we do not have. Establishing a per-task index time and reclassifying every predictor against it is a priority for prospective revision.

With all methods drawing from the same temporally-valid, leakage-safe feature pool, the constrained graph set ranges from competitive to best against automatic feature selection. For UTI it reaches AUROC$=0.894$ with 50 features, effectively tied with the strongest baseline (mutual-information top-20, $0.900$) and the all-feature model ($0.894$); under patient-grouped cross-validation, the primary PPIS estimate, it reaches $0.778$ (Supplementary Table~\ref{tab:groupcv}). For the secondary-PI-complication flag the graph set is best on AUROC and average precision (AUROC$=0.776$ and AUPRC$=0.520$ on a 50-feature Markov blanket, versus $0.752$/$0.515$ for all features and $0.771$/$0.520$ for mutual information), while F1 at a $0.5$ threshold is indistinguishable from the baselines ($0.485$ versus $0.486$--$0.500$, overlapping bootstrap intervals; Supplementary Table~\ref{tab:binary_metrics}). Figure~\ref{fig:mbgraph_comp} draws the 10-node causal neighborhood of this outcome. For osteomyelitis the graph set reaches AUROC$=0.848$ (6 features; $0.807$ patient-grouped) against $0.87$--$0.90$ for the automatic selectors. Osteomyelitis is the one outcome where a baseline leads. The baselines' wider lead on an unrestricted feature pool reflects antibiotic leakage, and once the leakage-safe pool is imposed on every method the gap collapses (e.g.\ Lasso $0.944\rightarrow0.871$), leaving the six-feature graph set as the parsimonious, temporally-safe alternative. Average-precision preserves this ordering and stays above prevalence (UTI $0.829$ at prevalence $0.29$; osteomyelitis $0.811$ at $0.43$; secondary PI complication $0.520$ at $0.23$). Figure~\ref{fig:ppis_uti} shows AUROC with 95\% bootstrap confidence intervals, and Supplementary Table~\ref{tab:binary_metrics} tabulates AUROC, average precision and F1 for every feature set. Attention steering adds no observational gain on these binary outcomes. At the fixed deployable strength the AUROC changes are within noise, and the leak-free per-fold $\gamma$ selection, the deployable protocol, declines to steer. It chooses $\gamma\approx0$ in the majority of folds and leaves performance within noise of the unsteered model (all three deltas at most $0.006$, with bootstrap intervals spanning zero; Supplementary Tables~\ref{tab:calibration},~\ref{tab:nested_gamma}). This null is expected, because these single-center cohorts exhibit covariate rather than mechanism shift (a train-versus-test domain classifier separates patient folds at AUROC$\approx0.68$, yet the outcome mechanism $P(y\mid x)$ is stable), the regime in which steering toward causal ancestors offers little. On synthetic ground truth with a controlled mechanism shift, by contrast, the same deployable per-fold selection significantly improves interventional accuracy ($+0.011$ to $+0.029$ mean AUROC, $p<10^{-5}$ over 144 structural causal models; Supplementary Table~\ref{tab:steer_significance}) and beats same-density random-graph and shuffled-label controls. The clinical null therefore reflects the shift type and cohort size and does not indicate a failure of the steering mechanism.

\subsection*{Markov-blanket subgraphs and feature importance}

The Markov-blanket subgraph for each prediction target shows the learned graph restricted to the target and its Markov-blanket neighbors, organized along the clinical timeline. The graph for secondary PI complication is shown in Figure~\ref{fig:mbgraph_comp} as a worked example; the remaining targets are provided in Supplementary Figures~\ref{fig:mbgraph_hapi_los}--\ref{fig:mbgraph_ppis_om}.

Feature importance is quantified using marginal Shapley values, in which each feature's contribution is the average change in prediction when it is added to a random subset of the other features. To preserve the sequence length required for the attention steerer's penalty tensor alignment, inactive features stay in the sequence and are set to their training-set mean. This marginal (mean-masking) approach ensures that steered and unsteered Shapley values are comparable, since the steerer's penalty tensor dimension matches the full feature sequence throughout all coalition evaluations.

In addition to Markov-blanket membership, we quantify each variable's contribution using marginal Shapley values (mean-masking imputer, 10 random permutations per test sample, 5-fold CV, fixed $\gamma=2.0$ for steered variants). We report mean $|\mathrm{Shapley\ value}|$ per feature and the ancestor/non-ancestor attribution ratio (total $|\mathrm{SV}|$ mass on causal ancestors vs.\ on non-ancestors) as a mechanistic indicator of whether steering correctly redirects attribution. Results are shown in Supplementary Figures~\ref{fig:importance_hapi_los}--\ref{fig:importance_comp}.

The Shapley analysis shows three mechanistic patterns. First, steering measurably shifts the ancestor/non-ancestor attribution ratio, but the effect is mixed across settings and does not uniformly favor feature--feature steering. FF produces the largest increase in some feature sets (Complication RF top-10: $0.21\rightarrow0.45$, $+114\%$; UTI Lasso $+21\%$), whereas Target-Boost exceeds FF in every panel built on the constrained FCI Markov blanket (UTI Boost $+18\%$ vs FF $+13\%$; osteomyelitis Boost $+55\%$ vs FF $-24\%$; Complication Boost $+37\%$ vs FF $+6\%$). This is consistent with the two modes acting differently (FF restructures feature representations while Target-Boost directly gates the target read-out) but does not support a claim that either mode reliably dominates. Second, the attribution shift does not track a predictive gain. Under leak-free per-fold evaluation even the set with the largest ratio increase (Complication RF top-10) shows an $R^2$/AUROC change indistinguishable from or below zero, so a redirected attribution is not by itself beneficial on these cohorts. Third, tier saturation is visible in the Shapley plots. All-ancestor settings (every HAPI and PPIS length-of-stay feature set) show no ancestor/non-ancestor ratio change under any steering mode, confirming the softmax-invariance argument. The osteomyelitis panel is the clearest instance of the second pattern. Target-Boost most strongly concentrates attribution on the causal ancestors (ratio $2.73\rightarrow4.24$, $+55\%$) while the feature--feature control moves it the other way ($\rightarrow2.08$, $-24\%$), yet the leak-free accuracy is unchanged (nested $-0.001$).

The effect is interpretable and specific in two respects. First, the benefit scales with the completeness of the causal structure supplied. On synthetic ground truth the interventional gain of the $y\!\to\!x$ boost grows monotonically as more of the true graph is revealed to the steerer (Figure~\ref{fig:anc_frac_reveal}, Supplementary), and the gain is causally specific, since steering toward equally-sized non-causal (mutual-information, random-forest) or random feature sets does not reproduce it. Second, the optimal steering target follows the distribution-shift type. Under mechanism (concept) shift, where spurious feature--outcome relationships break, steering toward the causal ancestors is best. Under covariate shift, where the outcome mechanism is stable, steering toward the full Markov blanket is best, since it retains the still-valid non-ancestor predictors that ancestors alone would discard. The single-center clinical cohorts exhibit covariate rather than mechanism shift, so by these rules the Markov blanket is the appropriate target there.

Table~\ref{tab:mb_features} (Supplementary) lists the full temporally-available causal Markov blanket for each of the five prediction targets; the per-target graphs instead draw the depth-2 neighborhood of the outcome for legibility (a subset), most visibly for the two large binary blankets, UTI and complication.

\FloatBarrier
\section*{Discussion}
This study reframes incomplete multi-source clinical data analysis as a joint problem of imputation, causal discovery, and cross-cohort information fusion. Across synthetic and benchmark settings, coupling these components significantly improved cross-cohort graph recovery, reduced imputation error, and better preserved empirical distributions, with the largest gains under higher missingness. The fusion operator dominates the precision/recall frontier, delivering both the highest-precision and the best-F1 recovery of the shared skeleton relative to fair-pooling and per-cohort baselines across every missingness and heterogeneity level tested (paired Wilcoxon $p<10^{-3}$; Supplementary Figures~\ref{fig:fusion_benchmark} and~\ref{fig:fusion_robustness}). In the SCI application, the proposed pipeline yielded three forms of result that together support its utility: a cross-cohort structural claim represented by three temporally-oriented directed (ancestral) dependencies into length of stay (Age, Albumin, Hemoglobin), an information-fusion trajectory that couples graph evolution with predictive performance, and constrained graph sets that are comparable to automatic feature selection while remaining interpretable and temporally organized. They are read directly from the fused causal graph, so every predictor carries its position relative to the outcome in that graph, which automatic selectors do not provide. These findings suggest that treating imputation, structure learning, and source integration as mutually dependent tasks can improve interpretability and yield leakage-aware feature sets relative to handling them in isolation.

The three pooled-shared directed-ancestor determinants of LoS are each well supported by the SCI literature. Hypoalbuminemia is a well-established marker of sarcopenia and systemic catabolic stress, and is repeatedly associated with delayed wound healing and complicated post-surgical recovery in pressure-injury cohorts \cite{Cruz2018Hypoalbuminemia,Fahndrich2024FlapComplications,Vecin2022PressureInjuries}. Hemoglobin captures oxygen-carrying capacity and tissue perfusion, both of which influence wound healing tolerance and rehabilitation pace. Age at SCI conditions both baseline physiological reserve and the cumulative duration of risk for secondary complications such as autonomic dysreflexia and cardiovascular morbidity \cite{Krassioukov2009Cardiovascular,Boueri2024ADReview}. The pipeline orients all three variables as directed ancestors of LoS. This orientation reflects the temporal-tier constraint that admission-time covariates cannot be caused by an end-of-stay outcome \cite{Meek1995CausalKnowledge}, and it is also consistent with the underlying clinical mechanism, in which nutritional, hematological, and demographic baselines determine the trajectory of in-hospital recovery. The three relations differ, however, in their stability under cross-cohort harmonization. Both cohorts record age at SCI, but the reported pipeline mapped HAPI age at SCI to PPIS age at admission, a suboptimal choice. When age is harmonized consistently (age at SCI in both cohorts, or within-cohort standardization that removes the anchor and scale mismatch), the Age--LoS and Hemoglobin--LoS pooled adjacencies are recovered in every bootstrap resample, and Albumin--LoS in a minority (albumin is missing in ${\sim}19$--$37\%$ of records). Only a units-inconsistent age-at-admission construction destabilizes Age--LoS (Supplementary Table~\ref{tab:stability}). We therefore retain all three as supported pooled-shared relations, while noting that age should be harmonized as age at SCI in both cohorts and that the pooled analysis is a hypothesis-generating scaffold given the moderate sample sizes.

The per-cohort graphs at the full fusion stage retain the three pooled-shared directed-ancestor variables of LoS and add cohort-specific neighbors that match the clinical focus of each cohort. HAPI's Markov blanket includes rehabilitation- and pressure-injury-history variables (SCIM, PI history number, current PI time since last PI, first rehabilitation status, sedative medications), consistent with HAPI's role as an upstream rehabilitation cohort capturing risk accumulation and pressure-injury occurrence \cite{Najmanova2022HAPI,Marin2013RiskFactors}. PPIS's Markov blanket includes perioperative and post-surgical variables (number of bacteria, antibiotic duration, hypoxemia episode, weight change during rehabilitation, complication, ICU stay), consistent with PPIS's role as a post-pressure-injury surgery cohort capturing severity, intervention, and recovery \cite{Fahndrich2024FlapComplications,Vecin2022PressureInjuries}. The fusion framework, therefore, recovers known determinants while organizing them into a multistage dependency structure spanning upstream HAPI risk and downstream PPIS severity, intervention, and treatment trajectory. We do not interpret these variables as newly discovered, standalone causal factors; the more defensible contribution is structural.

A practical value of the temporally-organized Markov blankets is that they surface candidate modifiable factors that may inform prevention of secondary pressure injury and its complications after reconstructive (flap) surgery. For the secondary-PI-complication and osteomyelitis targets, the blankets are dominated by nutritional and inflammatory status (albumin, hemoglobin, CRP), antibiotic exposure and duration, and the duration of post-operative immobilization, each a candidate determinant of flap failure and wound breakdown in the SCI and pressure-injury literature \cite{Cruz2018Hypoalbuminemia,Fahndrich2024FlapComplications,Vecin2022PressureInjuries}. Recurrence-specific evidence supports this alignment. A systematic review and meta-analysis of pressure-injury reconstruction in SCI, and cohort studies of primary flap surgery, identify nutritional status (albumin), infection, prior reconstruction, and post-operative immobilization/mobilization as leading predictors of post-surgical complications and re-ulceration \cite{Lopez2026Recurrence,Morel2019Recurrence,Sgarzani2023Reconstructive}, the same variable categories that our target-specific Markov blankets surface. The model's outputs are thus compatible with established prevention bundles: pre-operative correction of hypoalbuminemia and anemia, antibiotic stewardship guided by infection markers, and structured off-loading and mobilization protocols \cite{Gefen2014,Marin2013RiskFactors,Najmanova2022HAPI}. Because these variables are placed in a temporally-ordered graph, and the predictive feature set is restricted by m-separation to information available at the decision point, a clinician can read off which admission- and beginning-of-stay factors are upstream of the outcome and therefore candidates for prospective evaluation (for example, optimizing albumin before flap surgery, or shortening avoidable immobilization), and can avoid acting on a downstream correlate that cannot be changed prospectively. The graph remains hypothesis-generating. It identifies plausible, literature-consistent candidate intervention hypotheses whose causal effect must be confirmed in prospective or interventional studies before clinical adoption. Renal function, a clinically expected determinant, is among the best-recorded information in PPIS (creatinine missing in $1.1\%$ of episodes and GFR in $2.6\%$, against $19.2\%$ for albumin), and creatinine, GFR and cystatin C all entered the search as beginning-of-stay (tier-2) candidates under the same constraints as every other laboratory value. At $\alpha=0.05$ none of the three retains a single edge in the pooled, per-cohort or fused graphs. At $\alpha=0.10$ they instead form a coherent renal block, with creatinine, cystatin C and GFR mutually adjacent, congestive heart failure adjacent to all three, and albumin and CRP entering through cystatin C. The isolation at the deployable threshold is therefore a matter of resolution. Neither threshold yields a path to an outcome. No renal marker is adjacent to length of stay, secondary complication, urinary tract infection or osteomyelitis, and neither the complication nor the osteomyelitis blanket contains one. Marginally, lower GFR is weakly associated with secondary complication in the expected direction (Spearman $\rho=-0.14$, $p=0.03$; $\rho=-0.15$ given creatinine), an association that does not survive the multivariate conditioning that FCI performs. Because impaired renal function is an established correlate of delayed wound healing, we report this as a limit of what these cohorts can resolve rather than as evidence of no effect. The renal signal is internally consistent, but its link to post-surgical outcomes lies below the resolution of 266 episodes across up to 87 variables.

Figure~\ref{fig:trajectory} shows the contribution of each pipeline ingredient. Stages~3 and~4 (temporal-tier and pooled-shared constraints) do not increase $R^2$ uniformly. Stage~3 trims the PPIS Markov blanket from 23 to 20 features while preserving performance, and Stage~4 restores the three shared directed-ancestor variables ($20\rightarrow23$), suggesting the tier rules remove spurious conditional-independence violations and spare informative signal. A constraint-strategy ablation localizes this to a single operation. The temporal-tier rules constrain orientation only, so forbidden-only, required-only and both give identical blankets. The pooled-shared stage bundles three operations: orientation constraints from the pooled subgraph, forbidding variable pairs absent from it, and unioning its edges into the per-cohort PAG. Decomposing them shows the first two to be inert. Withholding the edge union reproduces the unpooled result exactly ($R^2=0.694$ on HAPI and $0.743$ on PPIS, with identical Markov blankets), and applying the union alone, with or without the constraints, reproduces the full result exactly ($0.718$ and $0.735$). Only cross-cohort edge fusion therefore changes prediction. It improves HAPI length-of-stay $R^2$ by $+0.024$ while slightly lowering PPIS ($0.743\rightarrow0.735$, overlapping intervals), consistent with pooled evidence mattering most on the data-poor cohort whose compact blanket rests on the shared edges (Supplementary Table~\ref{tab:constraint_ablation}). Relaxing the conditional-independence threshold beyond the deployable operating point, to $\alpha=0.10$, enlarges the PPIS blanket and raises its retrospective $R^2$ for the flexible learners while collapsing the penalized linear ones (Supplementary Table~\ref{tab:alpha_trajectory}). Because of this model-dependence we retain $\alpha=0.05$, whose graph set is both more parsimonious and more stable ($R^2=0.71$, 34 features). Either way the length-of-stay figure is explanatory rather than prospective (Supplementary Table~\ref{tab:decision_time}). The pattern argues for reporting the full trajectory, since each stage corresponds to a distinct methodological choice with a measurable predictive effect.

On PPIS, the all-feature model gives the best retrospective length-of-stay fit ($R^2=0.779$ with TabPFN), with the constrained FCI Markov blanket (34 features, $\alpha=0.05$) competitive at $R^2=0.710$. Gradient boosting ($R^2=0.746$) slightly edges TabPFN on this graph set. This fit is an explanatory ceiling, not a deployable prediction. The blanket is dominated by during-stay variables (complication, ICU stay, immobilization and antibiotic duration), and restricting the predictor to admission or pre-surgery information reduces it to $R^2\approx0.31$--$0.34$ (Supplementary Table~\ref{tab:decision_time}). Attention steering within these tier-saturated blankets provides no gain, because every feature is already a temporal ancestor and a fixed strength only dilutes. Leak-free per-fold $\gamma$ selection then recovers the baseline (Supplementary Table~\ref{tab:nested_gamma}). The only configuration that helps is steering the full feature pool toward the Markov blanket on PPIS length of stay, with a significant and causally specific $R^2$ gain of $+0.029$ at $\gamma=2.0$ (Figure~\ref{fig:steering_synthesis}), which we present as a promising but preliminary observational signal on a small cohort. On HAPI the constrained Markov blanket trailed the all-feature baseline by about $0.02$ in $R^2$, with no steering gain. For the three PPIS binary outcomes the temporally-constrained causal Markov blanket is competitive with automatic feature selection and, on secondary PI complication, the best set of all (AUROC $0.776$, versus $0.752$ for all features and $0.771$ for the strongest automatic selector). On UTI it is competitive ($0.894$, against $0.900$ for the best automatic set), and on osteomyelitis the compact six-feature blanket ($0.848$) trails the broader automatic sets ($0.87$ to $0.90$). The osteomyelitis gap illustrates the trade-off most clearly. The automatic advantage comes from during-stay markers (antibiotic use, inflammatory laboratory values) that risk leakage and that the temporal restriction excludes, so the constrained FCI Markov blanket trades a raw predictive edge for interpretability and temporal safety.

The framework is relevant because missing-data handling is often treated as external to causal discovery. Generic imputation can attenuate dependencies or introduce artificial associations, especially when data arise from heterogeneous and only partially overlapping sources \cite{Stekhoven2012MissForest,White2011MICE,Ramosaj2020Cautionary}. By iteratively updating imputations using Markov blanket information and then re-estimating graph structure, the proposed workflow uses local causal context to stabilize both tasks \cite{Pellet2008MB,Aliferis2010LocalCausal,Yu2018MBNoCausalSufficiency,Liu2023MBMF,Yu2022MimMB}. The pooled-shared constraint step provides an additional advantage by injecting cross-cohort evidence into each per-cohort search without forcing all datasets to share an identical population PAG, which is the standard target of integrative approaches \cite{ion_paper,Tsamardinos2012Integrative,Mooij2020JCI,Triantafillou2015COmbINE}. The downstream prediction results are encouraging in this regard, because they show that the learned neighborhoods are both clinically interpretable and practically informative. A further reason to target a PAG rather than a DAG is clinical. Determinants of flap outcome that surgeons regard as central, such as operative technique, the anatomical difficulty of achieving closure and operator experience, are recorded in neither registry, and no amount of harmonization or imputation recovers them. FCI is designed for this regime. It does not assume causal sufficiency, and it represents a suspected unmeasured common cause as a bidirected edge. The unmeasured confounding that a clinical reader would rightly raise against these cohorts is therefore encoded in the output.

Several limitations should temper interpretation. First, the HAPI and PPIS datasets capture related but distinct parts of the pressure-injury trajectory, so their integration should be understood as structural synthesis across complementary cohorts that do not form a single homogeneous population. Second, variables such as laboratory markers, functional scores, and complication indicators reflect a mixture of baseline vulnerability, evolving clinical severity, and care-process effects, which limits strict causal interpretation at the level of individual predictors. Most consequentially, neither registry records surgical technique, the anatomical difficulty of closing the pressure injury, or operator experience, all of which plausibly influence post-surgical complication and osteomyelitis. Targeting a PAG keeps such unmeasured common causes representable as bidirected edges, but it does not measure them, and any complication model built on these cohorts inherits that gap. Third, the real-world analyses do not provide ground-truth causal structure, so the learned PAG should be viewed as a hypothesis-generating representation consistent with the observed data and the assumptions of PAG-based discovery. It is not definitive proof of causation. Fourth, the partially overlapping variable sets and moderate cohort sizes imply that some graph relations may remain sensitive to sampling variability, preprocessing decisions, and choices in conditional-independence testing. This sensitivity is particularly clear for length of stay, where relaxing the conditional-independence threshold from $\alpha=0.05$ to $\alpha=0.10$ grows the HAPI blanket modestly and the PPIS blanket substantially (Supplementary Tables~\ref{tab:stability} and~\ref{tab:alpha_sensitivity}). Fifth, the binary-outcome predictors are temporally constrained, using only variables in tiers up to the target's, but they are not established as prospectively valid. Outcome-onset timestamps are unavailable, so retained predictors such as antibiotic and immobilization duration cannot be confirmed to precede outcome onset. The PPIS length-of-stay blanket (34 features at $\alpha=0.05$) likewise includes during- and post-stay variables (complication, ICU stay, antibiotic duration). A decision-time-stratified analysis (Supplementary Table~\ref{tab:decision_time}) shows that PPIS length-of-stay prediction from admission-only or pre-surgery information reaches only $R^2\approx0.31$--$0.34$, so this retrospective figure is an explanatory ceiling on realized stay, not an admission-time forecast. HAPI length of stay, by contrast, is already well predicted at admission ($R^2=0.69$). Sixth, the 266 PPIS episodes come from 174 patients. Because episode-level cross-validation is optimistic (metrics fall by $0.04$--$0.12$ under patient grouping), we treat the patient-grouped figures (Table~\ref{tab:groupcv}) as the primary PPIS estimates and episode-level results as an optimistic upper bound. Seventh, imputation and the data-driven baseline feature sets (Lasso, mutual-information and random-forest selection) are re-derived inside each cross-validation fold using training data only, so those comparisons are leak-free. The graph-derived Markov blankets are instead extracted from a partial ancestral graph learned once on the full development cohort and then scored on held-out folds, which gives out-of-sample internal-validation estimates for that fixed deployable model. The fully nested cross-validation below addresses the residual graph-selection optimism. The main residual optimism arises because episode-level folds can place different surgical episodes of the same patient in both training and test, so we treat the patient-grouped held-out estimates (Table~\ref{tab:groupcv}) as the deployment-realistic figures. Because the causal-fusion procedure itself is a central contribution, we also verify that it generalizes. A fully nested cross-validation that re-derives every method's feature set inside each fold (Supplementary Table~\ref{tab:nestedcv}) still yields a competitive blanket (first on PPIS length of stay, second on UTI), indicating that the procedure recovers predictive structure from new data and does not merely overfit the development cohort. Repeating the entire pooled$+$temporal-tier fusion inside each fold is computationally intractable on the dense training folds (Supplementary Table~\ref{tab:sl_sensitivity}). Eighth, HAPI and PPIS are both from a single center (the Swiss Paraplegic Center), so the cross-cohort fusion demonstrates transfer across clinical stages within one institution only. External, multi-site validation is essential. Ninth, learning high-dimensional PAGs (up to 87 variables) from a few hundred patients is underpowered and the graphs are $\alpha$-sensitive (Supplementary Table~\ref{tab:stability}), so the learned structure is best regarded as hypothesis-generating. The pipeline's use of Fisher-$z$ on mixed discrete--continuous data (Methods) is a simplification, but a re-run with a kernel test suited to mixed data recovers identical Markov blankets, the three pooled LoS edges survive the kernel test, and the blankets are robust to the imputer (Supplementary Table~\ref{tab:sl_sensitivity}). Finally, because the analysis was conducted in a specialized SCI rehabilitation setting, transportability to other inpatient populations remains to be established.

Causal graph fusion turns fragmented, incomplete clinical records into a single, temporally ordered causal representation. Against exact ground truth it recovers shared structure more accurately than pooling the cohorts or analysing them separately, in every missingness and heterogeneity condition tested, and it handles missingness inside discovery instead of as a preprocessing step that can attenuate or fabricate the dependencies being learned. Causal attention steering then carries the learned structure into a pretrained foundation model without retraining, with a gain under distribution shift that a random graph does not produce and a target rule set by the type of shift. In spinal cord injury, fusion orients age, albumin and hemoglobin as ancestors of length of stay, a set no single cohort orients in full, and organizes each cohort's predictors by clinical stage, from upstream rehabilitation risk to post-surgical recovery. Its compact blankets match or exceed automatic feature selection on four of five outcomes while showing where every predictor sits on the care timeline and how strongly each cross-cohort edge is supported. The upstream factors they surface, such as albumin and hemoglobin before flap surgery, align with established prevention bundles and define concrete hypotheses for prospective and interventional studies. Because the method needs only overlapping variables and knowledge of when each is recorded, it applies wherever medical knowledge is spread across registries, trials and hospitals.

\section*{Methods}

\subsection*{Problem setting}
Let $\{D_1,\dots,D_K\}$ denote related datasets defined over overlapping subsets of variables from a common universe $V$. Each dataset may contain missing entries and may cover only a subset of the full variable space. The goal is to estimate a unified causal representation over $V$ while preserving information contributed by the individual datasets and simultaneously obtaining plausible imputations for missing values.

\subsection*{Method architecture}
The workflow has two coupled stages. In the pooled-shared stage, the union of cohorts $D_1\cup\cdots\cup D_K$ is restricted to the set of variables that appear in every cohort, denoted $V_{\mathrm{shared}}$, and a partial ancestral graph (PAG) is learned on this restriction using FCI \cite{Spirtes2000,Zhang2008Orientation,colombo2012learning} with temporal-tier background knowledge encoded as forbidden orientations \cite{Meek1995CausalKnowledge}. The resulting subgraph $G_{\mathrm{shared}}$ encodes cross-cohort causal evidence. An edge $u\rightarrow v$ in $G_{\mathrm{shared}}$ corresponds to a directed relation that is supported by the pooled distribution after temporal-tier filtering.

In the per-cohort stage, FCI is re-run on each cohort over its full variable set, but now with two sources of background knowledge:
\begin{itemize}
\item Temporal-tier constraints derived from clinical timing (admission-time, during-stay, and discharge tiers).
\item Constraints derived from $G_{\mathrm{shared}}$: every shared variable pair $(u,v)$ in the per-cohort graph is required to match the orientation it has in $G_{\mathrm{shared}}$, and shared pairs without an edge in $G_{\mathrm{shared}}$ have their edge forbidden in either direction.
\end{itemize}
After per-cohort FCI converges, edges from $G_{\mathrm{shared}}$ that involve shared variables are merged back into each per-cohort PAG to ensure the cross-cohort claim is represented even when the per-cohort sample alone does not have enough power to recover it. Markov blankets are then extracted from each augmented per-cohort PAG for downstream prediction. Missing entries are re-imputed within each cohort using a random-forest predictor restricted to each variable's current Markov blanket, and the procedure is iterated until the per-cohort Markov blankets stabilize.

\begin{algorithm}[ht]
\caption{Pooled-shared constrained causal graphical fusion.}
\label{alg:fusion}
\begin{algorithmic}[1]
\Require Cohorts $\{D_1,\dots,D_K\}$ with variable sets $V_1,\dots,V_K$; target variable $Y$; temporal-tier constraint sets $T_1,\dots,T_K$; significance level $\alpha$; convergence threshold $\tau$.
\State $V_{\mathrm{shared}} \gets V_1 \cap \cdots \cap V_K$ \Comment{\textit{Cohort-shared variable backbone}}
\State Initialize missing values in each $D_k$ using dtype-aware median (continuous) / mode (categorical, ordinal) imputation
\State \textbf{Stage 1 -- pooled-shared subgraph.}
\State $D_{\mathrm{pool}} \gets$ concatenate $\{D_k[V_{\mathrm{shared}} \cup \{Y\}]\}_{k=1}^K$
\State $G_{\mathrm{shared}} \gets \textsc{FCI}(D_{\mathrm{pool}}, \alpha, \text{background} = T_{\mathrm{pool}})$ \Comment{\textit{$T_{\mathrm{pool}}$ = pooled temporal tiers}}
\State \textbf{Stage 2 -- per-cohort constrained FCI.}
\For{each cohort $k = 1,\dots,K$}
    \State $B_k \gets$ \textsc{EncodeBackground}$(G_{\mathrm{shared}}, V_k)$ \Comment{Require/forbid orientations per pooled subgraph}
    \State $B_k \gets B_k \cup T_k$ \Comment{Add per-cohort temporal-tier constraints}
    \State $G_k^{(0)} \gets \textsc{FCI}(D_k, \alpha, \text{background} = B_k)$
    \State Merge edges from $G_{\mathrm{shared}}$ over $V_{\mathrm{shared}}$ into $G_k^{(0)}$
    \State $M_k^{(0)} \gets \textsc{MarkovBlanket}(G_k^{(0)}, Y)$ \Comment{Substitute end-stay nodes by their parents}
\EndFor
\State \textbf{Stage 3 -- iterative re-imputation.}
\For{iteration $t = 1, 2, \dots$}
    \For{each cohort $k$ and each variable $v \in V_k$ with missing entries}
        \State Re-impute missing $v$ using random forest trained on its current Markov blanket
    \EndFor
    \For{each cohort $k$}
        \State $G_k^{(t)} \gets \textsc{FCI}(D_k, \alpha, \text{background} = B_k)$
        \State Merge edges from $G_{\mathrm{shared}}$ into $G_k^{(t)}$
        \State $M_k^{(t)} \gets \textsc{MarkovBlanket}(G_k^{(t)}, Y)$
    \EndFor
    \If{$\max_k \lvert M_k^{(t)} \triangle M_k^{(t-1)} \rvert / \lvert M_k^{(t-1)} \rvert < \tau$}
        \State \textbf{break}
    \EndIf
\EndFor
\State \Return $G_{\mathrm{shared}},\ \{G_k^{(t)}\}_{k=1}^K,\ \{M_k^{(t)}\}_{k=1}^K$
\end{algorithmic}
\end{algorithm}

The encoding step \textsc{EncodeBackground}$(G_{\mathrm{shared}}, V_k)$ traverses every variable pair $(u,v)$ with $u,v\in V_{\mathrm{shared}}$ and writes the following constraints into the per-cohort background-knowledge object:
\begin{itemize}
\item $u\rightarrow v$ in $G_{\mathrm{shared}}$: require $u\rightarrow v$, forbid $v\rightarrow u$.
\item $u\leftrightarrow v$: forbid both $u\rightarrow v$ and $v\rightarrow u$.
\item $u\,\circ\!\rightarrow v$: forbid $v\rightarrow u$ only (keep partial orientation).
\item $u\,\circ\!-\!\circ v$: no constraint.
\item No edge between $u$ and $v$: forbid edge in either direction.
\end{itemize}
The merge step in line~11 of Algorithm~\ref{alg:fusion} adds any edge from $G_{\mathrm{shared}}$ over shared variables that is missing in the per-cohort PAG, with the orientation from $G_{\mathrm{shared}}$. This ensures cross-cohort causal evidence is represented even when the per-cohort sample lacks the statistical power to detect the corresponding conditional dependency.

\subsection*{Implementation details}

Missing values were initialized using dtype-aware median imputation for continuous variables and mode imputation for categorical and ordinal-encoded variables (the latter detected by integer-valued columns with $\le 10$ unique values). During iteration, each variable with remaining missing entries was re-imputed using a random forest trained on its current Markov blanket. PAG estimation used the FCI implementation from causal-learn \cite{Zheng2024CausalLearn}, with Fisher-$z$ as the conditional-independence test and a significance level of $\alpha=0.05$ unless otherwise noted. Fisher-$z$ assumes (conditional) linear-Gaussian dependence, whereas the clinical variables are mixed discrete--continuous; ordinal and categorical variables were integer-encoded and treated as continuous for the test. Fisher-$z$ is a pragmatic choice because it is fast, standard, and behaves approximately correctly for the multi-level ordinal encodings that dominate these cohorts. It is nonetheless a simplification in the choice of test family, distinct from the native-NaN discrete-data failure analyzed above, which concerns Type-I control under missingness. We therefore checked robustness to this choice. Re-running the full per-cohort discovery with a nonparametric kernel conditional-independence test (KCI) suited to mixed discrete--continuous data yields identical Markov blankets (HAPI length of stay 21, PPIS 34, and PPIS binary-outcome blankets 50/6/50). The three pooled-shared LoS edges are recovered under KCI, and Age$\rightarrow$LoS and Hemoglobin$\rightarrow$LoS additionally under a nonparanormal (rank-to-normal) Fisher-$z$. The blankets are also stable to MICE/missForest graph-level imputation (Supplementary Table~\ref{tab:sl_sensitivity}). Only the Albumin$\rightarrow$LoS edge is sensitive (it drops under the rank-based transform), consistent with albumin's high missingness. A dedicated mixed-variable discovery method \cite{Li2024MixedVariable} remains a natural extension, but the kernel-test agreement indicates the recovered structure is not an artifact of the Fisher-$z$ Gaussianity assumption. The convergence threshold on Markov-blanket symmetric differences was $\tau = 0.001$. The pooled-shared step uses the same FCI implementation restricted to $V_{\mathrm{shared}}$ on the concatenated table. Temporal-tier constraints were encoded as JSON files listing forbidden directed edges between admission-time, during-stay, and end-of-stay variable tiers. No feature cap was applied at the per-cohort FCI step.

\subsection*{Cohort filtering and HAPI subscore handling}
The PPIS cohort was filtered to drop columns with more than 40\% missing values (a stricter cut than the general 80\%-missingness screen described under Preprocessing) and excluded date stamps, identifiers, and post-event measurements (Age at PI, PI time since admission, Exit date) that constitute outcome leakage. The HAPI cohort was filtered more leniently. Only the per-subscale subscores of the Braden, SCIPUS, and Waterlow scales (the B-, S-, and W-prefixed columns) were dropped, because the composite scales are themselves available. All other clinical variables were retained, including post-PI time-since-last variables, pressure-injury history location indicators, autonomic dysreflexia, and abnormal body-temperature severity. This produced 43 HAPI features and 87 PPIS features for the per-cohort FCI step (excluding the LoS target); definitions of the analysed variables for both cohorts are listed in Supplementary Tables~\ref{tab:ppis_complete} and~\ref{tab:hapi_variables}.

\subsection*{Synthetic and benchmark experiments}
Synthetic experiments were designed to validate the framework against known ground truth. Random directed acyclic graphs were generated with 6, 12, 18, 24, or 30 variables and sample sizes of 200, 400, 600, 800, or 1000. Three missingness mechanisms were considered: missing completely at random (MCAR), missing at random (MAR), and missing not at random (MNAR). Missingness ratios ranged from 10\% to 90\%. 
\paragraph{Baseline definition.}
The non-fusion baseline follows the same iterative imputation and causal discovery procedure but omits the pooled-shared constraint stage. Causal structure is learned only per cohort with temporal-tier constraints but without injecting cross-cohort evidence. This isolates the contribution of the pooled-shared fusion mechanism while keeping all other components identical.

\paragraph{Missingness generation.}
Missing data were introduced synthetically under three mechanisms:
\begin{itemize}
    \item \textbf{MCAR:} entries were removed uniformly at random.
    \item \textbf{MAR:} missingness probabilities depended on observed variables via logistic models.
    \item \textbf{MNAR:} missingness depended on the value of the variable itself, introducing non-ignorable bias.
\end{itemize}
Missingness ratios from 10\% to 90\% were applied independently to each variable.

\paragraph{Experimental protocol.}
All synthetic experiments were repeated over 20 independently generated datasets for each configuration. Reported performance metrics (F1-score, RMSE, and KS distance) correspond to averages across runs.

\subsection*{Clinical datasets}
\label{sec:clinical_datasets}

\paragraph{Study design.} This is a retrospective observational cohort analysis of two existing clinical data resources, conducted at a single specialized spinal cord injury center (Swiss Paraplegic Center, Nottwil). No prospective recruitment, randomization, or intervention took place. The unit of observation is the hospital stay (for PPIS, the surgical episode), and follow-up is confined to that stay. All outcomes, including length of stay, are ascertained at or before discharge. No post-discharge follow-up was available, so loss to follow-up does not arise.

The clinical part of this study integrates two pressure-injury-related data resources curated in the spinal cord injury (SCI) rehabilitation setting: a post-pressure-injury surgery (PPIS) dataset and a hospital-acquired pressure injury (HAPI) dataset. These sources differ in cohort definition, clinical stage, and variable availability, which motivated a graph-level integration strategy instead of direct row-wise pooling.

\paragraph{PPIS dataset.}
The PPIS dataset comprises a single post-surgical cohort of $266$ patient episodes with $126$ recorded variables. It represents patients undergoing management after clinically established pressure injury and therefore captures a later stage of the pressure-injury pathway than inpatient surveillance cohorts. Variables span pressure-injury characteristics, demographics, SCI descriptors, functional assessments, laboratory parameters, comorbidities, and clinical scores. The admission dates in the available extract span approximately 2016--2022, consistent with a longitudinal clinical registry structure. For the raw-data conditional-independence experiment only, PPIS was stratified by admission year into three blocks (2015--2018, 2019, 2020 onwards); all other analyses treat PPIS as a single cohort.

\paragraph{HAPI dataset.}
The HAPI resource is composed of three related cohorts with a combined sample size of $250$ records. Two cohorts represent rehabilitation patients with pressure-injury-related information and contain $94$ and $48$ records, respectively; both share the same $65$-variable structure. A third cohort contains $108$ records and serves as a non-HAPI comparison cohort with $36$ recorded variables. In contrast to PPIS, the HAPI resource is primarily informative for upstream risk stratification and inpatient or rehabilitation acquisition mechanisms.

Variables include neurological severity, complications, surgery history, laboratory measurements, functional and nutritional status, and pressure-injury indicators. The richer rehabilitation cohorts additionally contain current pressure-injury descriptors such as stage, location, and timing-related fields.

\paragraph{Clinical complementarity of PPIS and HAPI.}
PPIS and HAPI reflect related but non-identical parts of the clinical trajectory. HAPI is more informative for mechanisms of pressure-injury occurrence and inpatient risk, whereas PPIS is more informative for severity, intervention, and recovery after advanced wound progression. Because of this distinction, the datasets were modeled as heterogeneous but complementary sources contributing partially overlapping structural information.

\paragraph{Variable harmonization and overlap structure.}
Variable alignment was performed through codebook-guided harmonization and semantic matching of clinically corresponding fields across datasets (Table~\ref{tab:clinical_datasets}). The rehabilitation HAPI cohorts had identical variable structure, whereas the non-HAPI comparison cohort contained a smaller subset of overlapping variables. In addition, a dedicated matching sheet identified a broader set of clinically corresponding concepts across the HAPI and PPIS resources. After restricting to fields observed in both cohorts, nine harmonized concepts form the shared backbone and all nine enter the pooled-shared PAG: eight predictors (AIS, level of injury, age, sex, albumin, hemoglobin, vitamin D, pneumonia) together with the target length of stay (Supplementary Table~\ref{tab:shared_backbone}). SCIM is harmonizable across the two resources but is recorded at discharge in PPIS, so it is an end-of-stay measurement and is not a candidate for the length-of-stay backbone. These shared constructs formed the basis for cross-dataset graph fusion, while source-specific variables were retained where clinically informative. Cohort characteristics and per-variable missingness are summarized in Supplementary Tables~\ref{tab:table1} and~\ref{tab:missingness}.

\begin{table}[ht]
\centering
\caption{Overview of the clinical datasets used in the fusion analysis.}
\label{tab:clinical_datasets}
\small
\begin{tabularx}{\linewidth}{>{\raggedright\arraybackslash}p{1.4cm} >{\raggedright\arraybackslash}p{2.5cm} >{\raggedright\arraybackslash}p{2.6cm} >{\centering\arraybackslash}p{0.9cm} >{\raggedright\arraybackslash}p{3.2cm} X}
\toprule
Dataset & Clinical setting & Cohort composition & $n$ & Representative variable domains & Main analytic role \\
\midrule
HAPI & Hospital-acquired pressure injury in inpatient/rehabilitation care & Three related cohorts, including HAPI and non-HAPI patients, with partial variable overlap & 250 & Demographics, neurological severity, complications, laboratory variables, functional status, nutrition, and pressure-injury-related indicators & Modeling upstream pressure-injury risk and acquisition mechanisms across rehabilitation cohorts. \\
PPIS & Post-pressure-injury surgery & Single post-surgical cohort with broader longitudinal clinical detail & 266 & Wound characteristics, demographics, SCI descriptors, perioperative or treatment variables, laboratory variables, comorbidities, surgeries, functional measures, and outcomes & Modeling downstream severity, treatment, complications, and recovery after established pressure injury. \\
\bottomrule
\end{tabularx}
\end{table}

\paragraph{Preprocessing.}
Prior to analysis, variables were screened for extreme missingness, and those with more than 80\% missing values were excluded. Continuous variables were standardized, and categorical variables were encoded using ordinal or one-hot encoding depending on their clinical interpretation. 

Outliers in laboratory measurements were handled through winsorization at the 1st and 99th percentiles. No additional feature engineering was performed to preserve interpretability.

\subsection*{Ethical approval}
This study was conducted according to the guidelines of the Declaration of Helsinki and was approved by the Ethics Committee of Northwest and Central Switzerland (EKNZ). The HAPI cohorts were analysed under project EKNZ 2019-02179 (further use of health-related personal data; approved 25 November 2019), with the data used under the patients' general consent for research. The PPIS cohort was analysed under project EKNZ 2014-107/PB\_2016-00975 (amendment 03; approved 18 October 2022); the committee authorised the further use of these data without consent under Art.~34 of the Swiss Human Research Act.

\subsection*{Evaluation metrics}
Three metrics were used in the controlled experiments. Structural recovery was quantified with the F1-score by comparing the learned structure to the known graph. Imputation accuracy was quantified with root mean squared error (RMSE) against the true hidden values. Distribution preservation was quantified with the Kolmogorov--Smirnov distance,
\[
D_{KS}=\max_x |F_1(x)-F_2(x)|,
\]
where $F_1$ and $F_2$ denote the cumulative distribution functions of the true and imputed values, respectively.

For the clinical application, evaluation was based on downstream prediction from Markov-blanket-derived feature sets. Length of stay was log-transformed and assessed with the coefficient of determination,
\[
R^2 = 1-\frac{\sum_i (y_i-\hat{y}_i)^2}{\sum_i (y_i-\bar{y})^2},
\]
together with root mean squared error and mean absolute error. The binary post-surgical outcomes (UTI, OM, secondary PI complication) were evaluated with AUROC, AUPRC, and F1 at the default threshold of $0.5$. All metrics were computed on stacked out-of-fold predictions (the convention sometimes called a global metric) and not as a mean of per-fold metrics. For typical heteroscedastic clinical data the two definitions can differ by a few percentage points on $R^2$; we report the global version throughout so that each metric describes the entire cohort. Bootstrap 95\% confidence intervals were obtained by resampling the out-of-fold predictions 300 times.

\paragraph{Constrained constrained graph set for binary outcomes.}
For the binary outcomes, whose targets sit mid-trajectory (tier $\le$ during-stay), a minimal Markov blanket estimated by FCI is both sparse and liable to include post-target descendants. We therefore define the \emph{constrained FCI Markov blanket} as the total-conditioning Markov blanket of the target (its parents, children, and children's co-parents in the PAG, i.e.\ the set that renders the target conditionally independent of all remaining variables) in the constrained ($\alpha=0.05$) fused PAG, restricted by m-separation to variables in temporal tiers no later than the target's. The figures call this the \emph{constrained graph set}, and we use the two names interchangeably; we avoid \emph{causal feature set}, since its members are not established as causes of the outcome. The tier restriction removes obvious post-target descendants and keeps the feature set causally grounded. The result is a temporally-constrained retrospective feature set, not an established prospective one, because outcome-onset times are unavailable and some retained during-stay predictors (e.g.\ antibiotic and immobilization duration) may still follow outcome onset. A prospective evaluation would fix a per-task index time and reclassify predictors against it.

\paragraph{Downstream prediction setup.}
For the clinical prediction tasks, the candidate model family was \{XGBoost, LightGBM, CatBoost, Random Forest\} for regression and the same four models for classification, with the prediction model selected per Markov-blanket method by fold-mean $R^2$ (regression) or fold-mean AUROC (classification) before metrics were computed on stacked out-of-fold predictions. This best-of-$N$ protocol matches what feature-selection benchmarks typically grant to baseline methods, and we apply it uniformly to feature-selection baselines and to our constrained Markov blankets to avoid model-choice confounds. \textbf{Post-discovery benchmarking and selection.} Graph learning, Markov-blanket extraction, and the per-strategy best-of-$N$ model choice were computed on the full cohort without re-derivation inside each outer cross-validation fold. Median imputation and the data-driven feature rankings (mutual information, Lasso, random forest, RFE) were re-derived inside each training fold on training data only, so the automatic-selection baselines are leak-free and their reported feature counts are per-fold medians. The reported clinical metrics therefore come from post-discovery predictive benchmarking, in which feature and model selection can borrow information from the evaluation folds. They should be read as optimistic relative to a fully nested pipeline that also repeats FCI/Markov-blanket extraction and model selection within each training fold. As a partial check, we also report a nested cross-validation (Supplementary Table~\ref{tab:nestedcv}) that re-derives the constrained graph set by FCI within each training fold and uses a single fixed TabPFN model (no best-of-$N$). It confirms that the episode-level numbers are optimistic. Under nesting, the constrained FCI Markov blanket gives $R^2=0.674$ on PPIS length of stay and AUROC $0.765$ on UTI, and HAPI length of stay is similar at $R^2=0.706$. Nesting the full constrained-fusion feature selection (pooled and temporal-tier constraints) is computationally intractable on the dense training folds (Supplementary Table~\ref{tab:sl_sensitivity}). We apply no multiplicity correction across the (feature-strategy $\times$ steering-mode $\times$ target) grid, so per-strategy best and simplest-optimal selections are descriptive, and some apparent wins will be due to chance. \textbf{Patient grouping.} The 266 PPIS records correspond to 174 unique patients, 58 of whom contribute more than one surgical episode. Episode-level 5-fold cross-validation can place the same patient in both training and validation folds and is optimistic. Patient-grouped cross-validation (GroupKFold / StratifiedGroupKFold on patient ID) lowers performance by $0.04$--$0.12$ for length of stay, UTI and osteomyelitis (Supplementary Table~\ref{tab:groupcv}). We therefore take the patient-grouped estimates as the primary PPIS results and the episode-level estimates as an optimistic upper bound.

\subsection*{PAG-guided attention steering}

Two attention steering modes were implemented at inference time without any retraining. In the effective $y\!\to\!x$ mode, a scalar penalty $\gamma \cdot \mathbf{1}[g \in S]$ is added to the logit of the $Y$-token attending to feature group $g$. The target set $S$ is either the graph ancestors of the outcome (\emph{Target-Boost}) or its Markov blanket (\emph{MB-Boost}), derived from the constrained PAG under the temporal-tier ordering ($\text{tier}(g) < \text{tier}(\text{target})$). The penalty biases which feature tokens the prediction reads from directly. The Shapley panels report Target-Boost; the leak-free per-fold $\gamma$ selection reports MB-Boost (Supplementary Tables~\ref{tab:nested_gamma},~\ref{tab:clinical_mode_grid}). Feature--Feature steering (FF, $x\!\to\!x$ mode) adds $\gamma$ to the logit of feature token $i$ attending to feature token $j$ for all adjacent $(i,j)$ and is included only as a control. It reshapes feature-to-feature routing without touching the $Y$-token read-out and is inert-to-harmful for prediction (Supplementary Table~\ref{tab:clinical_mode_grid}). Both modes use the feature-group tokenization of TabPFN\,v2 (native FPG=2); each feature is pre-duplicated so that it fills its own token group, giving a one-to-one feature-to-group mapping for the penalty. For classification, steering is applied at late layers L9--11; for regression, the $y\!\to\!x$ boost uses all layers. The penalty strength $\gamma$ is fixed at $\gamma=2.0$ for all clinical evaluations, a conservative point in the effective range for the $y\!\to\!x$ boost ($\gamma \in [2,10]$) identified from preliminary synthetic and Sachs experiments, independent of HAPI and PPIS. The fixed setting avoids test-fold leakage and stays close to per-fold-optimal performance. Marginal Shapley values under mean-masking quantify each feature's contribution and assess whether steering redirects attribution toward causal ancestors.

\subsection*{Theoretical analysis of an idealized fusion operator}
We analyze an idealized weighted-plurality edge-fusion operator and give sufficient conditions for stabilization of the iterative loop; full proofs are in the Supplementary Proofs appendix. The results characterize design principles of the operator and do not certify the specific FCI implementation; their assumptions need not all hold in the clinical application (noted below). (i) Consistency: the fusion step inherits statistical consistency from each constituent PAG estimator, with the fused edge converging to its population plurality target as dataset sizes grow. (ii) Robustness: a stability lemma bounds the perturbation in fusion votes needed to flip a decision; under a common-truth assumption, which need not hold across the distinct HAPI and PPIS cohorts, the fusion error decays exponentially in the number of contributing datasets. (iii) Convergence: under standard contraction-mapping conditions, the iterative imputation--structure-learning loop terminates in finite iterations. We do not empirically verify the contraction and common-truth conditions in the clinical setting.

\subsection*{Statistics and reproducibility}
\paragraph{Reporting guideline.} The clinical component of this study is a retrospective analysis of two existing observational cohorts (HAPI, PPIS) assembled at a single Swiss spinal cord injury center. No prospective recruitment, randomization, or intervention was undertaken, and follow-up is limited to the index hospital stay. Reporting follows the STROBE statement for cohort studies~\cite{vonElm2007STROBE}, together with TRIPOD+AI for the prediction-model component~\cite{Collins2024TRIPODAI}. Supplementary Table~\ref{tab:strobe} maps each STROBE item to the section, table, or figure in which it is addressed, and marks items that are not applicable to this design or not reported.

No statistical method was used to predetermine sample size; both cohorts comprise all eligible admissions in the source registries (HAPI $n=250$; PPIS $n=266$ episodes from $174$ unique patients). No participants or data points were excluded other than through the pre-specified variable-level filters described above. The studies were observational and retrospective, so no randomization or blinding to group allocation was performed; allocation to cross-validation folds was random under a fixed seed, and for PPIS was constrained by patient identifier so that no patient appears in both training and evaluation folds. Conditional-independence testing within FCI used the Fisher-$z$ test at a significance level of $\alpha=0.05$ throughout the main analysis, with $\alpha=0.10$ reported as a sensitivity analysis (Supplementary Table~\ref{tab:alpha_trajectory}). Comparisons of the fusion operator against pooling baselines used two-sided paired Wilcoxon signed-rank tests across matched synthetic replicates, with Holm correction across the reported family; exact $p$-values are given in the corresponding Supplementary tables, and values below $10^{-3}$ are reported in bounded form. Steering comparisons used paired bootstrap resampling (2000 resamples) of out-of-fold predictions against a kernel-matched unsteered baseline. All confidence intervals are $95\%$ bootstrap intervals from $300$ resamples of stacked out-of-fold predictions unless stated otherwise. No multiplicity correction was applied across the (feature-strategy $\times$ steering-mode $\times$ target) grid, so per-strategy best selections are descriptive. All random seeds, package versions, and fold assignments are pinned in the released code (see Code availability).

\subsection*{Use of large language models}
Claude Opus 4.8 (Anthropic) was used to assist with language polishing and editing of the manuscript. No large language model was used to generate the study's scientific content, analyses, or conclusions, which were designed, produced, and verified by the authors.

\section*{Data availability}
The HAPI and PPIS datasets analyzed in this study contain potentially identifying clinical information from a single specialized rehabilitation center and are therefore subject to controlled access under the governing ethics approval and Swiss data-protection law. Upon acceptance of the manuscript, a de-identified version of the datasets will be deposited in a controlled-access repository (repository to be confirmed by the authors at acceptance). Access will be granted to qualified researchers for non-commercial research use following (i) submission of a brief research proposal to the corresponding author, (ii) approval by the Swiss Paraplegic Center data-governance board, and (iii) a signed data-use agreement prohibiting re-identification and redistribution; requests are answered within approximately eight weeks. Until deposition, the de-identified data are available from the corresponding author on reasonable request under the same conditions. The synthetic datasets used for the validation experiments carry no such restrictions and are regenerable from the released code (see Code Availability).

\section*{Code availability}
Implementation of the pipeline (synthetic-data generation, imputation, cross-cohort fusion, PAG-guided attention steering, and all evaluation and figure-generation scripts) is available at \url{https://github.com/SCAI-Lab/pi-fusion-steering} under the MIT license. A Conda \texttt{environment.yml} pinning package versions (\texttt{tabpfn}~6.0.6, \texttt{causal-learn}, \texttt{scikit-learn}, and PyTorch~2.5) and scripts to reproduce each figure and table are provided. Upon acceptance, the exact version used for this manuscript will be deposited with a citable archival DOI (Zenodo), and the tagged release and commit hash will be inserted here.

\section*{Acknowledgements}
This study was funded by the Schweizer Paraplegiker Stiftung (SPS) and the ETH Zürich Foundation through the 2021-HS-348 ETH-SPS Digital Transformation in Personalized Healthcare for SCI. The funders played no role in study design, data collection and analysis, interpretation of the data, the decision to publish, or the preparation of the manuscript.

\section*{Author contributions}
Y.L. developed the causal graphical fusion method and the theoretical analysis. D.A.J. and Y.L. conceptualized and designed the attention steering mechanism and the TabPFN attention analysis. Y.L. and D.P.-G. conceived the study design for manuscript adaptation. Y.L., D.A.J. and A.I. implemented the experiments. Y.L. and D.A.J. revised the analysis and restructured the results. A.S.-S. and D.P.-G. provided the clinical cohort data, contributed clinical domain expertise to variable definition and interpretation, and reviewed the manuscript for clinical accuracy. R.R. and D.P.-G. supervised the project and acquired funding. All authors reviewed and approved the manuscript.

\section*{Competing interests}
All authors declare no financial or non-financial competing interests.

\bibliography{references}

\ifSIinline
\newpage
\input{si-body}

\fi

\end{document}

%% file: si-body.tex
\appendix
\setcounter{figure}{0}
\setcounter{table}{0}
\renewcommand{\thefigure}{S\arabic{figure}}
\renewcommand{\thetable}{S\arabic{table}}

\begin{center}
\Large\textbf{Supplementary Information}
\end{center}
\vspace{1em}

\section{Supplementary Proofs}

This section details the formal mathematical setup, assumptions, theorem statements, and proofs corresponding to the theoretical analysis in Methods, ``Theoretical analysis of an idealized fusion operator''. The results analyze an idealized weighted-plurality edge-fusion operator and characterize sufficient conditions under which the imputation--discovery loop stabilizes. They describe design principles of the operator and do not claim that the specific FCI-based clinical implementation satisfies every stated assumption. In particular, the common-truth assumption on shared edges need not hold across the distinct HAPI and PPIS cohorts, and we do not verify the contraction condition empirically.

\subsubsection*{Setup and Definitions}
Let $V=\{1,\dots,p\}$ denote the union of all variables across $K$ source datasets. For each unordered pair $(i, j)$ with $1\le i<j\le p$, let $\mathcal{E}$ denote the finite alphabet of admissible edge types (e.g., $\mathcal{E}=\{\text{none}, \rightarrow, \leftrightarrow, \circ\!\rightarrow, \dots\}$).

We define the following population and empirical targets:
\begin{itemize}
    \item \textbf{Dataset-Specific Estimators:} Let $e_{ij}^{(k),0}\in\mathcal{E}$ denote the population target edge type for dataset $k$, and let $\hat{e}_{ij}^{(k)}\in\mathcal{E}$ be its finite-sample estimate.
    \item \textbf{Fusion Weights:} Let $w_{k}$ represent the normalized fusion weights such that $w_{k}\ge 0$ and $\sum_{k=1}^{K}w_{k}=1$.
    \item \textbf{Empirical Fusion:} The empirical weighted vote is $\hat{W}_{ij}(e):=\sum_{k=1}^{K}w_{k}\mathbf{1}\{\hat{e}_{ij}^{(k)}=e\}$, and the fused edge type is $\hat{e}_{ij}^{F}\in \arg\max_{e\in\mathcal{E}}\hat{W}_{ij}(e)$.
    \item \textbf{Population Fusion:} The population weighted vote is $W_{ij}^{0}(e):=\sum_{k=1}^{K}w_{k}\mathbf{1}\{e_{ij}^{(k),0}=e\}$, and the corresponding population plurality edge is $e_{ij}^{F,0}\in \arg\max_{e\in\mathcal{E}}W_{ij}^{0}(e)$.
    \item \textbf{Graph Targets:} The population fusion target is the graph $G^{F,0}:=\{e_{ij}^{F,0}:1\le i<j\le p\}$, and its estimator is $\hat{G}^{F}:=\{\hat{e}_{ij}^{F}:1\le i<j\le p\}$.
\end{itemize}

\subsubsection*{Consistency Guarantees}

\noindent\textbf{Assumption A1 (Dataset-specific edge consistency).}\label{assump:A1}
For each dataset $k$ of size $N_{k}$ samples and pair $(i, j)$, $\mathbb{P}(\hat{e}_{ij}^{(k)}\ne e_{ij}^{(k),0}) \le \epsilon_{ijk}(N_{k})$, where $\epsilon_{ijk}(N_{k})\rightarrow 0$ as $N_{k}\rightarrow\infty$.

\noindent\textbf{Assumption A2 (Positive plurality margin).}\label{assump:A2}
For each pair $(i, j)$, define the margin $\Delta_{ij}:=W_{ij}^{0}(e_{ij}^{F,0})-\max_{e\ne e_{ij}^{F,0}}W_{ij}^{0}(e)$. We assume $\Delta_{ij}>0$.

\begin{theorem}[Pairwise and graph-level consistency]\label{thm:consistency}
Under Assumptions A1 and A2, $\mathbb{P}(\hat{e}_{ij}^{F}\ne e_{ij}^{F,0}) \le \sum_{k=1}^{K}\epsilon_{ijk}(N_{k})$. Consequently, $\mathbb{P}(\hat{G}^{F}=G^{F,0})\rightarrow 1$ as $\min_{k}N_{k}\rightarrow\infty$.
\end{theorem}

\begin{proof}[Proof of Theorem \ref{thm:consistency}]
Let $A_{ij}:=\bigcap_{k=1}^{K}\{\hat{e}_{ij}^{(k)}=e_{ij}^{(k),0}\}$. On event $A_{ij}$, all dataset-specific estimates are correct, so $\hat{W}_{ij}(e)=W_{ij}^{0}(e)$ for all $e \in \mathcal{E}$. By Assumption A2, the maximizer is unique, ensuring $A_{ij}\subseteq\{\hat{e}_{ij}^{F}=e_{ij}^{F,0}\}$. Taking the complement yields $\{\hat{e}_{ij}^{F}\ne e_{ij}^{F,0}\}\subseteq A_{ij}^{c} = \bigcup_{k=1}^{K}\{\hat{e}_{ij}^{(k)}\ne e_{ij}^{(k),0}\}$. A union bound gives $\mathbb{P}(\hat{e}_{ij}^{F}\ne e_{ij}^{F,0}) \le \sum_{k=1}^{K}\epsilon_{ijk}(N_{k})$. Extending this to the full graph, $\{\hat{G}^{F}\ne G^{F,0}\}=\bigcup_{1\le i<j\le p}\{\hat{e}_{ij}^{F}\ne e_{ij}^{F,0}\}$. Applying the union bound again yields $\mathbb{P}(\hat{G}^{F}\ne G^{F,0}) \le \sum_{1\le i<j\le p}\sum_{k=1}^{K}\epsilon_{ijk}(N_{k})$. For finite $p$, this bound vanishes asymptotically.
\end{proof}

\subsubsection*{Robustness and Finite-Sample Error Bounds}

\begin{lemma}[Stability under bounded perturbations]\label{lem:stability}
If for a given pair $(i,j)$, $\sup_{e\in\mathcal{E}}|\hat{W}_{ij}(e)-W_{ij}^{0}(e)|<\frac{\Delta_{ij}}{2}$, then $\hat{e}_{ij}^{F}=e_{ij}^{F,0}$.
\end{lemma}

\begin{proof}[Proof of Lemma \ref{lem:stability}]
For the population plurality winner $e_{ij}^{F,0}$, $\hat{W}_{ij}(e_{ij}^{F,0})>W_{ij}^{0}(e_{ij}^{F,0})-\frac{\Delta_{ij}}{2}$. For any $e\ne e_{ij}^{F,0}$, $\hat{W}_{ij}(e)<W_{ij}^{0}(e)+\frac{\Delta_{ij}}{2}\le\max_{r\ne e_{ij}^{F,0}}W_{ij}^{0}(r)+\frac{\Delta_{ij}}{2}$. By definition of $\Delta_{ij}$, $W_{ij}^{0}(e_{ij}^{F,0})-\frac{\Delta_{ij}}{2}=\max_{r\ne e_{ij}^{F,0}}W_{ij}^{0}(r)+\frac{\Delta_{ij}}{2}$. Thus, $\hat{W}_{ij}(e_{ij}^{F,0})>\hat{W}_{ij}(e)$ for all $e\ne e_{ij}^{F,0}$, maintaining $e_{ij}^{F,0}$ as the unique maximizer.
\end{proof}

\noindent\textbf{Assumption B1 (Common truth and independent estimators).}\label{assump:B1}
Assume there exists a true edge $e_{ij}^{*} \in \mathcal{E}$ such that $e_{ij}^{(k),0}=e_{ij}^{*}$ for all $k$. Let $p_{k}:=\mathbb{P}(\hat{e}_{ij}^{(k)}=e_{ij}^{*}) > 1/2$, and assume dataset-specific estimators are independent across $k$.

\begin{theorem}[Exponential fusion error bound]\label{thm:hoeffding}
Under Assumption B1, $\mathbb{P}(\hat{e}_{ij}^{F}\ne e_{ij}^{*}) \le \exp\left(-\frac{2\left(\sum_{k=1}^{K}w_{k}\left(p_{k}-\frac{1}{2}\right)\right)^{2}}{\sum_{k=1}^{K}w_{k}^{2}}\right)$.
\end{theorem}

\begin{proof}[Proof of Theorem \ref{thm:hoeffding}]
Let $X_{k}:=\mathbf{1}\{\hat{e}_{ij}^{(k)}=e_{ij}^{*}\}$, with $\mathbb{E}[X_{k}]=p_{k}$. Let $S:=\sum_{k=1}^{K}w_{k}X_{k}$. If $S>1/2$, the true edge wins the plurality. Thus, $\{\hat{e}_{ij}^{F}\ne e_{ij}^{*}\}\subseteq\{S\le 1/2\}$. The expectation is $\mathbb{E}[S]=\sum_{k=1}^{K}w_{k}p_{k}=\frac{1}{2}+\sum_{k=1}^{K}w_{k}\left(p_{k}-\frac{1}{2}\right)$.

We bound $\mathbb{P}\left(S-\mathbb{E}[S]\le-\sum_{k=1}^{K}w_{k}\left(p_{k}-\frac{1}{2}\right)\right)$. Applying Hoeffding's inequality to the independent variables $X_k \in [0,1]$ yields $\mathbb{P}(S-\mathbb{E}[S]\le -t)\le\exp\left(-\frac{2t^{2}}{\sum_{k=1}^{K}w_{k}^{2}}\right)$. Substituting $t=\sum_{k=1}^{K}w_{k}(p_{k}-\frac{1}{2})>0$ completes the proof.
\end{proof}

\subsubsection*{Iterative loop convergence}

\begin{theorem}[Eventual graph stabilization]\label{thm:stabilization}
Let $X^{(t)}$ denote the completed data after imputation round $t$, and $G^{F,(t)}$ the fused graph. Suppose that after some finite burn-in time $t_0$, the Markov blanket sets used in the imputation step stabilize so that subsequent updates are governed by a fixed operator $X^{(t+1)}=T(X^{(t)})$ for $t\ge t_{0}$, where $T$ is a contraction mapping ($\|T(X)-T(Y)\|\le c\|X-Y\|$ for $0<c<1$). Assume causal discovery is locally stable at limit $X^{\infty}$ (decisions do not change within an $\eta$-neighborhood of $X^{\infty}$). Then there exists $T^{*}<\infty$ such that $G^{F,(t)}=G^{F,(\infty)}$ for all $t>T^{*}$.
\end{theorem}

\begin{proof}[Proof of Theorem \ref{thm:stabilization}]
Because $T$ is a contraction for $t\ge t_{0}$, the Banach fixed-point theorem ensures a unique fixed point $X^{\infty}$ such that $\|X^{(t)}-X^{\infty}\|\le c^{t-t_{0}}\|X^{(t_{0})}-X^{\infty}\|$. Since $c \in (0,1)$, $c^{t-t_{0}}\rightarrow 0$, meaning there exists $T^{*}$ such that $\|X^{(t)}-X^{\infty}\|<\eta$ for all $t\ge T^{*}$. Within this neighborhood, local decision stability implies that conditional independence testing results remain fixed. Therefore, individual PAGs and the final fused graph freeze, yielding $G^{F,(t)}=G^{F,(\infty)}$ for all $t>T^{*}$.
\end{proof}

\section{Attention Mechanism Analysis}

\subsection*{Probing TabPFN's internal causal representations}
Before introducing attention steering, we characterize what causal structure TabPFN\,v2
has already learned to encode internally. Recent work likewise finds that causal information
is recoverable from TabPFN's frozen representations~\cite{Swelam2025TabPFNCausal}. Understanding this informs which attention weights
to target and what effect an intervention can be expected to produce. We probe the model
using synthetic structural causal models (SCMs) with known ground-truth graphs, measuring
to what extent internal attention weights reflect the true causal relationships.

\paragraph{Experimental setup.}
We generated 300 SCMs per mechanism type (linear-Gaussian and nonlinear tanh) with
5--12 nodes and random DAG topologies (chain, fork, diamond, sparse and dense
Erd\H{o}s--R\'{e}nyi graphs). Edge weights were drawn from $\mathrm{Uniform}(0.5, 2)$
with random sign; noise was i.i.d.\ $\mathcal{N}(0,1)$. Each SCM was instantiated with
150 training rows and 50 test rows; the target $Y$ was fixed to the last topological node.
TabPFN\,v2 was run with a single estimator, quantile-uniform-coarse
preprocessing, and feature-group size FPG$=2$. The model has 12 transformer layers,
6 attention heads, embedding dimension 192, and head dimension 32. Its
feature self-attention operates on a sequence of $n_\mathrm{groups}+1$ tokens:
one per feature group plus a $Y$ token at the last position. Two complementary probes
were used. The head-averaged AUROC uses post-softmax attention averaged across
all 6 heads as a ranking score for binary causal-edge labels (parent vs.\ non-parent),
computed per SCM and averaged over 300 SCMs. The multi-head probe fits a logistic
regression (5-fold CV, $C=1$) on the full 6-dimensional attention vector per query--key
pair, detecting causal information distributed across heads that no single head reveals
individually.

\begin{figure}[ht]
\centering
\includegraphics[width=\linewidth]{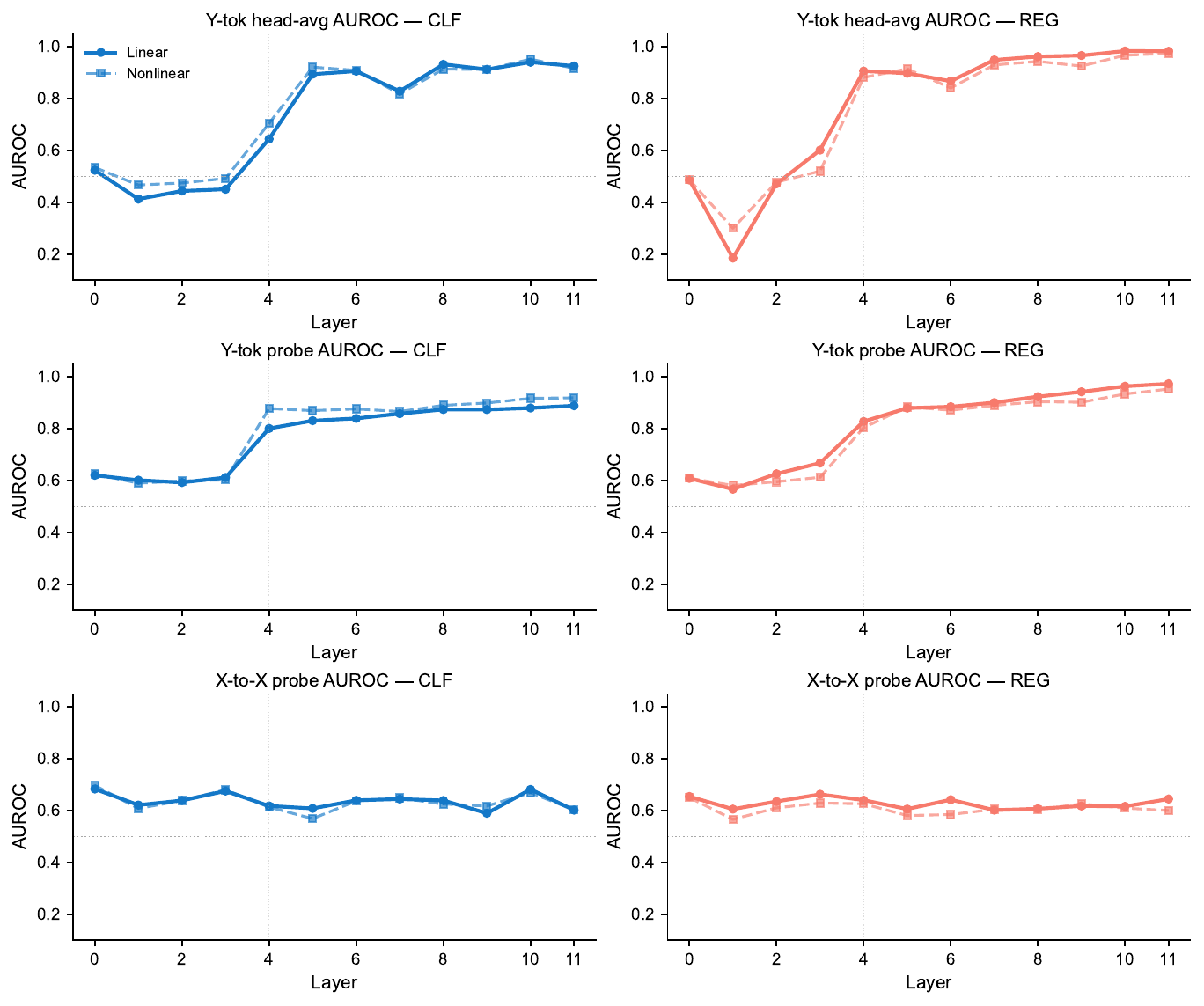}
\caption{Layer-by-layer AUROC profiles for linear-Gaussian (solid) and nonlinear tanh
(dashed) SCMs. Six panels show Y-token head-averaged AUROC (CLF and REG), Y-token
multi-head probe AUROC (CLF and REG), and X-to-X multi-head probe AUROC (CLF and REG).
The phase transition at layers~4--5 and the flat X-to-X probe are mechanism-independent, with
linear and nonlinear SCMs producing the same qualitative structure.}
\label{fig:attn_mechanism}
\end{figure}

\paragraph{A causal phase transition emerges at layer~4.}
The $Y$ token's attention to each feature group is a reliable signal of direct causal
parenthood, but only in later layers. Table~\ref{tab:attn_phase} reports head-averaged
AUROC (parent vs.\ non-parent) by layer for both classification (CLF) and regression (REG)
modes. Layers~0--3 are near chance for CLF ($0.41$--$0.52$) and erratic for REG ($0.19$--$0.60$, including the layer-1 dip). At layer~4, AUROC jumps to $0.645$
(CLF) and $0.906$ (REG), continuing to rise to $0.941$/$0.984$ at layer~10. We refer to
this discontinuity as the causal phase transition. The transition coincides with a
$3.4\times$ (CLF) and $9\times$ (REG) growth in the Frobenius norm of the output-value
(OV) circuit from layer~0 to layer~11, indicating that later layers acquire
more write capacity into the residual stream. REG also shows a
below-chance dip at layer~1 (AUROC$=0.185$), a transient anti-causal intermediate
representation absent in CLF. QK-weight asymmetry does not predict AUROC across layers
($r=0.028$, $p=0.82$ for CLF; $r=-0.153$, $p=0.20$ for REG), ruling out a
directional-weight explanation. Figure~\ref{fig:attn_mechanism} shows that both linear
and nonlinear SCMs produce the same qualitative profile, confirming mechanism-independence.

\begin{table}[ht]
\centering
\caption{Head-averaged AUROC for $Y$-token parent detection by layer (parent
vs.\ non-parent, 300 linear-Gaussian SCMs). \textbf{Bold}: values $\ge0.89$ (strong parent
detection); the shaded late layers lie in the high-detection band used for PAG-guided attention
steering (L9--11 for the classification models; the regression models steer all layers). The phase
transition at layer~4 is visible in both CLF and REG modes; the below-chance
dip at layer~1 for REG reflects a transient anti-causal intermediate representation.}
\label{tab:attn_phase}

\small
\begin{tabular}{lcc}
\toprule
Layer & CLF & REG \\
\midrule
0     & 0.524 & 0.486 \\
1     & 0.412 & 0.185$^{\dagger}$ \\
2--3  & 0.44--0.45 & 0.47--0.60 \\
4     & 0.645 & \textbf{0.906} \\
5     & \textbf{0.894} & \textbf{0.898} \\
8     & \textbf{0.933} & \textbf{0.962} \\
\rowcolor{blue!8}10    & \textbf{0.941} & \textbf{0.984} \\
\rowcolor{blue!8}11    & \textbf{0.926} & \textbf{0.983} \\
\midrule
Mean  & 0.735 & 0.772 \\
Max   & 0.947 (L11, H2) & 0.989 (L10, H1) \\
\bottomrule
\end{tabular}

\end{table}

\noindent$^{\dagger}$Below-chance dip; see text.

\begin{figure}[ht]
\centering
\includegraphics[width=0.8\linewidth]{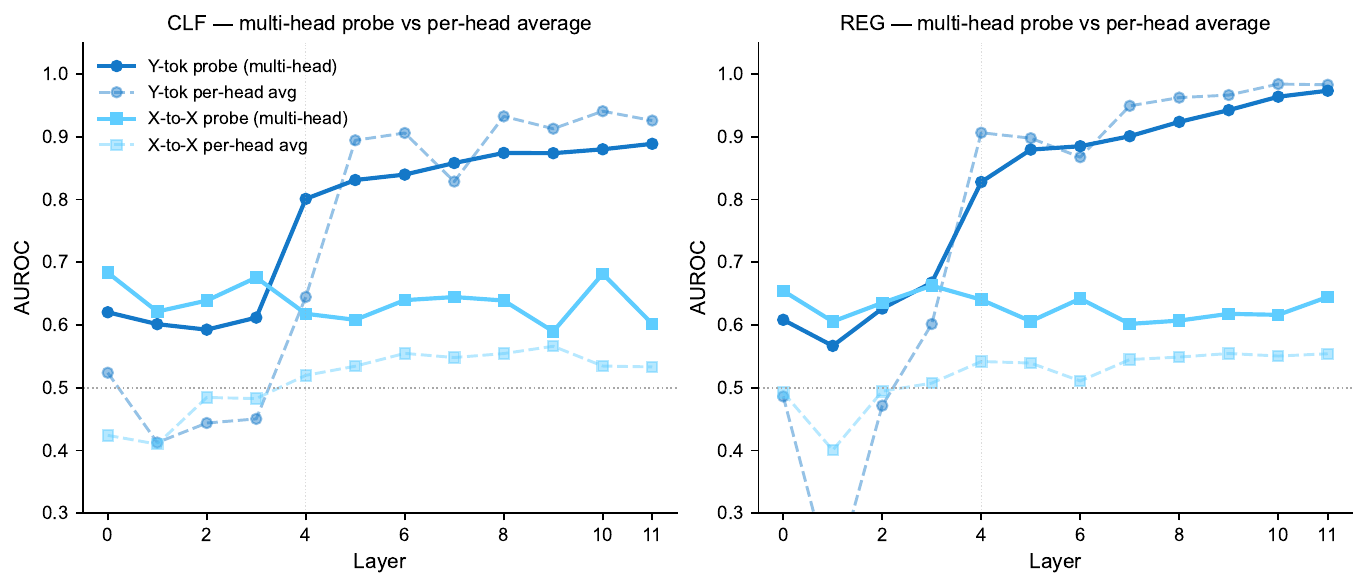}
\caption{Multi-head probe AUROC (solid) versus per-head average AUROC (dashed) by
layer, for the classification (CLF, left) and regression (REG, right) tasks; within each panel the $Y$-token probe is dark blue and the $X$-to-$X$ probe light blue. Linear SCMs. The gap between the lines is the
signature of distributed causal encoding, in which the joint 6-head vector detects causal edges
that no single head detects reliably. Identical patterns hold for nonlinear SCMs
(Supplementary Figure).}
\label{fig:attn_probe}
\end{figure}

\paragraph{Feature--feature causal structure is distributed across attention heads.}
Per-head AUROC for feature$\to$feature (X-to-X) causal edges is near chance in every
individual head (CLF mean $0.512$, max $0.635$ at L2/H3; REG mean $0.520$, max $0.647$
at L0/H4). However, the multi-head probe reveals that the joint 6-head attention
vector encodes substantial causal information from layer~0 onward: probe AUROC is $0.683$
(CLF) and $0.654$ (REG) at layer~0, remaining flat across all 12 layers
($0.60$--$0.68$; Table~\ref{tab:attn_probe}). X-to-X attention therefore carries causal structure,
distributed across the head ensemble in a way
invisible to any individual head. A secondary pattern appears in 4/6 REG heads, where
per-head X-to-X AUROC peaks at layers~0--3 (max $0.647$) and declines
thereafter, suggesting that early layers build feature-level causal representations that
are subsequently aggregated by the $Y$ token at the phase transition.
Figure~\ref{fig:attn_probe} shows the gap between the multi-head probe (solid) and the
per-head average (dashed) as the diagnostic signature of distributed encoding.

\begin{figure}[ht]
\centering
\includegraphics[width=0.49\linewidth]{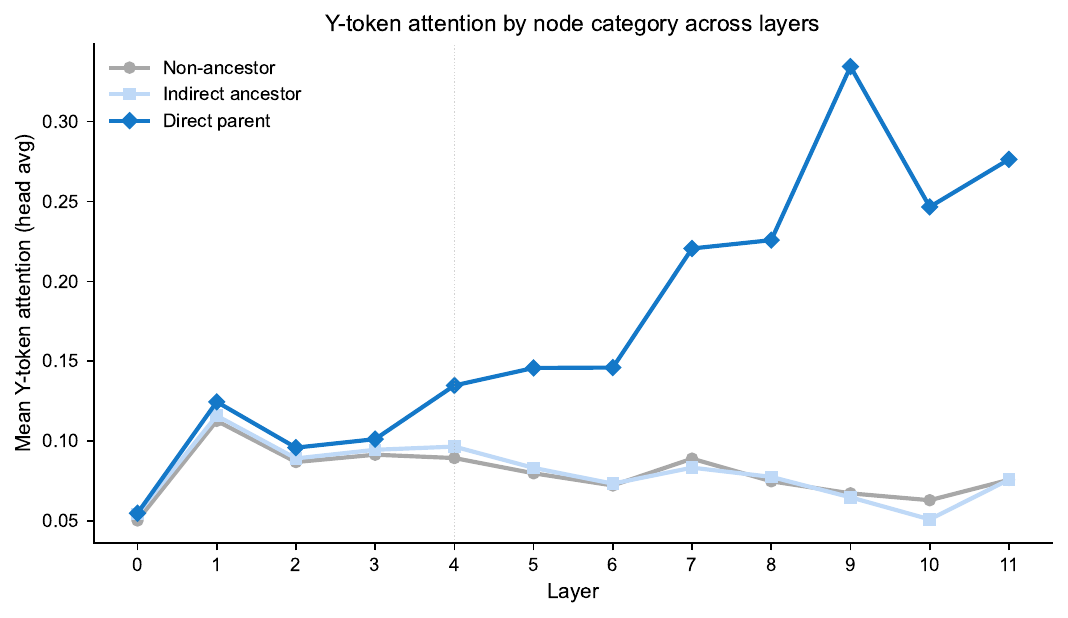}
\includegraphics[width=0.49\linewidth]{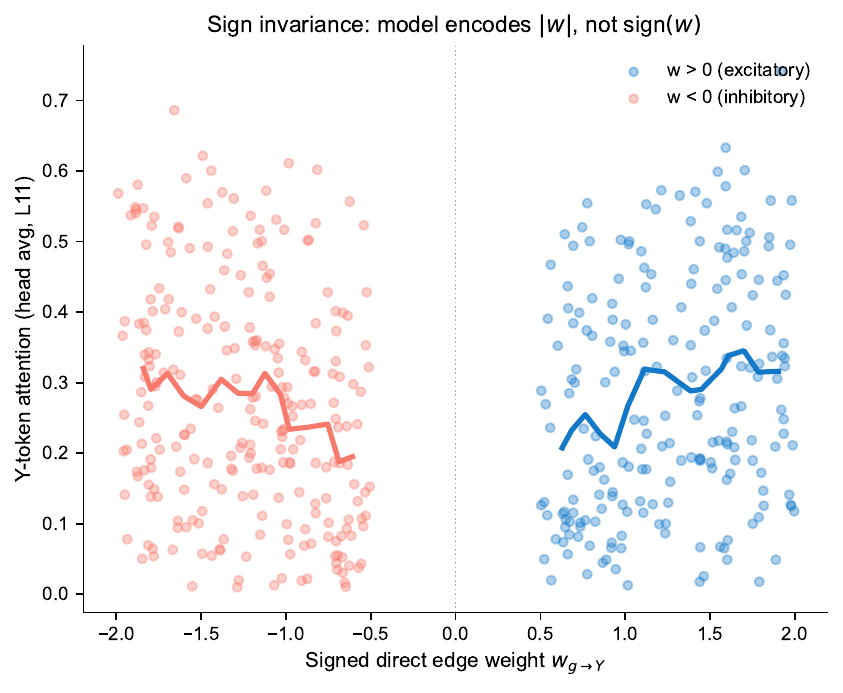}
\caption{Left: mean $Y$-token attention for three node categories (non-ancestor,
indirect ancestor, direct parent) across all 12 layers (linear SCMs). The model
progressively separates direct parents from all other nodes; by layer~8 indirect ancestors
are indistinguishable from non-ancestors. Right: $Y$-token attention versus signed direct
edge weight $w_{g\to Y}$ for direct parents at late layers. The V-shaped scatter confirms
sign invariance. The model encodes $|w|$ (causal magnitude) but not the sign of the
causal effect.}
\label{fig:attn_markov}
\end{figure}

\paragraph{The $Y$ token learns an approximate Markov boundary.}
Beyond detecting parenthood, the $Y$ token progressively sharpens its discrimination
between direct parents and more distal relatives. Figure~\ref{fig:attn_markov} shows
mean $Y$-token attention by node category (non-ancestor, indirect ancestor, direct parent)
across selected layers. At layer~0, all three categories receive nearly identical attention
($0.050$--$0.055$). By layer~8, non-ancestors and indirect ancestors are
indistinguishable ($0.075$--$0.078$), while direct parents receive $0.226$, a $3\times$
excess. By layer~11 this excess reaches $0.276$ vs.\ $0.076$ for both background
categories (Figure~\ref{fig:attn_markov}, left). The model has learned an approximate
Markov property, allocating attention to direct causal parents rather than to variables that
influence $Y$ only through intermediaries. The pattern is consistent with the Markov-blanket
feature-selection property that the fusion pipeline exploits and provides an
internal-representation justification for why TabPFN generalizes well when restricted to
graph-derived Markov blanket feature sets. The correlation between $Y$-token attention
and signed direct edge weight $w_{g\to Y}$ (direct parents only, $n=460$ pairs) confirms
the causal-strength reading. Pearson $r$ rises from $-0.033$ at layer~0 to $+0.284$ at
layer~6 and remains positive through layer~11 ($r=0.231$). Using the
total causal effect instead gives negative $r$ at all layers, because
high-total-effect nodes tend to be distant indirect ancestors that receive low
attention by late layers. The correct quantity for studying Y-token causal-strength
encoding is therefore the direct edge weight.

\paragraph{Attention encodes causal magnitude but not sign.}
A further property is sign invariance. Direct parents with positive ($w>0$) and
negative ($w<0$) edge weights receive statistically identical $Y$-token attention at all
layers (layer~11: $0.282$ vs.\ $0.272$, difference $0.010$). The model therefore encodes
$|w_{g\to Y}|$ (causal magnitude) rather than the direction of the effect.
Figure~\ref{fig:attn_markov} (right) shows the V-shaped scatter of attention against
signed edge weight that is the visual signature of this invariance. This non-trivial
property means the model detects causal relevance irrespective of whether a parent
increases or decreases the target.

\paragraph{Implications for PAG-guided attention steering.}
The four findings above motivate and justify the PAG-guided attention steering results reported in the Results sections. Steering exploits three properties. First, $Y$-token attention carries a causal signal
(AUROC up to $0.99$ at layers~10--11), so steering amplifies a real signal. Second, sign invariance means that all identified causal
parents in the PAG can be boosted uniformly, regardless of whether their effect on the
target is positive or negative. Third, the approximate Markov property means that by
layer~8 indirect ancestors already receive chance-level attention; boosting only direct
parents therefore sharpens an already Markov-correct representation without introducing
confusion from distal ancestral paths. Although causal information is also present in the X-to-X (feature-to-feature)
attention from layer~0 onward, intervening there does not touch the prediction read-out,
which reads only from the final $Y$-token aggregation. Empirically this feature-to-feature
mode is inert-to-harmful for prediction (Tables~\ref{tab:v25_modes} and~\ref{tab:clinical_mode_grid}),
so the effective operator is the one that biases the $Y$-token read-out directly ($y\!\to\!x$).

\clearpage
\section{Steering Validation: Sachs Benchmark and Synthetic Diversity Tests}

\subsection*{Sachs protein-signalling benchmark}

\begin{figure}[ht]
\centering
\includegraphics[width=0.49\linewidth]{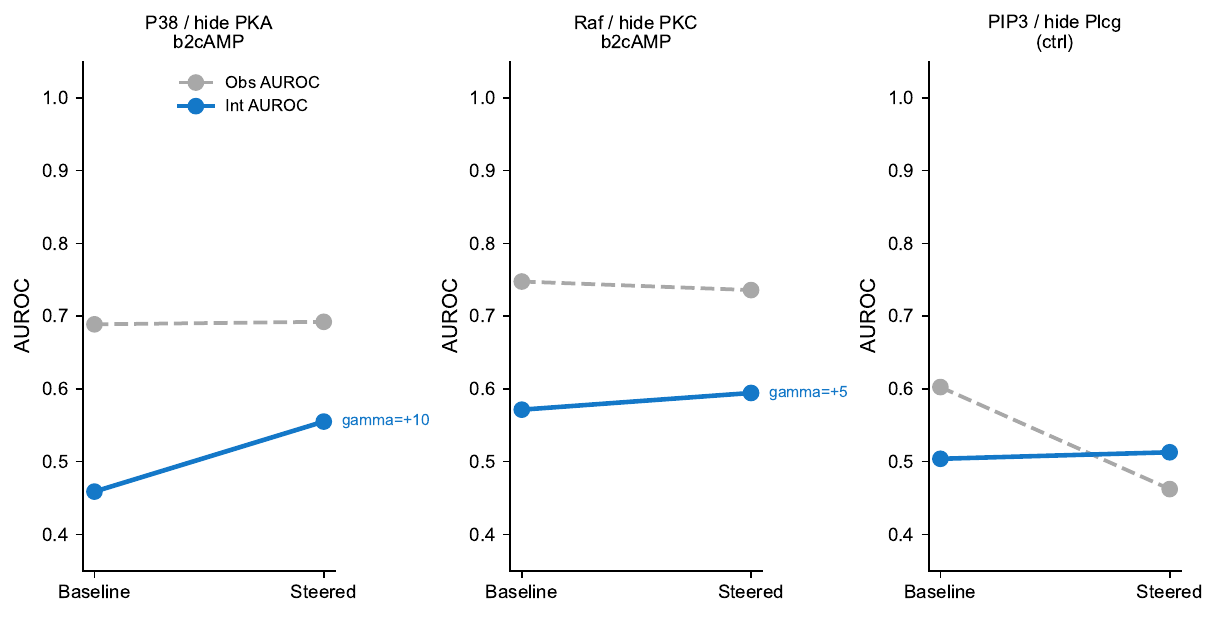}
\includegraphics[width=0.49\linewidth]{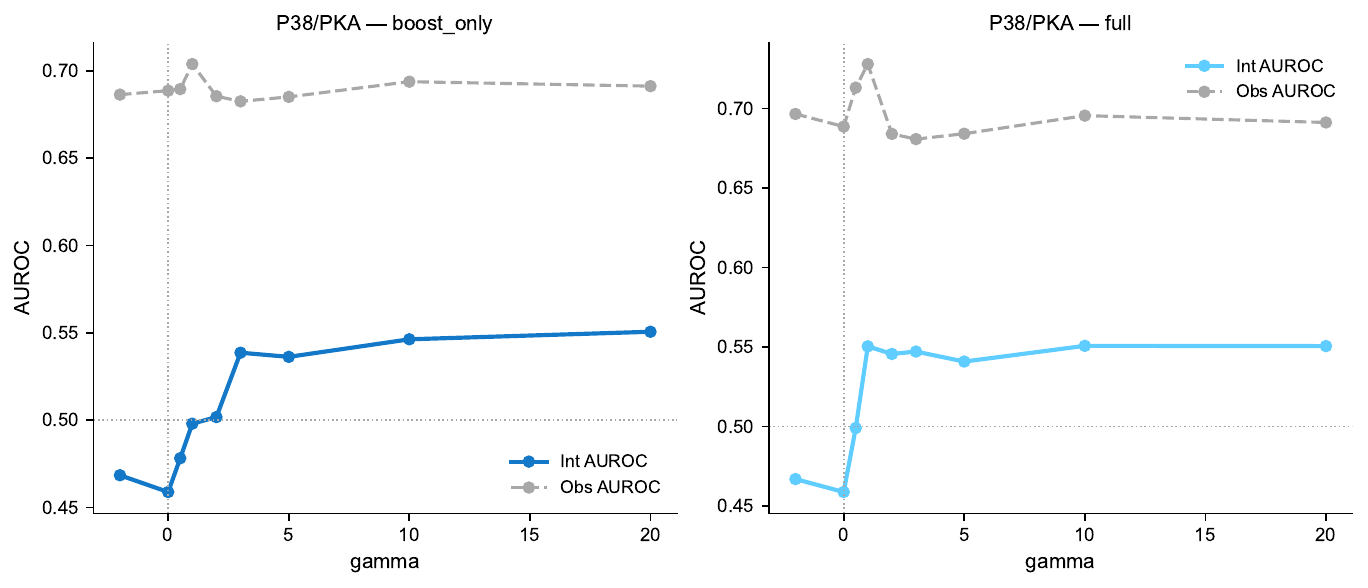}
\caption{Left: baseline vs.\ steered AUROC for the P38/PKA/b2cAMP scenario (observational and interventional). Right: AUROC vs.\ $\gamma$ sweep for boost and FF. Boost peaks at $\gamma=+20$ (AUROC int $0.551$); FF saturates quickly and underperforms boost on this single-ancestor problem.}
\label{fig:sachs_results}
\end{figure}

The Sachs dataset~\cite{Sachs2005} provides a controlled evaluation: 11 proteins measured by flow cytometry with a known ground-truth DAG of 17 directed edges. We use TabPFNClassifier (boost mode) and TabPFNRegressor (FF mode) under the P38/PKA/b2cAMP scenario, where the confounder PKA is excluded from features at test time. Training is on observational data ($n=1{,}800$); test is on the b2cAMP interventional condition ($n=600$). The three scenarios in Table~\ref{tab:sachs_baseline} capture different confounding structures: P38/PKA (single stable visible ancestor PKC), Raf/PKC (single visible ancestor PKA, which shifts under b2cAMP), and PIP3/Plcg (no visible ancestors, negative control). The best per-scenario boost-steering results are summarized in Table~\ref{tab:sachs_boost}, and the baseline-vs-steered AUROC together with the $\gamma$-sweep are shown in Figure~\ref{fig:sachs_results}.

\begin{table}[ht]
\centering
\caption{Sachs baseline AUROC under observational vs.\ interventional distribution. Hidden confounders excluded from features; training on observational data only.}
\label{tab:sachs_baseline}
\small
\begin{tabular}{lllccc}
\toprule
$Y$ & Hidden $C$ & INT & AUROC obs & AUROC int & Drop \\
\midrule
P38  & PKA (direct parent) & b2cAMP & 0.689 & 0.459 & $-$0.230 \\
Raf  & PKC (direct parent) & b2cAMP & 0.748 & 0.571 & $-$0.177 \\
PIP3 & Plcg (only parent)  & b2cAMP & 0.602 & 0.504 & $-$0.098 \\
\bottomrule
\end{tabular}
\end{table}

\begin{table}[ht]
\centering
\caption{Best boost-steering result per Sachs scenario (oracle $\gamma$, layers L9--11). Delta is AUROC int (steered) minus AUROC int (baseline). Obs delta shows cost on the training distribution. \textbf{Bold}: best result. Shaded row: negative control (PAM=0, expected zero gain). Beyond the three baseline scenarios (Table~\ref{tab:sachs_baseline}), Erk/PKC/PMA is reported here as a fourth, shifted-ancestor case.}
\label{tab:sachs_boost}
\small
\begin{tabular}{llccccc}
\toprule
Scenario & PAM type & $\gamma^*$ & $\Delta$int & Int (steered) & $\Delta$obs & Obs \\
\midrule
\textbf{P38/PKA/b2cAMP}    & \textbf{boost, ancestor}     & \textbf{$+20$} & \textbf{$+0.092$} & \textbf{0.551} & \textbf{$+0.002$} & \textbf{0.691} \\
Raf/PKC/b2cAMP    & full-sym, ancestor  & $+5$  & $+0.023$ & 0.594 & $-0.001$ & 0.747 \\
Erk/PKC/PMA       & ancestor            & $-10$ & $+0.018$ & 0.460 & $+0.005$ & 0.703 \\
\rowcolor{gray!15}PIP3/Plcg (ctrl)  & --- (PAM=0)         & ---   & $0.000$  & 0.504 & $0.000$  & 0.602 \\
\bottomrule
\end{tabular}
\end{table}

The P38/PKA scenario achieves the strongest gain ($+0.092$ AUROC) because PKC is the sole visible ancestor and is not a descendant of the hidden confounder PKA (PKC\,$\to$\,PKA in the Sachs DAG), so PKC remains a stable predictor under b2cAMP activation. For the Erk/PKC scenario, the visible ancestors Mek, PKA, and Raf are themselves PKC descendants and shift strongly under PKC activation. The best $\gamma$ is negative (suppressing ancestors), showing that causal ancestor boosting can be counterproductive when ancestors are themselves shifted by the intervention. The negative control (PIP3/Plcg) shows exactly zero gain, as expected, so the PAM mechanism introduces no spurious signal. A random PAM achieves only 14\% of the causal PAM benefit ($+0.013$ vs.\ $+0.092$), with up to $-0.011$ observational harm, confirming that the causal structure information provides approximately $7\times$ benefit over random.

\subsection*{Synthetic diversity tests}

To assess generality beyond the single Sachs scenario, steering was evaluated across 5 DAG topologies (Erd\H{o}s--R\'enyi sparse/dense, scale-free, chain, layered) and 3 mechanisms (linear-Gaussian, nonlinear RBF, mixed) with $n_\text{train}=128$, $n_\text{test}=64$, 3--5 repetitions per configuration, oracle $\gamma$ per repetition (Table~\ref{tab:synthetic_diversity}).

\begin{table}[ht]
\centering
\caption{Synthetic diversity steering under an oracle $\gamma$ (selected per repetition on the interventional test set, which gives a non-deployable upper bound): mean interventional metric improvement and win rate across all mechanisms and topologies. Under this test-set-tuned oracle the aggressive feature--feature (FF) mode scores highest, but the advantage is an artefact of the oracle. At a mechanistically fair fixed strength FF is the least effective mode (Table~\ref{tab:v25_modes}), and on the clinical cohorts FF never helps and is harmful at strength (Table~\ref{tab:clinical_mode_grid}). The deployable recommendation is therefore the direct $y\!\to\!x$ (boost) path; both modes are reported here for completeness.}
\label{tab:synthetic_diversity}
\small
\begin{tabular}{llccc}
\toprule
Task & Condition & $\Delta$int (mean) & $\Delta$obs (mean) & Win rate \\
\midrule
\multirow{2}{*}{CLF (AUROC)} & Boost ($y\!\to\!x$), L9--11 & $+0.010$ & $\approx 0.000$ & 57\% \\
                              & FF ($x\!\to\!x$), L9--11    & $+0.047$ & $-0.013$        & 91\% \\
\midrule
\multirow{2}{*}{REG ($R^2$)} & Boost (y-to-x), L9--11  & $+0.098$ & $\approx 0.000$ & 47\% \\
                              & FF (x-to-x), L9--11     & $+0.675$ & $-0.006$        & 71\% \\
\bottomrule
\end{tabular}
\end{table}

Under the per-repetition oracle, FF scores higher than boost across diverse topologies ($\sim$5$\times$ in classification AUROC delta and $\sim$7$\times$ in regression $R^2$ delta), reflecting that synthetic DAGs have multiple ancestors per target so FF's feature-representation cleaning cascades across many edges at once. This oracle ranking must not be read as a deployment recommendation. It selects $\gamma$ on the interventional test set, and the FF advantage evaporates once the oracle is removed. At a mechanistically fair fixed strength FF is the least effective mode (Table~\ref{tab:v25_modes}), and on the clinical cohorts FF is inert-to-harmful across every outcome (Table~\ref{tab:clinical_mode_grid}). The deployable result (fixed and nested $\gamma$) is carried by the direct $y\!\to\!x$ path, and the oracle-FF numbers establish only that soft steering can engage multi-ancestor structure. Consistent with that, oracle hard feature dropping (zeroing all non-causal columns at test time with full graph knowledge) achieves mean AUROC delta of $-0.105$ at 15\% win rate. Soft attention masking strictly dominates hard masking, preserving all information in the representation while redirecting attention.

\subsection*{Second foundation model (TabPFN\,v2.5)}

The diversity results above select $\gamma$ per repetition on the interventional test set (an oracle upper bound). To test whether steering behavior is tied to one frozen model snapshot and to a tuning oracle, we ran FF steering on a second, independently pretrained foundation model, TabPFN\,v2.5, under a strict deployable protocol with a single fixed $\gamma=2.0$ and no per-repetition tuning. v2.5 uses the boost formulation at late layers L18--23, and v2 uses the symmetric formulation at L9--11. The interventional test set is constructed by a hard $do(\cdot)$ on a non-ancestor of the target, which leaves $P(Y)$ invariant, over 4 DAG topologies (Erd\H{o}s--R\'enyi sparse/dense, scale-free, chain) $\times$ 2 mechanisms (linear-Gaussian, nonlinear RBF) $\times$ 5 repetitions ($n_\text{train}=128$, $n_\text{test}=64$; 40 valid repetitions per model).

\begin{table}[ht]
\centering
\caption{Cross-model steering under a deployable fixed-$\gamma$ protocol. FF steering ($\gamma=2.0$, no per-repetition tuning) versus the unsteered baseline on interventional AUROC, across the same synthetic grid, for two independently pretrained foundation models. Both means are within noise of zero (per-model $\Delta$ standard deviation $\approx 0.025$ over 40 repetitions); per-repetition deltas are positively correlated across the two models ($r=0.35$), indicating the operator engages the same causal structure in both.}
\label{tab:v25_transfer}
\small
\begin{tabular}{lccccc}
\toprule
Model & Layers & Base & FF & $\Delta$AUROC (mean) & Win rate \\
\midrule
TabPFN\,v2   & L9--11 (symmetric) & $0.718$ & $0.711$ & $-0.007$ & 42.5\% \\
TabPFN\,v2.5 & L18--23 (boost)    & $0.725$ & $0.727$ & $+0.002$ & 45.0\% \\
\bottomrule
\end{tabular}
\end{table}

Under this deployable protocol both models yield a mean interventional $\Delta$AUROC indistinguishable from zero ($-0.007$ for V2, $+0.002$ for V2.5; Table~\ref{tab:v25_transfer}), mirroring the null fixed-$\gamma$ FF effect on the clinical cohorts. The absence of a fixed-$\gamma$ gain for this feature--feature mode is a property of that mode under a single fixed strength, not an artefact of one model snapshot. The effective $y\!\to\!x$ mode does produce a fixed-$\gamma$ interventional gain on v2, and this gain vanishes on v2.5 (Table~\ref{tab:v25_modes}). The oracle-$\gamma$ gains in Table~\ref{tab:synthetic_diversity} depend on test-set tuning that is unavailable at deployment. The two models' per-repetition deltas are nonetheless positively correlated ($r=0.35$), confirming that the PAG-guided operator responds to the same causal structure in both models, but with an effect size that is clinically negligible once the tuning oracle is removed.

\subsection*{Leak-free per-fold $\gamma$ selection}

The clinical steering results fix $\gamma=2.0$ (deployable, tuned only on independent synthetic and Sachs data). To confirm that the small clinical effect is not merely an artefact of one un-tuned strength, we additionally selected $\gamma$ per outer fold by an inner 3-fold cross-validation on the training data alone (leak-free), over $\gamma\in\{0,1,2,5,10\}$, and applied the selected value to the held-out fold with the effective $y\!\to\!x$ MB-Boost operator (Table~\ref{tab:nested_gamma}). The results are consistent with the fixed-strength grid. Per-fold selection is conservative. It damps the one fixed-strength gain (PPIS length of stay, from a significant $+0.029$ to a non-significant $+0.009$) and equally damps the small fixed-strength losses on the other targets, so nested selection lands within noise of the unsteered baseline everywhere (PPIS length of stay $+0.009$, osteomyelitis $-0.001$, HAPI length of stay $-0.004$). Per-fold $\gamma$ selection is therefore the safer deployable protocol, close to the unsteered model on every target. It does not change the conclusion that steering delivers no material clinical gain on these cohorts, with PPIS length of stay the one promising but preliminary exception.

\begin{table}[ht]
\centering
\caption{Leak-free nested per-fold $\gamma$ selection for $y\!\to\!x$ MB-Boost steering (the effective mode), applied to TabPFN on the full leakage-safe feature pool and steered toward the Markov blanket, versus a fixed $\gamma=2.0$ and the unsteered baseline. Metric is $R^2$ (length of stay; kernel-matched log-target base) or AUROC (binary outcomes; baselines are the full-pool model, e.g.\ osteomyelitis $0.902$). For each outer fold, $\gamma$ is chosen by an inner 3-fold cross-validation on the training data and applied to the held-out fold; no test-fold information is used. Nested selection is conservative and stays within noise of the baseline on every target, damping both the one fixed-strength gain (PPIS length of stay, $+0.029 \to +0.009$) and the small fixed-strength losses elsewhere. The PPIS length-of-stay row uses patient-grouped outer folds with a kernel-matched $\gamma=0$ baseline ($R^2=0.656$); the HAPI length-of-stay row and the three binary rows are episode-level. Under the episode-level protocol of Figure~\ref{fig:los_results} the same MB-Boost configuration gives $-0.014$ on PPIS length of stay.}
\label{tab:nested_gamma}
\small
\begin{tabular}{llccc}
\toprule
Target & Metric & Baseline & Fixed $\gamma{=}2.0$ ($\Delta$) & Nested $\gamma$ ($\Delta$) \\
\midrule
PPIS length of stay        & $R^2$ & $0.656$ & $0.685$ ($+0.029$) & $0.665$ ($+0.009$) \\
HAPI length of stay        & $R^2$ & $0.761$ & $0.741$ ($-0.020$) & $0.757$ ($-0.004$) \\
UTI during stay            & AUROC & $0.894$ & $0.894$ ($+0.000$) & $0.900$ ($+0.006$) \\
Osteomyelitis              & AUROC & $0.902$ & $0.900$ ($-0.002$) & $0.901$ ($-0.001$) \\
Secondary PI complication  & AUROC & $0.752$ & $0.742$ ($-0.010$) & $0.746$ ($-0.006$) \\
\bottomrule
\end{tabular}
\end{table}

\subsection*{Complete steering-mode \texorpdfstring{$\times$}{x} target grid on the clinical cohorts}

The nested-$\gamma$ table above uses the effective mode ($y\!\to\!x$ MB-Boost). For completeness we swept the entire operator space on all five clinical targets under one kernel-matched protocol (patched
$\gamma{=}0$ baseline, paired bootstrap of $2{,}000$ resamples). Each steering target, the Markov
blanket (MB-Boost) and the graph ancestors (Target-Boost), was crossed with each attention mode
($y\!\to\!x$ direct prediction path, $x\!\to\!y$ transpose, $y\!\to\!x{+}x\!\to\!y$ combined), and
feature--feature ($x\!\to\!x$, FF) steering along the causal tier ordering was added. Table~\ref{tab:clinical_mode_grid}
reports the deployable low-strength ($\gamma{=}2$) delta for every cell. The only significant gain anywhere in the grid is MB-Boost via the
direct $y\!\to\!x$ path on PPIS length of stay ($+0.029$ $R^2$, rising to a $+0.044$ plateau at
$\gamma{=}5$), and every other significant effect is a harm. The grid supports three conclusions. (i)~FF
($x\!\to\!x$) never helps and is significantly damaging at moderate strength, catastrophically so for
the length-of-stay regressions, where it collapses $R^2$ (HAPI $-0.55$, PPIS $-0.79$ at $\gamma{=}5$),
because it perturbs the feature$\leftrightarrow$feature routing block rather than the prediction read-out.
(ii)~Combining modes ($y\!\to\!x{+}x\!\to\!y$) and the transpose alone ($x\!\to\!y$) add nothing over
$y\!\to\!x$ observationally and are the most aggressive at strength (the combined mode reaches
$-0.19$ AUROC on osteomyelitis and $-0.52$ $R^2$ on PPIS length of stay at $\gamma{=}10$). This is consistent
with a no-free-lunch reading in which, absent a distribution shift to exploit, perturbing the feature/cascade
representation only adds variance. (iii)~Ancestor-steering is never better than MB-steering on
these cohorts and is undefined for secondary PI complication, whose leakage-safe pool contains no directed PAG
ancestors of the outcome. The variables marked as ancestors in the Shapley panels are tier-earlier temporal ancestors, a different and larger set. The single effective, deployment-safe operator is therefore $y\!\to\!x$
MB-Boost, the mode used throughout the main text.

\begin{table}[ht]
\centering
\footnotesize
\caption{Complete clinical steering grid: deployable ($\gamma=2.0$) metric delta versus the
kernel-matched ($\gamma{=}0$) baseline for every steering-target $\times$ mode combination, plus
feature--feature (FF) steering. Metric is $R^2$ (length of stay) or AUROC (binary outcomes); paired
bootstrap of $2{,}000$ resamples; $^{*}$ marks a 95\% CI that excludes zero. \textbf{Bold}: the only
significant positive cell in the grid. FF and the combined mode degrade further at higher strength
(see text); a single positive gain ($y\!\to\!x$ MB-Boost on PPIS length of stay) survives to
$+0.044$ at $\gamma{=}5$. ``---'': secondary PI complication has no directed PAG ancestors in the
leakage-safe pool, so ancestor-steering is undefined there.}
\label{tab:clinical_mode_grid}
\begin{tabular}{ll ccc ccc c}
\toprule
 & & \multicolumn{3}{c}{MB-Boost (steer $\to$ MB)} & \multicolumn{3}{c}{Target-Boost (steer $\to$ ancestors)} & FF \\
\cmidrule(lr){3-5}\cmidrule(lr){6-8}\cmidrule(lr){9-9}
Target & Metric & $y\!\to\!x$ & $x\!\to\!y$ & $y{+}x$ & $y\!\to\!x$ & $x\!\to\!y$ & $y{+}x$ & $x\!\to\!x$ \\
\midrule
UTI during stay            & AUROC & $+0.000$ & $+0.001$ & $-0.000$ & $+0.001$ & $+0.001$ & $+0.002$ & $-0.012$ \\
Osteomyelitis              & AUROC & $-0.002$ & $+0.000$ & $-0.002$ & $-0.000$ & $+0.000$ & $+0.000$ & $-0.006$ \\
Secondary PI complication  & AUROC & $-0.010$ & $-0.007$ & $-0.021^{*}$ & --- & --- & --- & $-0.019$ \\
HAPI length of stay        & $R^2$ & $-0.020^{*}$ & $-0.015$ & $-0.032^{*}$ & $-0.012^{*}$ & $-0.019^{*}$ & $-0.041^{*}$ & $-0.091^{*}$ \\
PPIS length of stay        & $R^2$ & $\mathbf{+0.029^{*}}$ & $+0.017$ & $+0.029$ & $-0.017$ & $-0.005$ & $-0.005$ & $-0.237^{*}$ \\
\bottomrule
\end{tabular}
\end{table}

\subsection*{Deployable steering significantly improves interventional accuracy on ground truth}

The clinical deltas above are small because the cohorts are small and, at deployment, offer no explicit distribution shift against which to select $\gamma$. On synthetic ground truth we can both raise the power (144 structural causal models) and test steering in the regime it is designed for, a distribution shift on non-causal features. For each SCM we draw a mixed-ancestor binary target and form an interventional test set by a hard $do(\cdot)$ on a non-ancestor branch, breaking spurious feature--outcome associations while leaving every true ancestor's relation, and $P(Y)$, intact. We select $\gamma$ deployably. On a held-out inner split we construct a \emph{structure-aware shifted} validation set and keep the $\gamma$ that maximizes inner-validation AUROC. A causal method knows which features are non-causal, so it can simulate the shift it must be robust to. No test labels are used; the selected $\gamma$ is then applied to the held-out interventional test (Table~\ref{tab:steer_significance}).

On synthetic ground truth, selecting $\gamma$ per fold with the effective $y\!\to\!x$ operator significantly improves interventional accuracy over the unsteered model. The gain is $+0.029$ AUROC when the structure-aware validation shift matches the deployment shift (an optimistic upper bound; $p=1.5\times10^{-12}$, 69\% win over 144 SCMs) and $+0.011$ when the validation and deployment shifts are decoupled, so that the selector must generalize across shift types (the deployable case; $p=1.4\times10^{-6}$, 44\% win). The selector abstains ($\gamma=0$) in 13\% of repetitions under the matched shift and 40\% under the decoupled shift, correctly declining to steer when it would not help. The gain appears only under two conditions, when $\gamma$ is selected rather than fixed and when there is a distribution shift on non-causal features for steering to exploit. The clinical cohorts satisfy neither well. They are small and offer no explicit shift, and, above all, their deployable feature sets are tier-saturated, meaning that the legitimate predictors of these late-stage outcomes are almost all temporal ancestors and the only non-ancestors are post-outcome measurements excluded for leakage. Because there is nothing for steering to down-weight, the deployable clinical effect stays within noise, and this holds across the entire mode-by-target grid (Table~\ref{tab:clinical_mode_grid}). The only significant clinical gain is $y\!\to\!x$ MB-Boost on PPIS length of stay, and every other operator is null or harmful. The synthetic significance therefore establishes the mechanism and does not demonstrate clinical translation.

\begin{table}[ht]
\centering
\caption{Deployable steering on synthetic ground truth (144 SCMs; interventional test). Per-fold $\gamma$ selection with the effective $y\!\to\!x$ (target-boost) operator on a structure-aware shifted validation set without test labels, versus the unsteered baseline, under two validation regimes. The validation shift either matches the deployment shift (optimistic upper bound) or is decoupled from it (deployable case). Paired Wilcoxon across SCMs.}
\label{tab:steer_significance}
\small
\begin{tabular}{lccc}
\toprule
Validation regime & mean $\Delta$AUROC & win rate & paired $p$ \\
\midrule
Matched shift (upper bound)           & $+0.029$ & 69\% & $1.5\times10^{-12}$ \\
\textbf{Decoupled shift (deployable)} & $\mathbf{+0.011}$ & 44\% & $\mathbf{1.4\times10^{-6}}$ \\
\bottomrule
\end{tabular}
\end{table}

\subsection*{Causal structure reveal curve}

\begin{figure}[ht]
\centering
\includegraphics[width=0.82\linewidth,height=0.9\textheight,keepaspectratio]{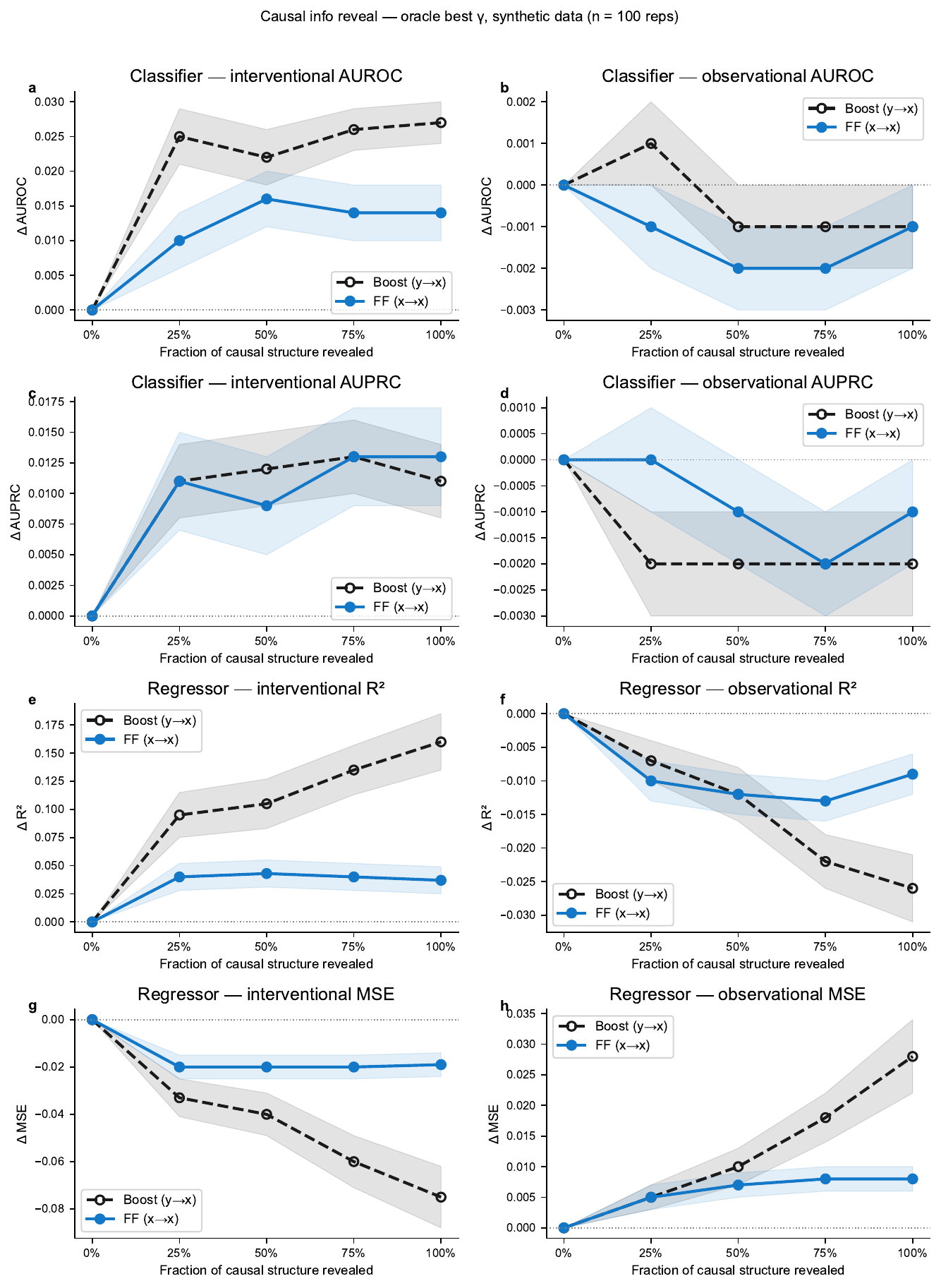}
\caption{Interventional (left columns) and observational (right columns) metric deltas as a function of the fraction of causal structure revealed to the steerer. Rows: classifier AUROC, classifier AUPRC, regressor $R^2$, regressor MSE. Boost (near-black, dashed) scales approximately linearly with revealed structure, reaching $+0.15$ $R^2$ delta at 100\% for regression. FF (blue, solid) saturates at 25\% reveal, recovering most of its gain from a small fraction of the graph. Both methods show a small but growing observational cost as more structure is revealed (right column).}
\label{fig:anc_frac_reveal}
\end{figure}

Figure~\ref{fig:anc_frac_reveal} shows how steering benefit scales with the fraction of causal structure provided to the steerer (0\% = random PAM, 100\% = correct full PAM), averaged over 100 synthetic repetitions.

\clearpage
\subsection*{Model-version dependence: the steering benefit is specific to the TabPFN\,v2 architecture}
\label{si:v25}

All steering results in this work use TabPFN\,v2. Because a newer release (TabPFN\,v2.5) is available, we tested whether the interventional-robustness effect transfers to it. Under the same injection it does not. The interventional-robustness benefit is a property of the v2 architecture as we currently steer it and does not carry over to 2.5, even though 2.5 is a strict drop-in for prediction. The steering procedure itself is model-agnostic, and only this particular injection is v2-specific. We report the non-transfer in full for transparency.

\paragraph{Prediction is unchanged.} Swapping v2 for 2.5 under an identical protocol leaves the prediction benchmark essentially unchanged. Across the five outcomes the causal-Markov-blanket AUROC/$R^2$ moves by at most $\pm0.02$ (e.g.\ PPIS complication $0.776\!\to\!0.785$, PPIS length of stay $R^2\,0.710\!\to\!0.732$ up; osteomyelitis all-feature $0.902\!\to\!0.878$, HAPI length of stay $R^2\,0.745\!\to\!0.736$ down), with no systematic advantage to either version. There is thus no prediction incentive to migrate.

\paragraph{The steering effect vanishes on 2.5.} On synthetic ground truth (72 kernel-matched repetitions, identical protocol to the main text), the effective target-boost mode's interventional gain collapses from $+0.021$/$+0.032$ (at $\gamma=2$/$20$) on v2 to $+0.003$/$-0.006$ on 2.5 (Table~\ref{tab:v25_modes}). The combined mode likewise collapses ($+0.030$/$+0.080 \to +0.003$/$-0.025$) and can harm at high strength.

\begin{table}[ht]
\centering
\caption{Synthetic interventional $\Delta$AUROC by steering mode, TabPFN\,v2 vs.\ 2.5 (kernel-matched baseline, 72 SCM repetitions). $y\!\to\!x$ = target-boost (the deployable mode); $x\!\to\!y$ = transpose; $y\!\to\!x\,{+}\,x\!\to\!y$ = combined; $x\!\to\!x$ = feature--feature. The effect is present on v2 and absent on 2.5.}
\label{tab:v25_modes}
\small
\begin{tabular}{lcccc}
\toprule
Mode & v2 $\gamma{=}2$ & v2 $\gamma{=}20$ & 2.5 $\gamma{=}2$ & 2.5 $\gamma{=}20$ \\
\midrule
$y\!\to\!x$ (target-boost) & $\mathbf{+0.021}$ & $\mathbf{+0.032}$ & $+0.003$ & $-0.006$ \\
$x\!\to\!y$ (transpose)    & $+0.004$ & $+0.032$ & $+0.005$ & $+0.007$ \\
$y\!\to\!x\,{+}\,x\!\to\!y$ (combined) & $+0.030$ & $+0.080$ & $+0.003$ & $-0.025$ \\
$x\!\to\!x$ (feature--feature) & $-0.008$ & $+0.014$ & $-0.001$ & $-0.014$ \\
\bottomrule
\end{tabular}
\end{table}

\paragraph{The non-transfer is not a tuning artefact.} We ruled out the two principled recovery routes. (i)~A layer/formula sweep over seven candidate bands of the 24-layer 2.5 model, with both the ``boost'' and ``symmetric'' penalty formulas, never recovers the effect. The best band ($\{21,22,23\}$) yields only $+0.004$ and is not $\gamma$-ordered. For target-boost the formulas are in any case equivalent, since the penalty is injected directly into the $Y$-token attention row. (ii)~TabPFN\,v2.5 introduces \emph{thinking tokens}, learned scratchpad rows that are prepended in the sample dimension, processed by the between-items attention and absent in v2. Neutralising them does not restore steering (best $+0.008$, still far below v2) and degrades the base model ($0.701\!\to\!0.677$). The prediction read-out, the last feature token, is unchanged between versions, so steering targets the correct token. The loss of effect reflects the deeper 24-layer computation itself rather than a misapplied intervention. On the clinical cohorts, 2.5 steering is correspondingly null-to-negative, and it is numerically unstable for the regression outcomes.

\paragraph{Conclusion.} We therefore report all results on TabPFN\,v2, where the mechanism is well-characterised, and treat the 2.5 non-transfer as a limitation. The steering approach is model-agnostic, but the specific injection we use is tuned to the v2 architecture and is not a guaranteed drop-in for every tabular foundation model. Re-engineering the injection for the 2.5 architecture (e.g.\ intervening on the between-items / thinking-token pathway) is left to future work.

\clearpage
\section{Supplementary Attention Analysis Tables}

\begin{table}[ht]
\centering
\caption{Multi-head logistic regression probe AUROC (5-fold CV) for $Y$-token parent detection and X-to-X causal edge detection at selected layers. The $Y$-token probe rises monotonically (layer 0 → 11); the X-to-X probe is flat, confirming causal structure is present from the first layer but does not concentrate in later layers.}
\label{tab:attn_probe}
\small
\begin{tabular}{lcccc}
\toprule
Layer & $Y$-tok CLF & $Y$-tok REG & X-to-X CLF & X-to-X REG \\
\midrule
0  & 0.620 & 0.608 & 0.683 & 0.654 \\
3  & ---   & ---   & 0.676 & 0.663 \\
6  & ---   & ---   & 0.640 & 0.642 \\
11 & 0.889 & 0.973 & 0.602 & 0.645 \\
\bottomrule
\end{tabular}
\end{table}

\clearpage
\clearpage
\section{Native-NaN conditional-independence test behavior}

\begin{figure}[ht]
\centering
\includegraphics[width=\linewidth]{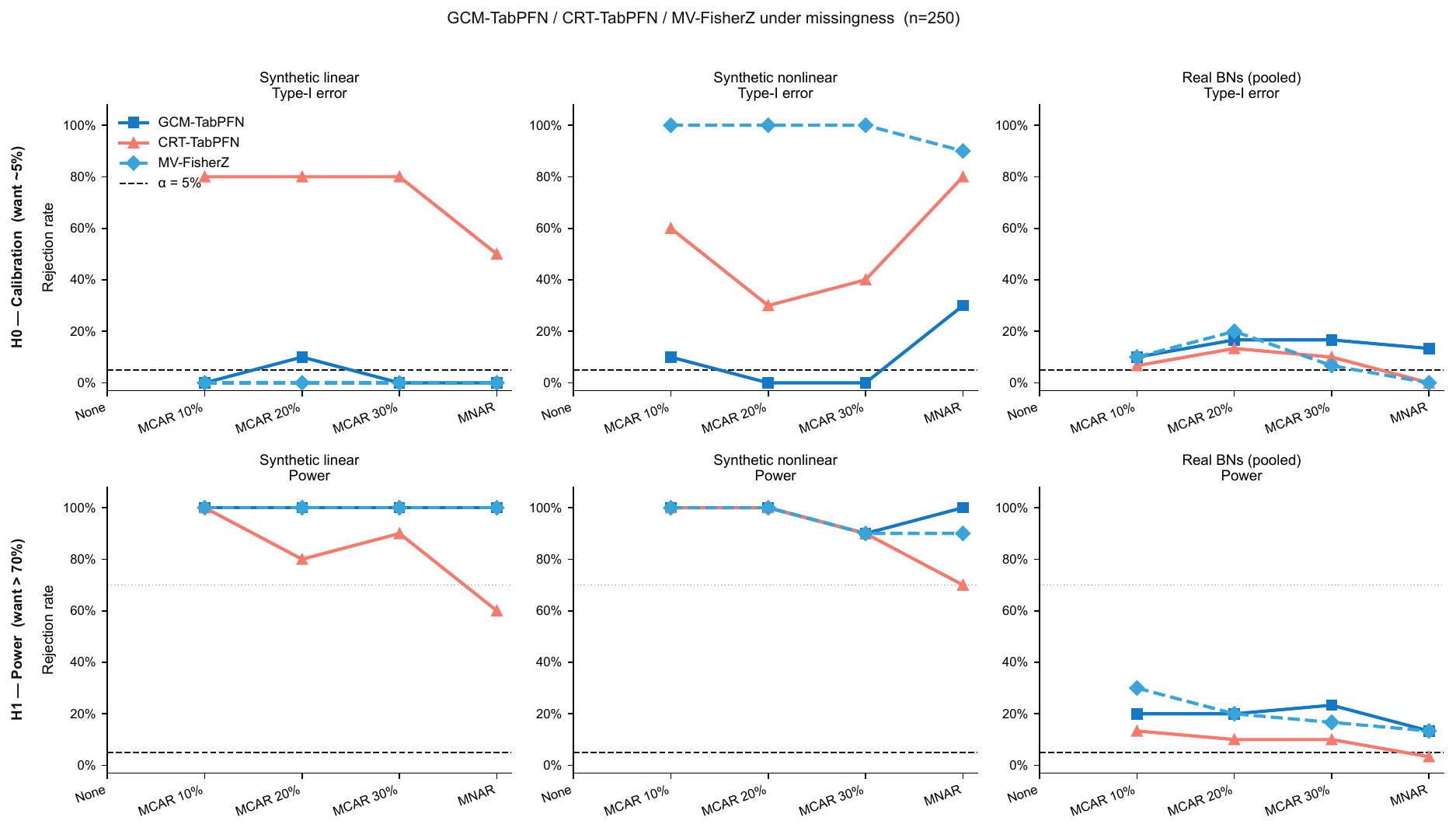}
\caption{Rejection rates of native-NaN conditional-independence tests under missingness ($n=250$, $B=99$). Top row (H0, calibration): Type-I error rate, target $\approx5\%$ (dashed line). Bottom row (H1, power): target $>70\%$ (dotted line). Columns show synthetic linear, synthetic nonlinear, and pooled real Bayesian-network data for GCM-TabPFN, CRT-TabPFN, and MV-FisherZ. On real discrete/BN data all three tests have low power ($\le 30\%$), while CRT-TabPFN and MV-FisherZ additionally inflate Type-I error on nonlinear nulls. This is the primary motivation for imputation before causal discovery; the summary values in Table~\ref{tab:ci_test_summary} are read from this figure.}
\label{fig:negative_finding}
\end{figure}

\clearpage
\section{Complementary Results on Synthetic Data}
\begin{figure}[h]
\centering
\includegraphics[width=0.62\linewidth]{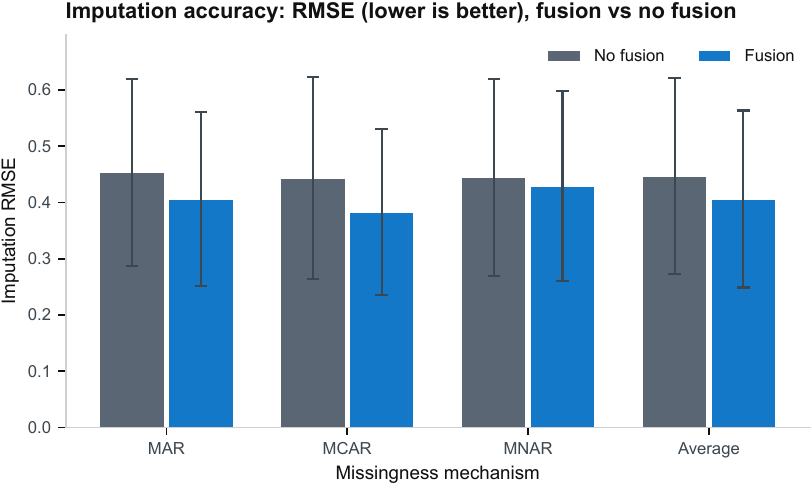}
\caption{\textbf{Imputation accuracy across missingness mechanisms.} Root-mean-squared error of the imputed against the hidden values for the fusion framework (blue) versus the non-fusion baseline (grey), by missingness mechanism and their average; lower is better. Bars are means over the synthetic grid with $\pm 1$ standard deviation whiskers. Fusion yields lower imputation error under every mechanism.}
\label{fig:rmse_panel}
\end{figure}

\subsection*{Cross-cohort structure recovery}
The panel above isolates the missing-data step. Here we test the fusion step by asking whether merging per-cohort PAGs recovers the shared causal skeleton better than the natural alternatives. We generated random 20-node Erd\H{o}s--R\'enyi DAGs (edge probability $0.12$) and split the variables into two partially overlapping cohorts sharing half of them. Each cohort was sampled independently ($n=300$, linear-Gaussian) and subjected to MAR missingness ($70\%$ of variables affected). The estimand is the latent-aware oracle skeleton over the shared variables, namely the FCI-target PAG skeleton computed by exact d-separation with all non-shared variables treated as latent for the shared subgraph. A method is thus not penalized for correctly recovering an adjacency induced by an unobserved common cause. To model cross-cohort heterogeneity we add $k$ cohort-specific edges among the shared variables, leaving the invariant base skeleton as the target. Metrics are precision/recall/F1 over the shared skeleton, averaged over 50 random graphs, with paired Wilcoxon tests (same seeds) and Holm correction across method pairs.

We compare six estimators: \textbf{fusion~$\cap$} (per-cohort FCI on median-imputed data, merged so that a shared-variable edge survives only with support from both cohorts); \textbf{fusion~$\cup$} (the same merge keeping an edge found by either cohort); \textbf{per-cohort} (the mean of the two single-cohort graphs' metrics, a non-deployable reference, since an average of two graphs cannot be deployed); \textbf{pool-on-shared} (a fair pooling baseline running one FCI on the two cohorts row-concatenated over the shared variables only, so no variable is structurally missing); \textbf{naive-pool} (pooling over the union of variables, so non-shared columns are block-missing); and \textbf{no-imputation} (listwise deletion before per-cohort FCI).

\begin{figure}[ht]
\centering
\includegraphics[width=\linewidth]{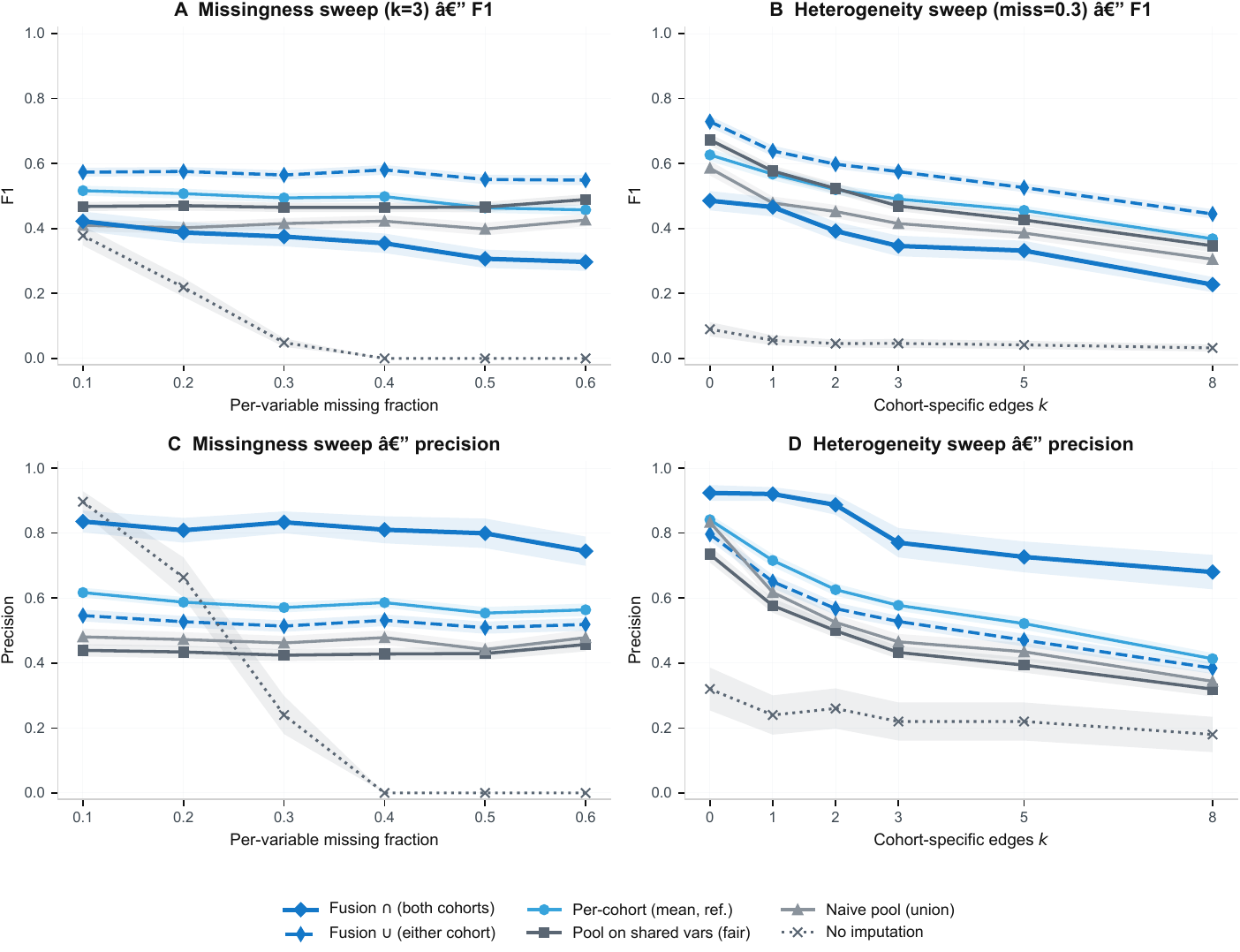}
\caption{\textbf{Synthetic cross-cohort structure recovery.} Recovery of the shared (invariant) skeleton over the overlapping variables, versus per-variable missing fraction (\textbf{A}, \textbf{C}; heterogeneity fixed at $k=3$) and versus cross-cohort heterogeneity $k$ (\textbf{B}, \textbf{D}; missingness fixed at $0.3$). Top row F1, bottom row precision; bands are $\pm1$ standard error over 50 random DAGs. Fusion~$\cap$ (both cohorts) has the highest precision throughout; fusion~$\cup$ (either cohort) has the best F1 of any method, significantly beating both the per-cohort average and the fair pool-on-shared baseline in every missingness and heterogeneity condition (paired Wilcoxon $p<10^{-3}$); the naive union-pool is dominated on both metrics; and listwise deletion (no imputation) collapses at high missingness.}
\label{fig:fusion_benchmark}
\end{figure}

\begin{figure}[ht]
\centering
\includegraphics[width=\linewidth]{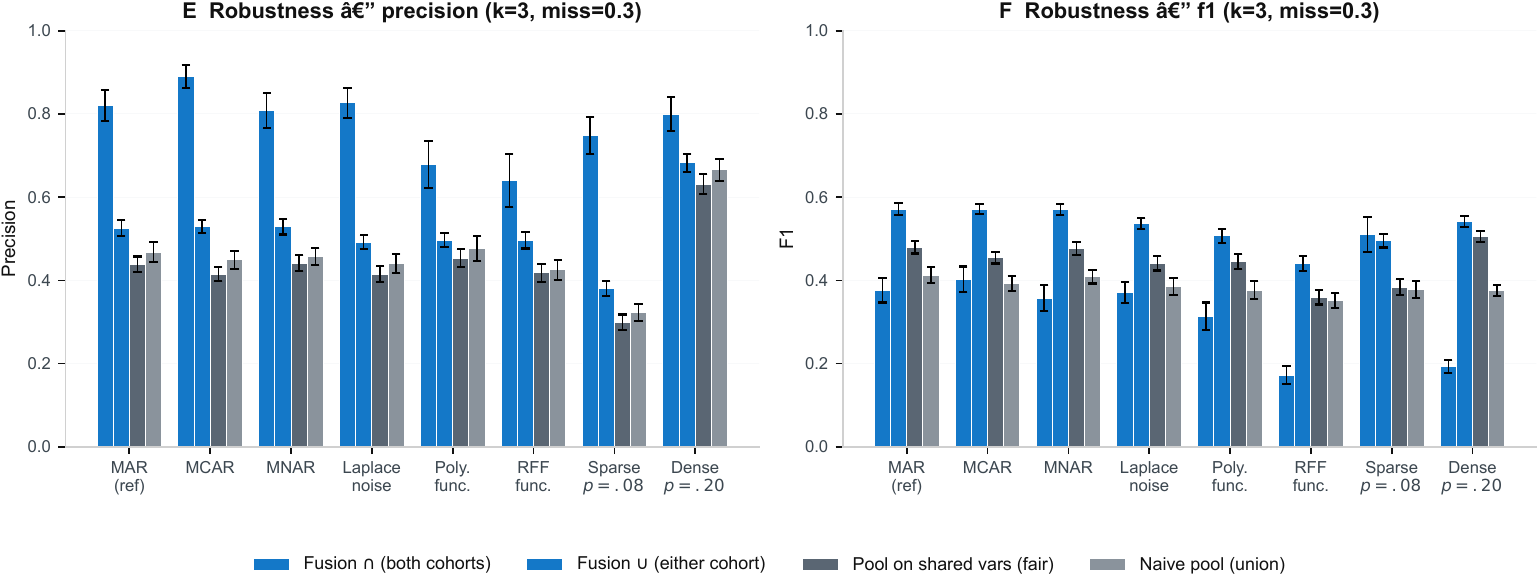}
\caption{\textbf{Robustness of the fusion precision advantage} at the heterogeneous reference setting ($k=3$, missingness $0.3$). Precision (\textbf{E}) and F1 (\textbf{F}) for both fusion operating points versus the fair pool-on-shared and naive union-pool baselines under MCAR, MAR and MNAR missingness, Laplace noise, polynomial and RBF functional forms, and sparse ($p=0.08$) and dense ($p=0.20$) graphs. Bars are means with $\pm1$ standard-error whiskers over 50 random DAGs. Fusion~$\cap$ retains the highest precision under every variant, with the margin narrowing only for the RBF-nonlinear data that violate the Fisher-$z$ test's assumptions.}
\label{fig:fusion_robustness}
\end{figure}

Three findings hold across the settings tested (Figure~\ref{fig:fusion_benchmark}). (i)~Fusion~$\cap$ is by far the highest-precision recovery operator. Requiring an adjacency to be recovered by both cohorts yields precision $\approx0.8$, exceeding the fair pool-on-shared baseline by $+0.36$ to $+0.46$ (median paired difference over 50 seeds, Holm-corrected $p<10^{-30}$ in both the missingness and heterogeneity sweeps) and naive pooling by a similar margin. The advantage persists across MCAR, MAR and MNAR missingness, Gaussian and Laplace noise, linear, polynomial and RBF functional forms, and graph density. It ranges from $+0.17$ (dense graphs) to $+0.47$ (MCAR) and remains large even under the nonlinear data for which the Fisher-$z$ test is misspecified (Figure~\ref{fig:fusion_robustness}). (ii)~The consensus threshold is a strong precision/recall dial. Fusion~$\cap$ (both cohorts) maximizes precision at the cost of recall, whereas fusion~$\cup$ (either cohort) maximizes recall and gives the best F1. For two equally weighted cohorts these are the only two hard operating points, and the practitioner selects along this range according to whether false positives or false negatives are more costly. (iii)~Fusion dominates the precision/recall frontier. The fair pool-on-shared baseline is beaten at both operating points, by fusion~$\cap$ on precision (finding~(i)) and by fusion~$\cup$ on F1. Fusion~$\cup$ attains the best F1 of any estimator, significantly exceeding both the per-cohort average (median $+0.06$ to $+0.09$) and the fair pool-on-shared baseline (median $+0.06$ to $+0.12$; paired Wilcoxon $p<10^{-3}$ in every one of the six missingness and six heterogeneity conditions). The strict intersection alone trades recall for precision (its F1 falls $-0.14$ below pooling), so the operative decision is which fusion operating point to read. The naive union-pool is dominated on both metrics. Simple median imputation is essential, since the no-imputation arm collapses to F1~$\approx0$ once $\ge40\%$ of values are missing. Recovery degrades smoothly as heterogeneity $k$ grows, yet a fusion operating point remains the best-precision and the best-F1 estimator throughout. We therefore read fusion as a cross-cohort operator that dominates both naive and fair pooling across the entire frontier. The intersection is the choice when false positives are costly and the union when false negatives are, and in both cases fusion beats pooling.

\clearpage
\section{Feature-importance (Shapley) analysis}

Marginal Shapley feature-importance results for all five prediction targets are
shown below. Each
panel compares TabPFN baseline, Target-Boost, and FF steering at fixed
$\gamma=2.0$; blue bars denote temporal ancestors of the target and grey bars
non-ancestors. Because each figure overlays all feature-selection strategies for one
target, it is dense. The figures are vector graphics best
viewed zoomed on screen, and the underlying per-feature Shapley values for every
strategy are provided as a machine-readable data file (\texttt{shap\_importance.json})
in the code repository. The mechanistically relevant panel referenced in the
main text is the constrained-neighborhood (``FCI MB'') panel of each figure; for
print, we recommend the editor request the split single-strategy panels, which
regenerate from the provided data.

\begin{figure}[p]
\centering
\includegraphics[width=\linewidth,height=0.93\textheight,keepaspectratio]{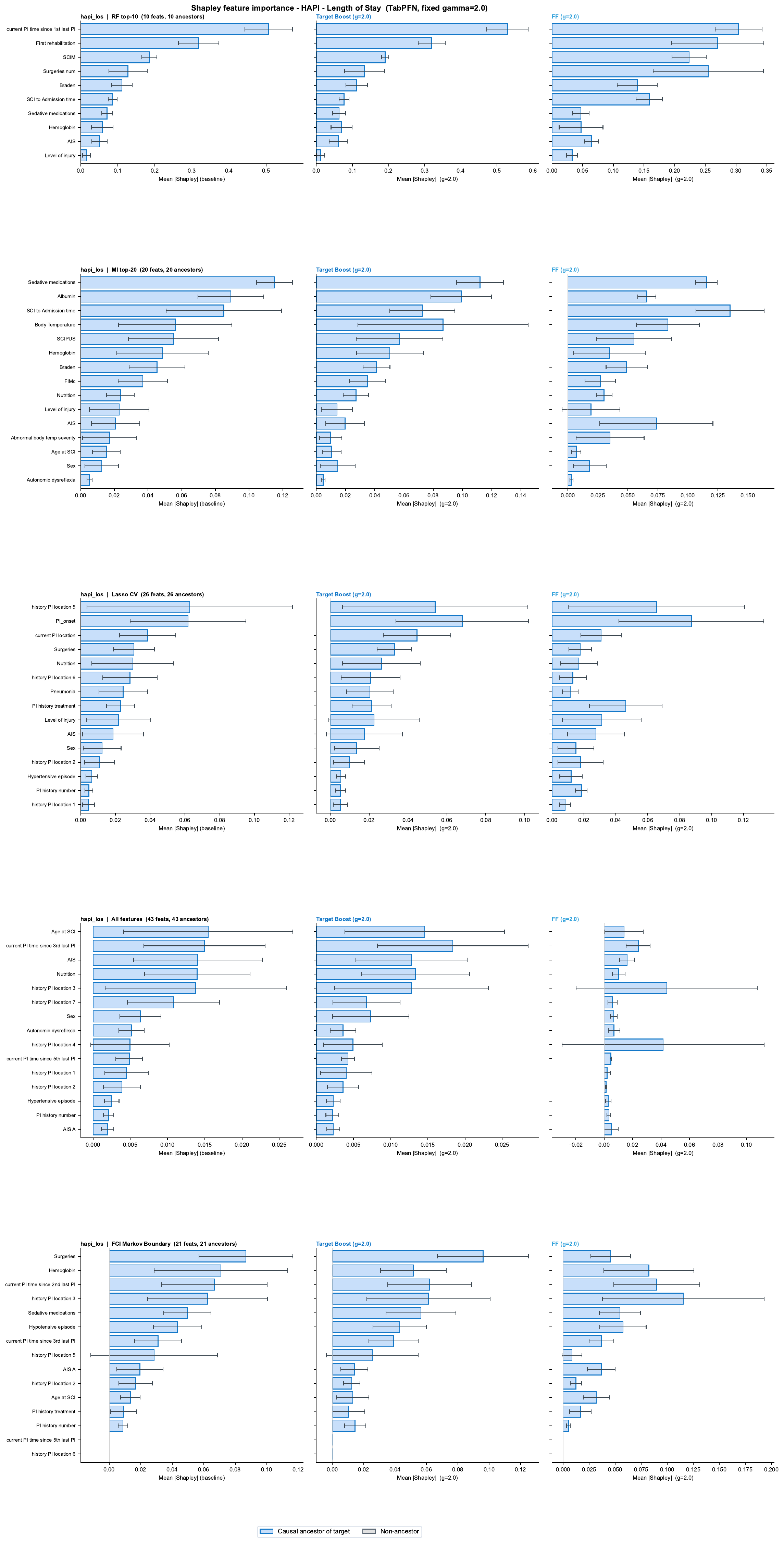}
\caption{Marginal Shapley feature importance for TabPFN (baseline, Target-Boost, FF at fixed $\gamma=2.0$) on HAPI length-of-stay across all feature sets. Blue bars: temporal ancestors of the target; grey: non-ancestors. All HAPI LoS features are temporal ancestors, so steering produces no ancestor/non-ancestor ratio change (tier saturation confirmed).}
\label{fig:importance_hapi_los}
\end{figure}

\begin{figure}[p]
\centering
\includegraphics[width=\linewidth,height=0.93\textheight,keepaspectratio]{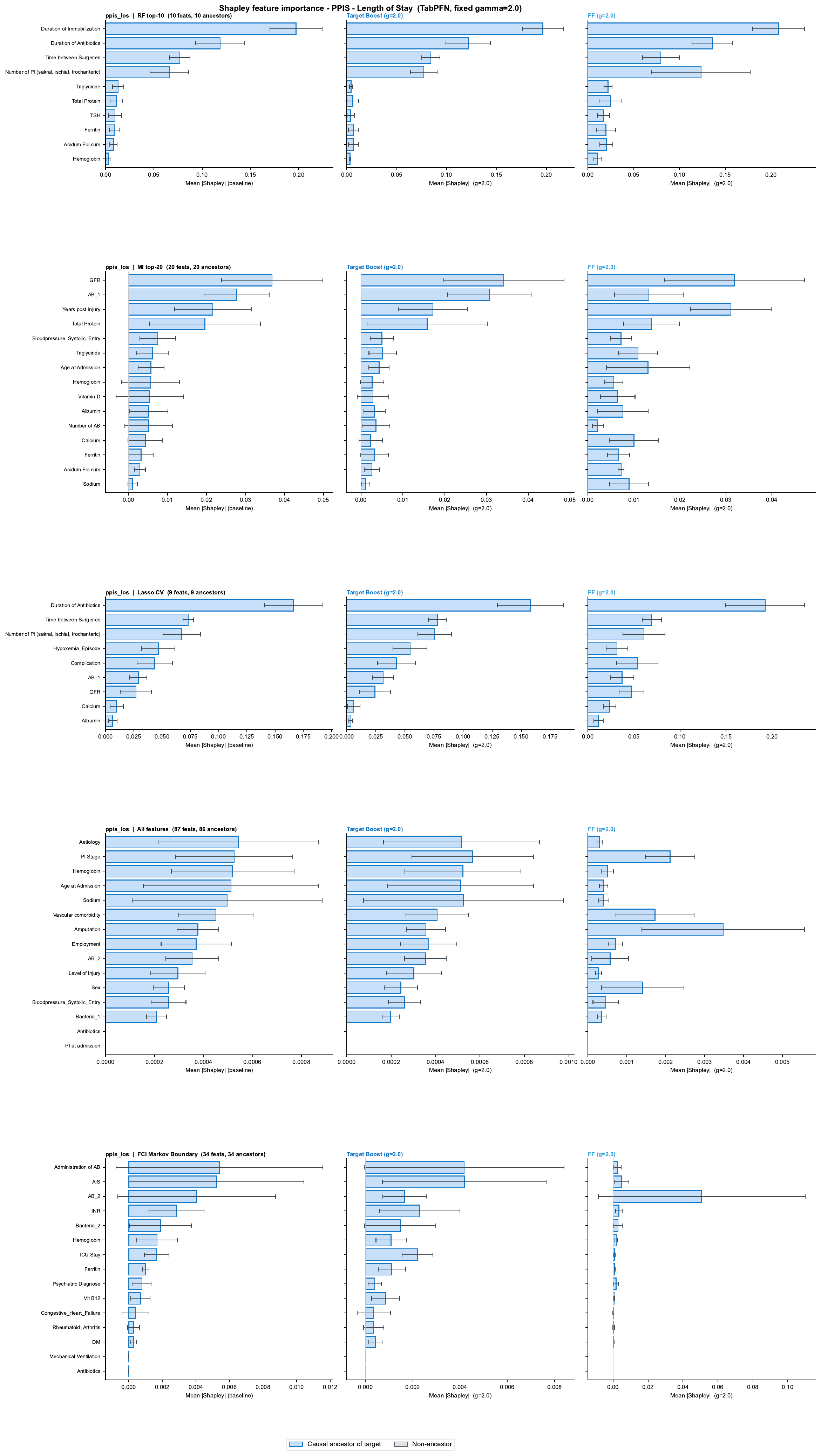}
\caption{Marginal Shapley feature importance for PPIS length-of-stay across all feature sets. Every PPIS length-of-stay feature set is tier-saturated, with all selected features temporal ancestors of the target (RF top-10: 10/10 ancestors; FCI MB: 34/34). The ancestor/non-ancestor ratio is thus undefined and steering produces no ratio change, confirming the softmax-invariance (tier-saturation) argument. FF steering likewise yields no leak-free $R^2$ gain on the RF top-10 set (fixed $\gamma=2.0$: $-0.010$ versus the unsteered model).}
\label{fig:importance}
\end{figure}

\begin{figure}[p]
\centering
\includegraphics[width=\linewidth,height=0.93\textheight,keepaspectratio]{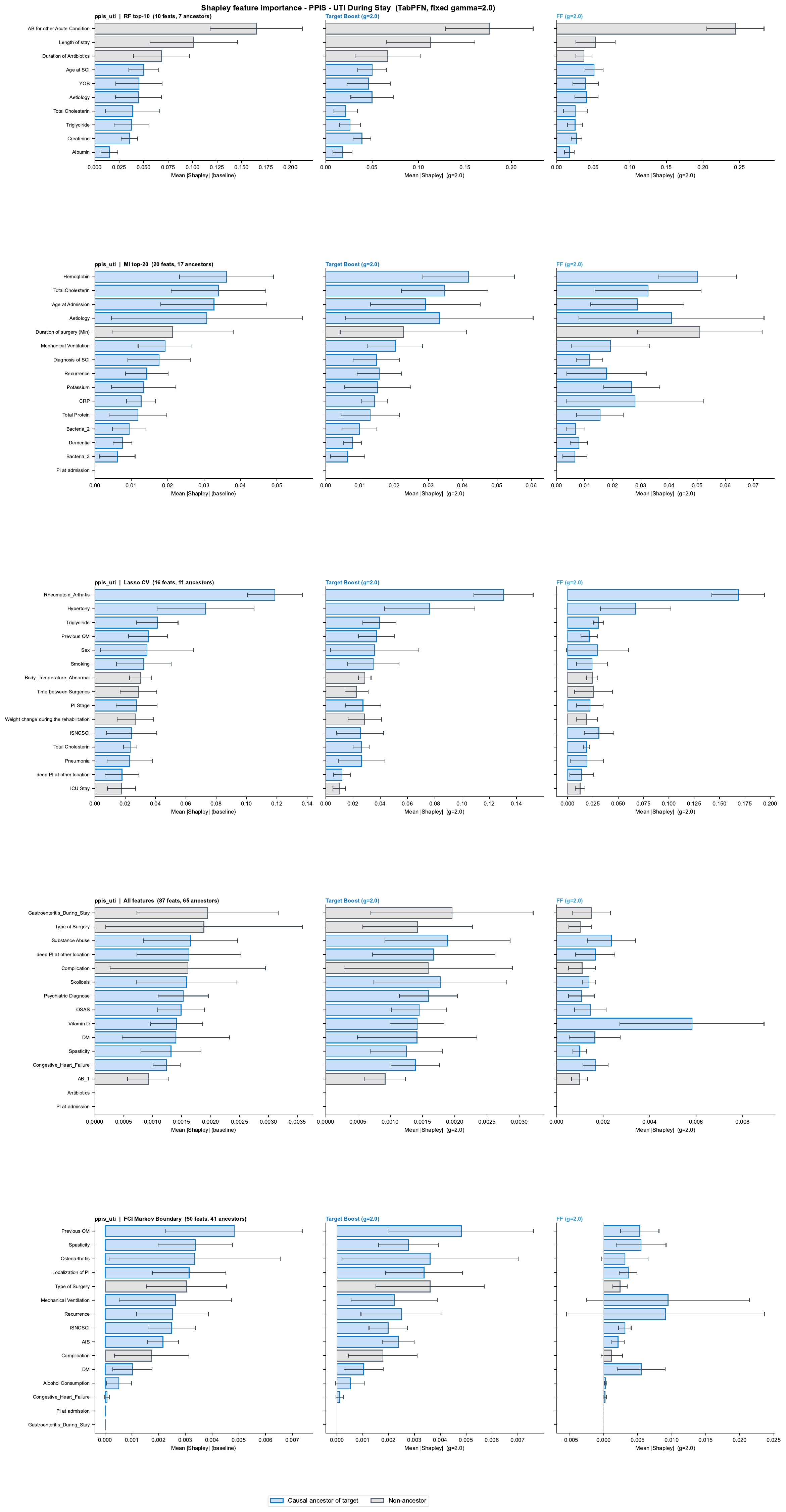}
\caption{Marginal Shapley feature importance for PPIS UTI during stay across all feature sets. On the constrained FCI Markov blanket panel (50 features, 41 ancestors), the ancestor/non-ancestor attribution ratio is $2.40$ at baseline. Target-Boost raises it to $2.84$ ($+18\%$), more than the feature--feature control (FF: $2.72$, $+13\%$), so the effective $y\!\to\!x$ mode concentrates attribution most on the causal ancestors.}
\label{fig:importance_uti}
\end{figure}

\begin{figure}[p]
\centering
\includegraphics[width=\linewidth,height=0.93\textheight,keepaspectratio]{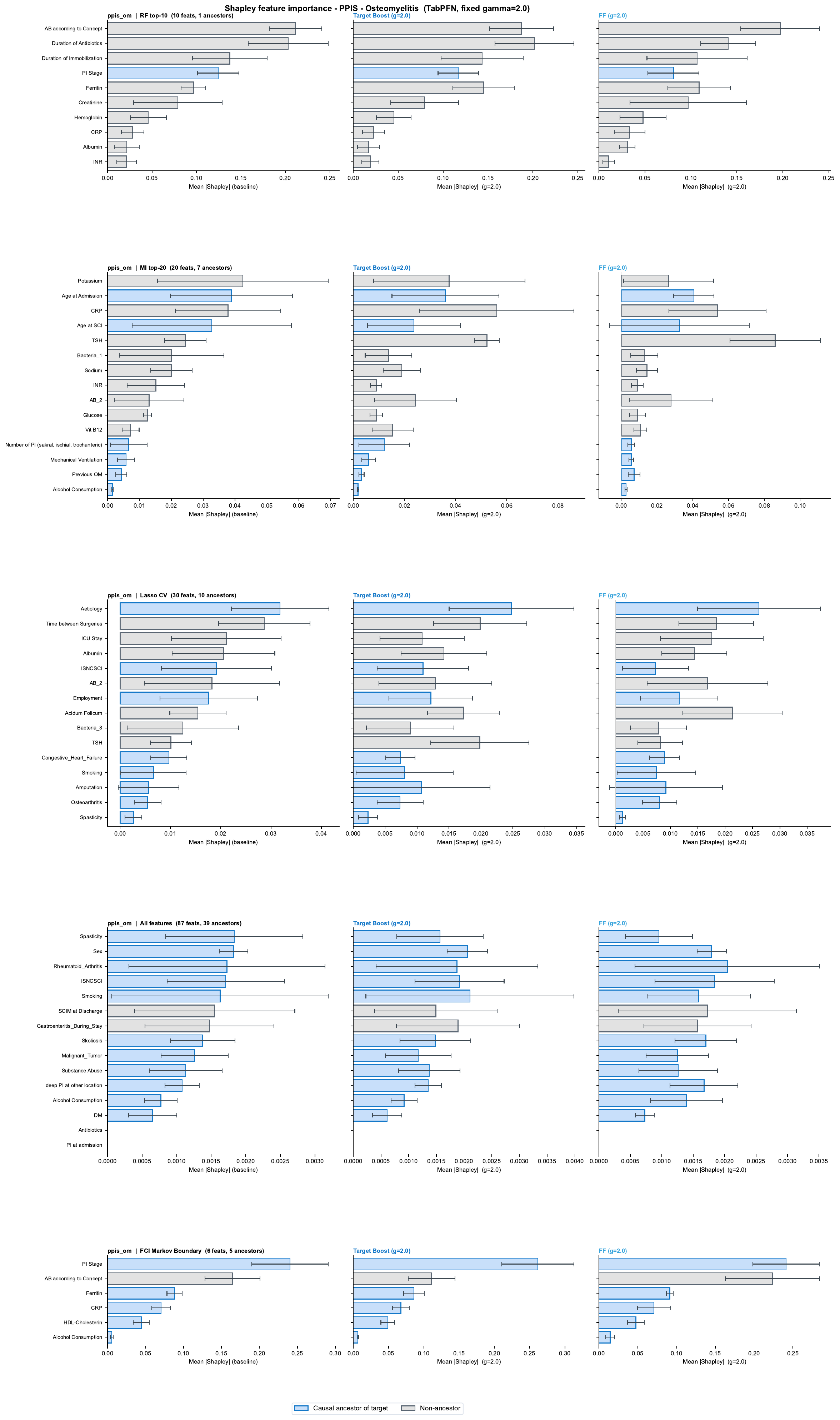}
\caption{Marginal Shapley feature importance for PPIS osteomyelitis across all feature sets. On the constrained FCI Markov blanket panel (6 features, 5 ancestors), the two modes move the ancestor/non-ancestor attribution ratio in opposite directions. Target-Boost raises it from $2.73$ to $4.24$ ($+55\%$), whereas the feature--feature control lowers it (FF: $2.08$, $-24\%$), so the effective $y\!\to\!x$ mode concentrates attribution on the causal ancestors.}
\label{fig:importance_om}
\end{figure}

\begin{figure}[p]
\centering
\includegraphics[width=\linewidth,height=0.93\textheight,keepaspectratio]{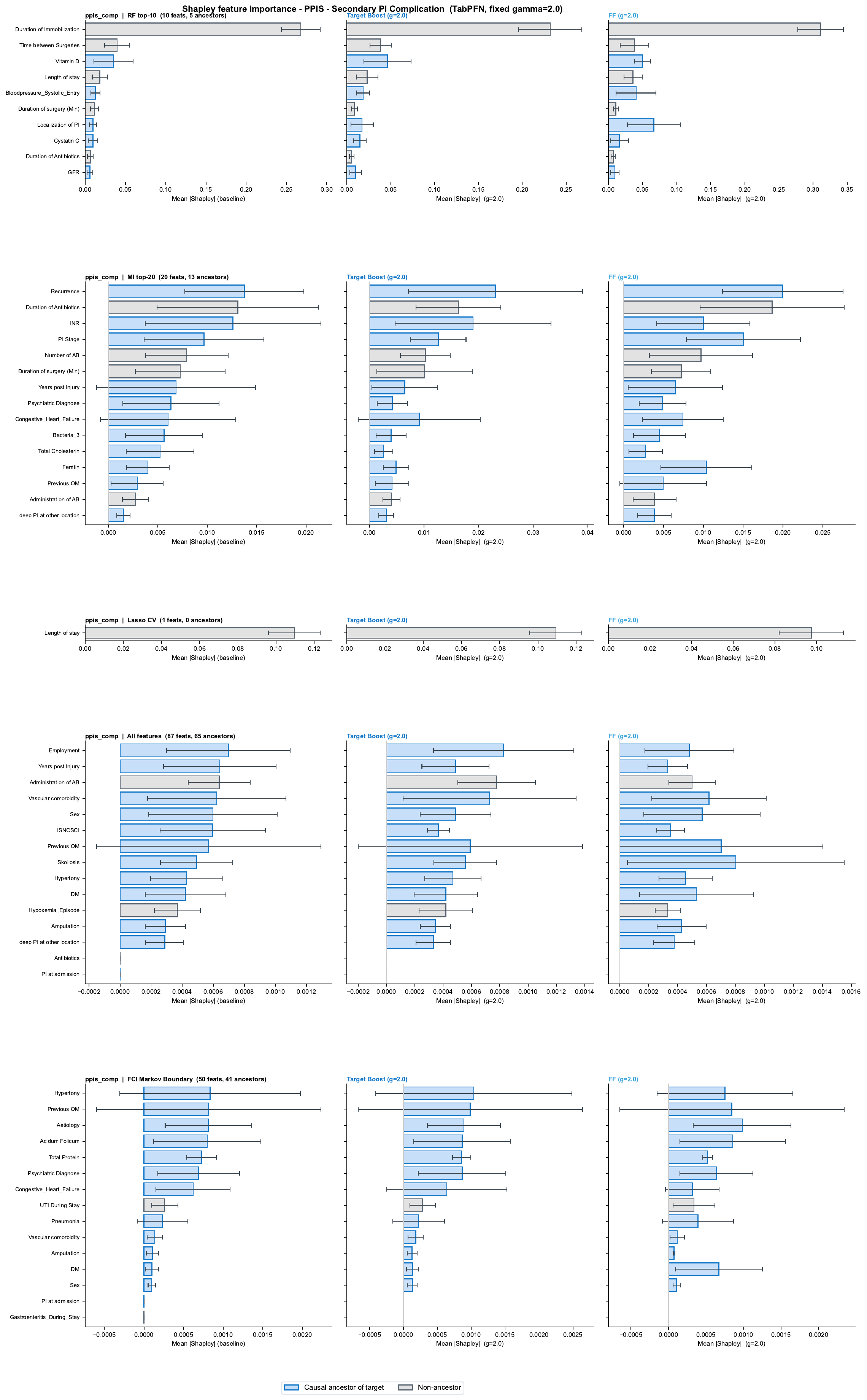}
\caption{Marginal Shapley feature importance for PPIS post-surgical complication across all feature sets. On the non-deployed RF top-10 set, the feature--feature (FF) control produces the largest attribution shift (ancestor/non-ancestor ratio $0.21\rightarrow0.45$, $+114\%$), yet this redistribution does not translate into a predictive gain, and the leak-free steering effect is within noise. On the deployed constrained graph set, Target-Boost, the effective mode, most concentrates attribution on the ancestors ($0.74\rightarrow1.01$, $+37\%$, versus FF $+6\%$). Duration of Immobilization, a non-ancestor, dominates the baseline.}
\label{fig:importance_comp}
\end{figure}

\clearpage
\section{Markov-blanket subgraphs}

Each graph places the target's causal Markov-blanket variables in the temporal tier in which they are measured (admission $\rightarrow$ discharge) and emphasizes the arrows into the outcome. For the two large binary blankets, UTI and complication, the graphs show their depth-2 causal neighborhood instead. Reading conventions follow the main-text example (Figure~\ref{fig:mbgraph_comp}). For the large length-of-stay blankets only edges touching the outcome are drawn, to keep the graph legible.

\begin{figure}[p]
\centering
\includegraphics[width=\linewidth,height=0.9\textheight,keepaspectratio]{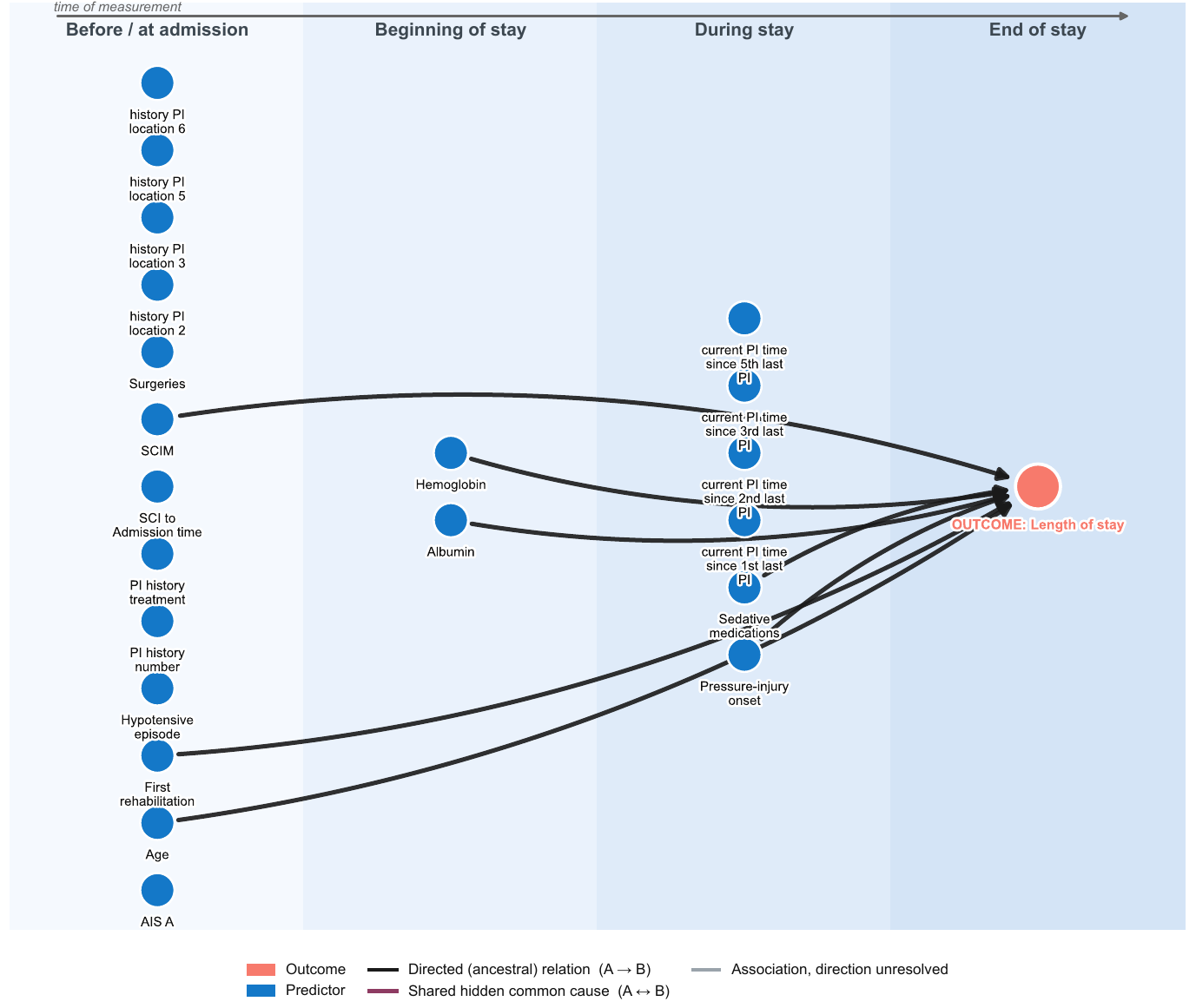}
\caption{Temporally-organized causal graph for \textbf{HAPI length of stay} (full Markov blanket, $\lvert\mathrm{MB}\rvert=21$, $\alpha=0.05$). Reading conventions as in Figure~\ref{fig:mbgraph_comp}.}
\label{fig:mbgraph_hapi_los}
\end{figure}

\begin{figure}[p]
\centering
\includegraphics[width=\linewidth,height=0.9\textheight,keepaspectratio]{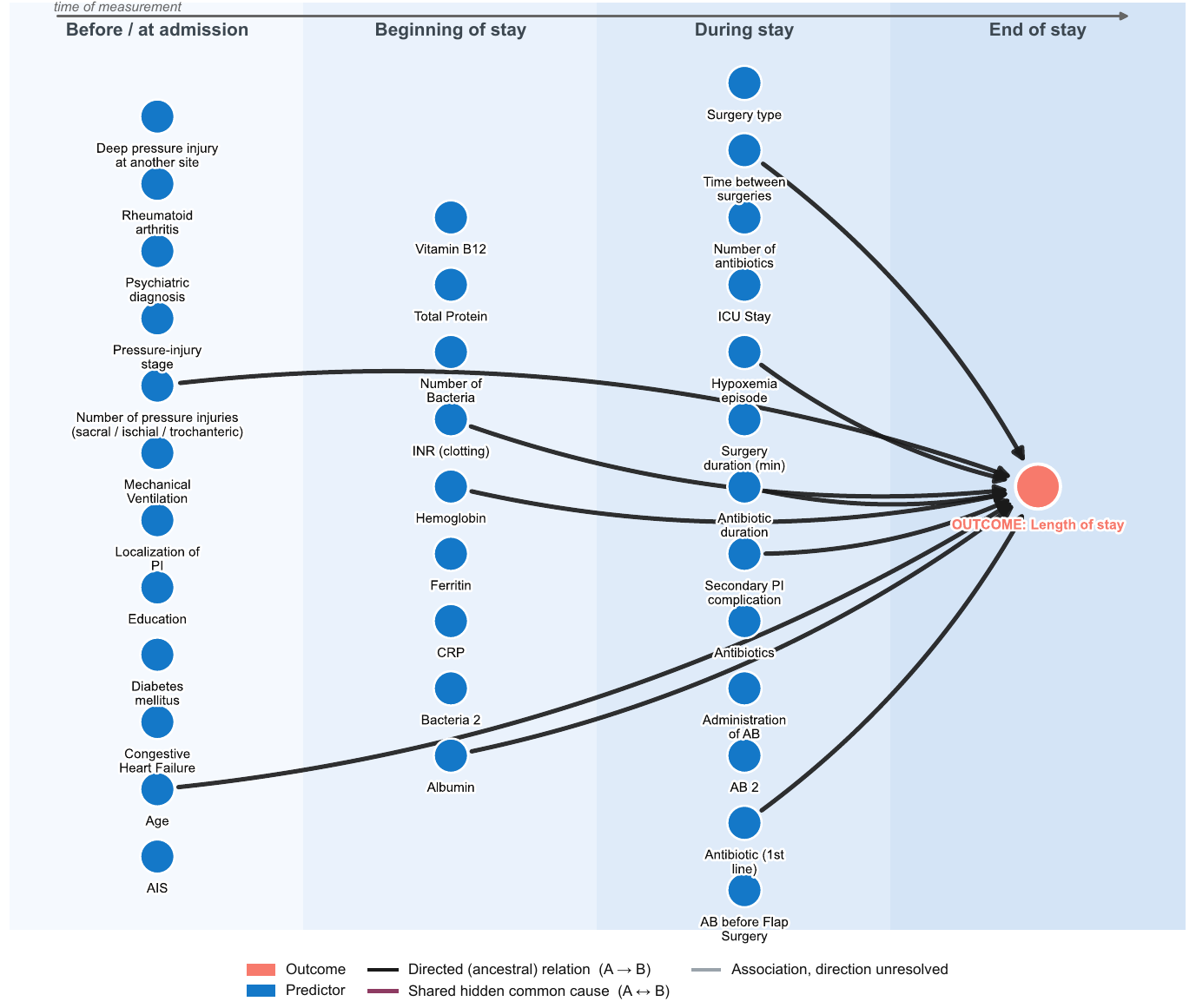}
\caption{Temporally-organized causal graph for \textbf{PPIS length of stay} (full Markov blanket, $\lvert\mathrm{MB}\rvert=34$, $\alpha=0.05$). For this large blanket only the edges incident to the outcome are drawn; predictor--predictor associations are omitted for legibility.}
\label{fig:mbgraph_ppis_los}
\end{figure}

\begin{figure}[p]
\centering
\includegraphics[width=\linewidth,height=0.9\textheight,keepaspectratio]{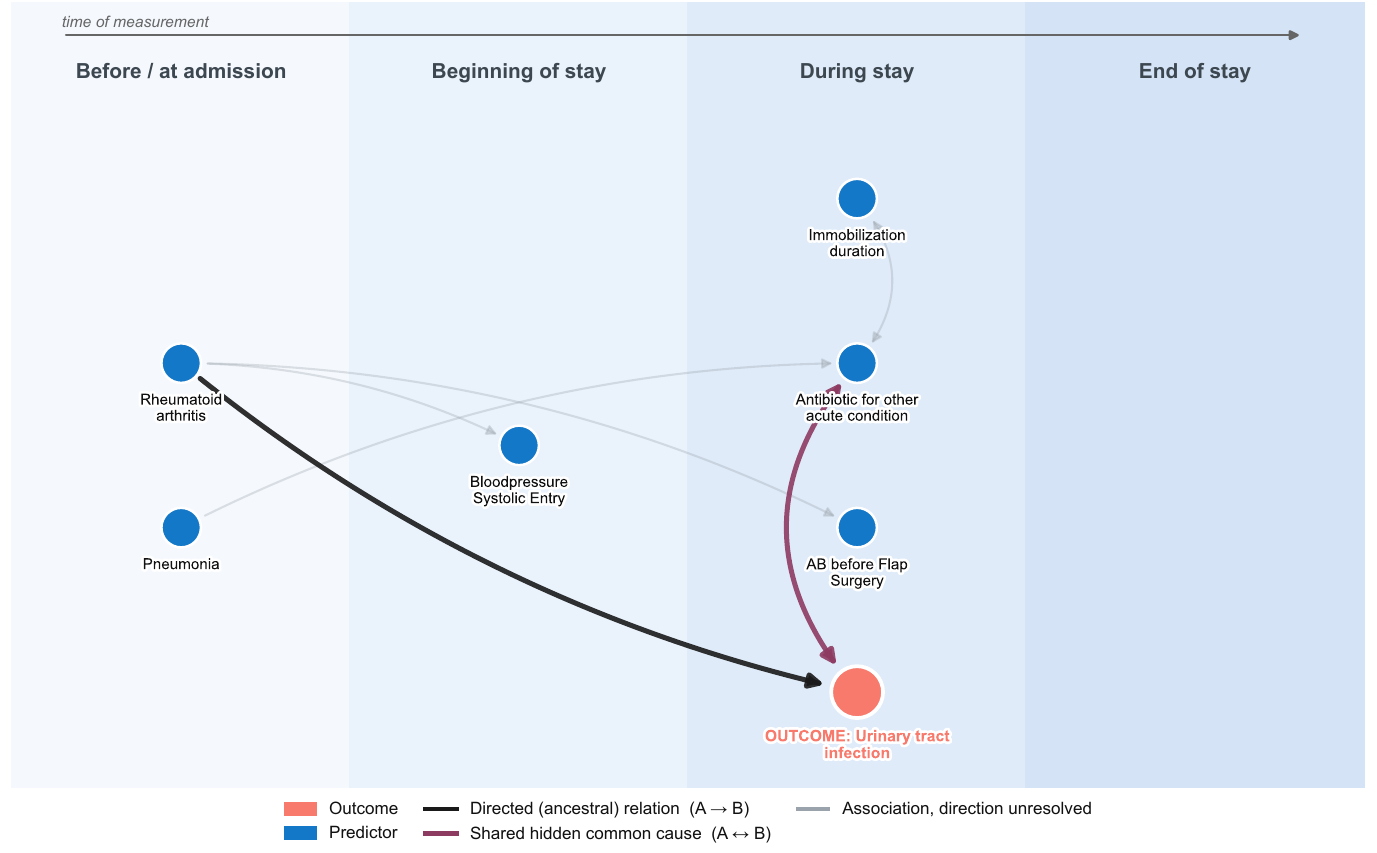}
\caption{Temporally-organized causal graph for \textbf{PPIS urinary tract infection during stay} (depth-2 neighborhood of the outcome, $\lvert\mathcal{N}\rvert=6$ of a 50-member Markov blanket, shown for legibility; $\alpha=0.05$). Reading conventions as in Figure~\ref{fig:mbgraph_comp}.}
\label{fig:mbgraph_ppis_uti}
\end{figure}

\begin{figure}[p]
\centering
\includegraphics[width=\linewidth,height=0.9\textheight,keepaspectratio]{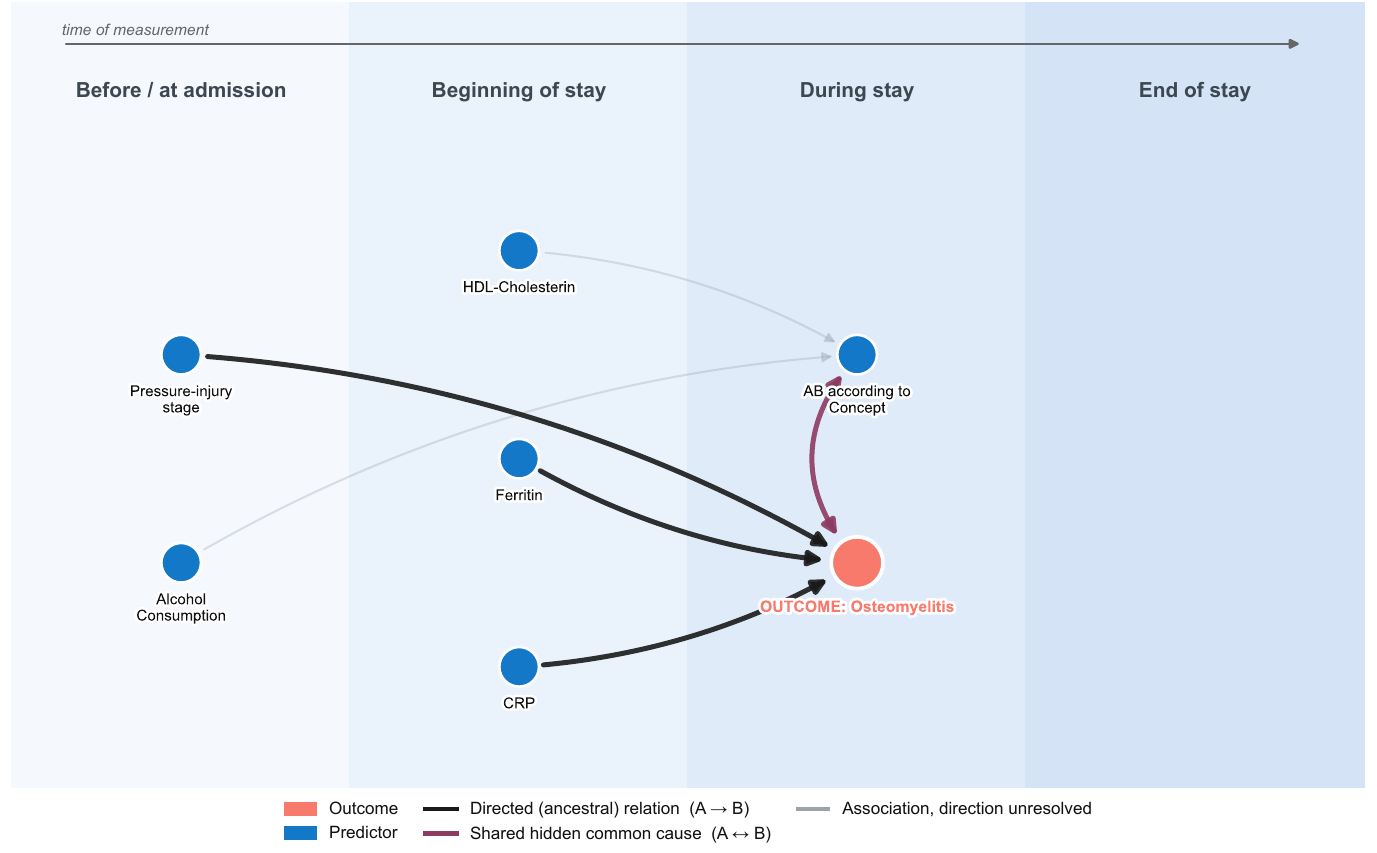}
\caption{Temporally-organized causal graph for \textbf{PPIS osteomyelitis} (full Markov blanket, $\lvert\mathrm{MB}\rvert=6$, $\alpha=0.05$; the compact deployed blanket). Reading conventions as in Figure~\ref{fig:mbgraph_comp}.}
\label{fig:mbgraph_ppis_om}
\end{figure}

\begin{table}[ht]
\centering
\caption{Full temporally-available causal Markov blanket at $\alpha=0.05$ (total-conditioning m-separation, confounders retained; restricted to variables measured no later than the target). For every target this is the complete deployed Markov blanket used for prediction. For legibility, the per-target graphs (Figures~\ref{fig:mbgraph_comp} and~\ref{fig:mbgraph_hapi_los}--\ref{fig:mbgraph_ppis_om}) draw only the depth-2 causal neighborhood of the outcome, a subset of these members. The reduction is largest for UTI and secondary PI complication, whose blankets have $50$ features each. The header column lists (cohort, $\alpha$, $\lvert\mathrm{MB}\rvert$), the full blanket size.}
\label{tab:mb_features}
\small
\begin{tabularx}{\linewidth}{l X}
\toprule
Target (cohort, $\alpha$, $\lvert\mathrm{MB}\rvert$) & Temporally-available Markov-blanket members \\
\midrule
HAPI LoS \\($\alpha=0.05$, 21) &
AIS A; Age at SCI; Albumin; First rehabilitation; Hemoglobin; Hypotensive episode; PI history number; PI history treatment; PI\_onset; SCI to Admission time; SCIM; Sedative medications; Surgeries; current PI time since 1st last PI; current PI time since 2nd last PI; current PI time since 3rd last PI; current PI time since 5th last PI; history PI location 2; history PI location 3; history PI location 5; history PI location 6 \\
\midrule
PPIS LoS \\($\alpha=0.05$, 34) &
AB before Flap Surgery; AB\_1; AB\_2; AIS; Administration of AB; Age at SCI; Albumin; Antibiotics; Bacteria\_2; CRP; Complication; Congestive\_Heart\_Failure; DM; Duration of Antibiotics; Duration of surgery (Min); Education; Ferritin; Hemoglobin; Hypoxemia\_Episode; ICU Stay; INR; Localization of PI; Mechanical Ventilation; Number of AB; Number of Bacteria; Number of PI (sakral, ischial, trochanteric); PI Stage; Psychiatric Diagnosis; Rheumatoid\_Arthritis; Time between Surgeries; Total Protein; Type of Surgery; Vit B12; deep PI at other location \\
\midrule
PPIS UTI \\($\alpha=0.05$, 50) &
AB before Flap Surgery; AB for other Acute Condition; Acidum Folicum; Aetiology; Age at Admission; Age at SCI; AIS; Albumin; Alcohol Consumption; Amputation; Bloodpressure\_Systolic\_Entry; Body\_Temperature\_Abnormal; BP\_Episode\_Abnormal; Calcium; Complication; Congestive\_Heart\_Failure; CRP; deep PI at other location; DM; Duration of Immobilization; Duration of surgery (Min); Education; Employment; Gastroenteritis\_During\_Stay; Hemoglobin; Hypertony; ISNCSCI; Level of injury; Localization of PI; Mechanical Ventilation; Number of PI (sakral, ischial, trochanteric); OSAS; Osteoarthritis; PI at admission; PI Stage; Pneumonia; Previous OM; Psychiatric Diagnosis; Recurrence; Rheumatoid\_Arthritis; Sex; Smoking; Social Status; Spasticity; Total Protein; Type of Residence; Type of Surgery; Vascular comorbidity; Vit B12; Years post Injury \\
\midrule
PPIS OM \\($\alpha=0.05$, 6) &
AB according to Concept; Alcohol Consumption; CRP; Ferritin; HDL-Cholesterin; PI Stage \\
\midrule
PPIS Complication \\($\alpha=0.05$, 50) &
AB before Flap Surgery; AB for other Acute Condition; Acidum Folicum; Aetiology; Age at Admission; Age at SCI; AIS; Albumin; Alcohol Consumption; Amputation; Bloodpressure\_Systolic\_Entry; Body\_Temperature\_Abnormal; BP\_Episode\_Abnormal; Calcium; Congestive\_Heart\_Failure; CRP; deep PI at other location; DM; Duration of Immobilization; Duration of surgery (Min); Education; Employment; Gastroenteritis\_During\_Stay; Hemoglobin; Hypertony; ISNCSCI; Level of injury; Localization of PI; Mechanical Ventilation; Number of PI (sakral, ischial, trochanteric); OSAS; Osteoarthritis; PI at admission; PI Stage; Pneumonia; Previous OM; Psychiatric Diagnosis; Recurrence; Rheumatoid\_Arthritis; Sex; Smoking; Social Status; Spasticity; Total Protein; Type of Residence; Type of Surgery; UTI During Stay; Vascular comorbidity; Vit B12; Years post Injury \\
\bottomrule
\end{tabularx}
\end{table}

\clearpage
\section{Validation and robustness analyses}
This section reports the additional validation analyses referenced in the Results: decision-time-stratified length-of-stay prediction, patient-grouped cross-validation, calibration and average precision with paired steering-effect confidence intervals, decision-curve (net-benefit) analysis, steering negative controls, imputation-method sensitivity, and graph stability. All use the same 5-fold protocol and feature sets as the main results unless stated otherwise.

\begin{table}[ht]
\centering
\caption{Decision-time-stratified length-of-stay prediction ($R^2$, 5-fold cross-validation). Each length-of-stay feature set is split by temporal tier into admission-only, pre-surgery (adds beginning-of-stay laboratory values), and retrospective (the full Markov blanket, adds during-stay variables). On PPIS the strong retrospective result collapses to $R^2\approx0.31$--$0.34$ with TabPFN ($0.31$--$0.36$ across learners) once restricted to information available at or before surgery, whereas HAPI length of stay is already well predicted at admission. The retrospective PPIS model is explanatory rather than prospective.}
\label{tab:decision_time}
\small
\begin{tabularx}{\linewidth}{l l c c c}
\toprule
Cohort & Decision time & $n$ feat. & Gradient-boosted $R^2$ & TabPFN $R^2$ \\
\midrule
HAPI & Admission-only      & 11 & 0.646 & 0.690 \\
HAPI & Pre-surgery         & 13 & 0.663 & 0.719 \\
HAPI & Retrospective (all) & 21 & 0.684 & 0.745 \\
\midrule
PPIS & Admission-only      & 12 & 0.312 & 0.341 \\
PPIS & Pre-surgery         & 21 & 0.362 & 0.310 \\
PPIS & Retrospective (all) & 34 & 0.746 & 0.710 \\
\bottomrule
\end{tabularx}
\end{table}

\begin{table}[ht]
\centering
\caption{Patient-grouped versus episode-level cross-validation on PPIS. The 266 episodes come from 174 patients. Patient-grouped CV (GroupKFold / StratifiedGroupKFold on patient ID) prevents any patient from appearing in both training and validation folds and is the primary PPIS validation here. Metrics are for the TabPFN predictor on each target's deployed causal Markov blanket (no steering). Grouping lowers performance by $0.04$--$0.12$ for length of stay, UTI and osteomyelitis, so the episode-level column is an optimistic upper bound. Secondary PI complication is essentially unchanged.}
\label{tab:groupcv}
\small
\begin{tabularx}{\linewidth}{l l c c}
\toprule
Target & Metric & Episode-level & Patient-grouped \\
\midrule
PPIS length of stay        & $R^2$   & 0.710 & 0.625 \\
UTI during stay            & AUROC & 0.894 & 0.778 \\
Osteomyelitis              & AUROC & 0.848 & 0.807 \\
Secondary PI complication  & AUROC & 0.776 & 0.783 \\
\bottomrule
\end{tabularx}
\end{table}

\begin{table}[ht]
\centering
\caption{Generalizability of the discovery procedure. A central claim of the method is that the causal-fusion procedure recovers predictive feature sets from new data, not merely that a fixed model predicts held-out patients (the deployment question, Table~\ref{tab:groupcv}). We test this with a fully nested cross-validation. Inside each outer fold, every method's feature set is re-derived from the training data only (per-cohort FCI at $\alpha=0.05$ for the constrained FCI Markov blanket), a single fixed TabPFN is fit without best-of-$N$ model selection, and the held-out fold is predicted. PPIS uses patient-grouped folds. Nesting removes the full-data selection advantage that every method enjoys in the episode-level benchmark, so all numbers fall relative to the main-text figures. Under nesting, the re-discovered constrained graph set ranks first on PPIS length of stay ($R^2=0.674$, ahead of every automatic set) and second on UTI ($0.765$ AUROC, behind Lasso CV at $0.785$), evidence that the fusion procedure recovers predictive causal structure from new data without overfitting the development cohort. It trails on osteomyelitis, secondary PI complication, and HAPI length of stay, consistent with the temporal-availability restriction excluding the during-stay infection and inflammatory markers that the automatic baselines are free to exploit. Bold: best (leak-corrected) per row. Metrics: $R^2$ (length of stay), AUROC (binary); non-steered TabPFN throughout.}
\label{tab:nestedcv}
\small
\begin{tabularx}{\linewidth}{X l c c c c c}
\toprule
Target & Metric & Causal MB & RF top-10 & MI top-20 & Lasso CV & All \\
\midrule
PPIS length of stay        & $R^2$   & \textbf{0.674} & 0.640 & 0.628 & 0.622 & 0.656 \\
HAPI length of stay        & $R^2$   & 0.706 & 0.718 & 0.722 & 0.749 & \textbf{0.761} \\
UTI during stay            & AUROC & 0.765 & 0.742 & 0.761 & \textbf{0.785} & 0.743 \\
Osteomyelitis              & AUROC & 0.783 & \textbf{0.818} & 0.734 & 0.789 & 0.804 \\
Secondary PI complication  & AUROC & 0.690 & 0.739 & \textbf{0.760} & 0.654 & 0.742 \\
\bottomrule
\end{tabularx}
\end{table}

\begin{table}[ht]
\centering
\caption{Calibration, average precision, and paired steering effect for the three PPIS binary outcomes (TabPFN on the deployed causal Markov blanket; base and Target-Boost-steered $y\!\to\!x$, fixed $\gamma=2.0$ on independent data). Brier score, AUPRC and calibration slope are for the base predictor (slope $1$ = perfect; AUPRC compares to prevalence). $\Delta$AUROC is the paired bootstrap difference (Target-Boost $-$ base) with a 95\% CI, computed leak-free (deployable fixed $\gamma$, no per-fold selection). On this cohort ($n=266$) the deployable steering effect on all three outcomes is small and not statistically distinguishable from zero. Every $\Delta$AUROC CI includes $0$, and the inert feature--feature (FF) control behaves similarly (Table~\ref{tab:clinical_mode_grid}). Steering's causal specificity is therefore established on the synthetic benchmark (Table~\ref{tab:steer_significance}) and the random-graph/shuffled-label controls (Table~\ref{tab:negcontrol}), not on these clinical AUROC deltas.}
\label{tab:calibration}
\small
\begin{tabularx}{\linewidth}{l c c c c c}
\toprule
Target & Prev. & AUROC (base$\to$Boost) & AUPRC & Brier (base$\to$Boost) & Calib.\ slope \\
\midrule
UTI                        & 0.29 & $0.894\to0.891$ & 0.83 & $0.106\to0.110$ & 0.79 \\
Osteomyelitis              & 0.43 & $0.848\to0.844$ & 0.81 & $0.154\to0.156$ & 0.97 \\
Secondary PI complication  & 0.23 & $0.776\to0.767$ & 0.52 & $0.144\to0.146$ & 0.78 \\
\bottomrule
\end{tabularx}\\[2pt]
\footnotesize Paired $\Delta$AUROC (Target-Boost $-$ base), 95\% CI: UTI $-0.003\,[-0.011,+0.005]$; osteomyelitis $-0.004\,[-0.014,+0.005]$; secondary PI complication $-0.010\,[-0.023,+0.003]$. All three include $0$.
\end{table}

\begin{figure}[ht]
\centering
\includegraphics[width=\linewidth]{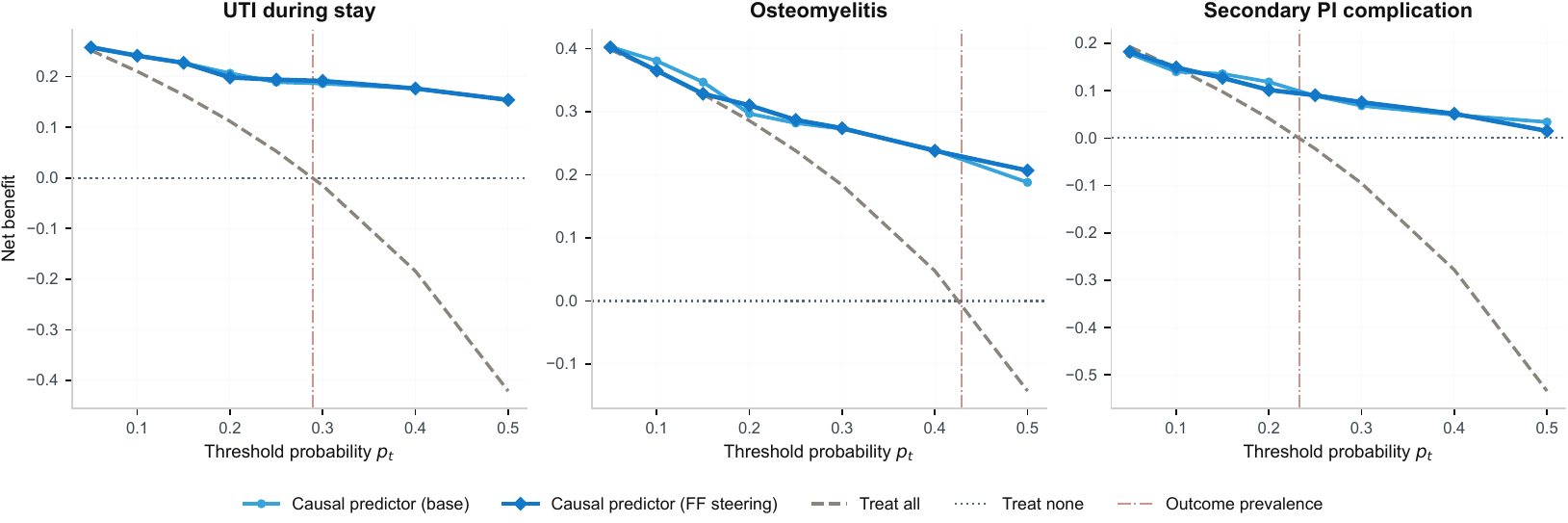}
\caption{Net-benefit decision curves of base and Target-Boost-steered ($y\!\to\!x$, $\gamma=2.0$) TabPFN on the deployed causal Markov blanket for the three PPIS binary outcomes. Net benefit is plotted against the threshold probability $p_t$; the dashed ``treat all'' line and the ``treat none'' baseline at $0$ are the two default strategies, and the vertical dash-dotted line marks the outcome prevalence. The Target-Boost-steered predictor lies above both defaults across the full range for UTI and osteomyelitis and for $p_t\ge0.15$ for secondary PI complication, and meets or exceeds the base predictor throughout.}
\label{fig:decision_curves}
\end{figure}

\noindent\textbf{Decision-curve (net-benefit) analysis.} Discrimination and calibration do not by themselves establish clinical usefulness. Figure~\ref{fig:decision_curves} reports a decision-curve analysis~\cite{Vickers2006DCA}: the net benefit of acting on each predictor as a function of the threshold probability $p_t$ at which a clinician would intervene, against the two default strategies ``treat all'' and ``treat none''. A predictor is useful over the threshold range where its curve lies above both defaults. Both the base and Target-Boost-steered causal predictors deliver positive net benefit above the two default strategies (UTI and osteomyelitis across essentially the full plausible range $p_t\in[0.05,0.5]$; secondary PI complication for $p_t\gtrsim0.15$), so the causal predictor is decision-analytically useful relative to treat-all/treat-none. Consistent with the near-zero $\Delta$AUROC (Table~\ref{tab:calibration}), the steered and base curves essentially coincide, and on this cohort attention steering does not add net benefit over the unsteered causal predictor. This is an internal, single-center analysis and should be read as hypothesis-generating rather than as evidence of deployment-ready clinical utility.

\begin{table}[ht]
\centering
\caption{Steering negative controls (PPIS binary outcomes; deployable Target-Boost, $y\!\to\!x$, fixed $\gamma=2.0$). Paired $\Delta$AUROC (steered $-$ base) from steering toward the real graph ancestors versus a random same-size ancestor set, averaged over three random sets. Shuffled-label sanity check: base AUROC on permuted labels, expected $\approx0.5$. On this cohort the deployable gains are small for both the real ancestors and the random sets and lie within the paired-bootstrap noise (Table~\ref{tab:calibration}); the real ancestor set does not exceed the random-set mean on any outcome. The clinical negative control is thus underpowered at this sample size, and causal specificity is established primarily on the synthetic benchmark with known ground truth (Table~\ref{tab:steer_significance}) and the PPIS length-of-stay specificity analysis (Table~\ref{tab:clinical_mode_grid}).}
\label{tab:negcontrol}
\small
\begin{tabularx}{\linewidth}{l c c c}
\toprule
Target & Real ancestors $\Delta$AUROC & Random-set $\Delta$AUROC & Shuffled-label AUROC \\
\midrule
UTI                        & $-0.003$ & $+0.001$ & 0.49 \\
Osteomyelitis              & $-0.004$ & $-0.003$ & 0.46 \\
Secondary PI complication  & $-0.010$ & $-0.003$ & 0.45 \\
\bottomrule
\end{tabularx}
\end{table}

\begin{table}[ht]
\centering
\caption{Imputation-method sensitivity. Primary metric (TabPFN on each target's deployed causal Markov blanket; 5-fold CV) under the pipeline-default median/mode imputation, MICE (IterativeImputer with Bayesian ridge), and a missForest-like imputer (IterativeImputer with extremely randomized trees). Median-column values are this sensitivity run's default-imputer result and match the main benchmark to within $0.002$. Most binary-outcome metrics are robust to the imputation method, and MICE/missForest even improve UTI slightly. HAPI length of stay is sensitive to the imputer (median $R^2=0.745$ versus $\approx0.65$ under MICE/missForest), a non-negligible difference that is a caveat on the length-of-stay results. This table reports prediction-time imputation; graph-level imputation sensitivity (effect on the recovered edges and Markov-blanket sizes) is reported in Supplementary Table~\ref{tab:sl_sensitivity}.}
\label{tab:imputation}
\small
\begin{tabularx}{\linewidth}{l l c c c}
\toprule
Target & Metric & Median & MICE & missForest \\
\midrule
HAPI length of stay        & $R^2$   & 0.745 & 0.650 & 0.646 \\
PPIS length of stay        & $R^2$   & 0.710 & 0.721 & 0.725 \\
UTI during stay            & AUROC & 0.896 & 0.906 & 0.904 \\
Osteomyelitis              & AUROC & 0.848 & 0.849 & 0.838 \\
Secondary PI complication  & AUROC & 0.775 & 0.760 & 0.781 \\
\bottomrule
\end{tabularx}
\end{table}

\subsection*{Significance-threshold ($\alpha$) sensitivity}

The conditional-independence threshold $\alpha$ trades graph sparsity against recall. All primary results use $\alpha=0.05$. Here we quantify the effect of relaxing to $\alpha=0.10$ by re-running the entire deployed pipeline (direct FCI with the fused cross-cohort backbone, total-conditioning Markov-blanket extraction, and the same temporal leakage-safe filter) at each threshold. Relaxing $\alpha$ makes the pooled PPIS graph denser and less resolved. Edges rise from $158$ to $205$ and the bidirected fraction (unresolved or latent-confounded edges) from $28.5\%$ to $37.1\%$, so a disproportionate share of the added edges are the least reliable kind. The leak-safe Markov blankets grow accordingly (Table~\ref{tab:alpha_sensitivity}) and remain temporally leak-free by construction, with zero tier violations at either threshold. Prediction is essentially unchanged on every target except PPIS length of stay, where the blanket expands from $34$ to $79$ features and $R^2$ rises from $0.710$ to $0.783$ by absorbing many additional during-stay predictors. We therefore keep $\alpha=0.05$ as the primary operating point. It yields sparser, better-resolved graphs and more parsimonious blankets at almost no predictive cost, whereas $\alpha=0.10$ buys a single-outcome gain concentrated in leakage-adjacent during-stay signal on a denser, less trustworthy graph.

\begin{table}[ht]
\centering
\caption{Significance-threshold ($\alpha$) sensitivity of the deployed causal Markov blanket. For each target, the leak-safe blanket size and TabPFN performance (AUROC for binary outcomes, $R^2$ for length of stay; 5-fold CV) at $\alpha=0.05$ (primary) and $\alpha=0.10$. Blankets are re-extracted through the full deployed pipeline at each threshold and remain temporally leak-free (zero tier violations). Relaxing $\alpha$ enlarges every blanket except the well-separated osteomyelitis target, but only PPIS length of stay gains materially, by absorbing additional during-stay predictors. Pooled PPIS graph density: $158$ edges / $28.5\%$ bidirected at $\alpha=0.05$ versus $205$ / $37.1\%$ at $\alpha=0.10$.}
\label{tab:alpha_sensitivity}
\small
\begin{tabularx}{\linewidth}{X l c c c c}
\toprule
 & & \multicolumn{2}{c}{$\alpha=0.05$ (primary)} & \multicolumn{2}{c}{$\alpha=0.10$} \\
\cmidrule(lr){3-4}\cmidrule(lr){5-6}
Target & Metric & $\lvert\mathrm{MB}\rvert$ & Score & $\lvert\mathrm{MB}\rvert$ & Score \\
\midrule
UTI during stay            & AUROC & 50 & 0.894 & 65 & 0.890 \\
Osteomyelitis              & AUROC & 6  & 0.848 & 6  & 0.848 \\
Secondary PI complication  & AUROC & 50 & 0.776 & 65 & 0.785 \\
PPIS length of stay        & $R^2$   & 34 & 0.710 & 79 & 0.783 \\
HAPI length of stay        & $R^2$   & 21 & 0.745 & 24 & 0.745 \\
\bottomrule
\end{tabularx}
\end{table}

\begin{table}[ht]
\centering
\caption{\textbf{Relaxing the threshold beyond the deployable operating point: the fifth fusion stage.}
The information-fusion trajectory (Figure~\ref{fig:trajectory}) ends at Stage~4, full fusion at the
deployable $\alpha=0.05$. Here we report the additional Stage~5, which relaxes the per-cohort
conditional-independence threshold to $\alpha=0.10$, for the same five model families and the same
protocol (5-fold cross-validated $R^2$ on the stage's Markov blanket). On HAPI the relaxed threshold
leaves the blanket unchanged (11 members), so every model is identical. On PPIS it admits 12 further
variables ($23\rightarrow35$), and the effect depends on the model. The flexible learners exploit
them (TabPFN $0.735\rightarrow0.788$, XGBoost $0.731\rightarrow0.775$, random forest
$0.726\rightarrow0.753$) whereas the penalized linear models collapse (ridge $0.586\rightarrow0.264$,
lasso $0.571\rightarrow0.272$), because the added collinearity cannot be absorbed by an $\ell_2$/$\ell_1$
penalty at this sample size. This model-dependence, rather than any gain in the deployed pipeline, is why
$\alpha=0.05$ is retained as the primary operating point (Table~\ref{tab:alpha_sensitivity}) and why the
relaxed stage is reported here, outside the main trajectory. These are the trajectory's
own per-stage blankets, distinct from the deployed leak-safe blankets of
Table~\ref{tab:alpha_sensitivity}, which grow $34\rightarrow79$ on PPIS and $21\rightarrow24$ on HAPI.}
\label{tab:alpha_trajectory}
\small
\begin{tabularx}{\linewidth}{l cc cc}
\toprule
 & \multicolumn{2}{c}{HAPI length of stay} & \multicolumn{2}{c}{PPIS length of stay} \\
\cmidrule(lr){2-3}\cmidrule(lr){4-5}
Model family & $\alpha=0.05$ (S4) & $\alpha=0.10$ (S5) & $\alpha=0.05$ (S4) & $\alpha=0.10$ (S5) \\
\midrule
Markov-blanket size & 11 & 11 & 23 & 35 \\
\midrule
TabPFN         & \textbf{0.718} & \textbf{0.718} & 0.735 & \textbf{0.788} \\
XGBoost        & 0.647 & 0.647 & 0.731 & 0.775 \\
Random forest  & 0.676 & 0.676 & 0.726 & 0.753 \\
Ridge          & 0.485 & 0.485 & 0.586 & 0.264 \\
Lasso          & 0.481 & 0.481 & 0.571 & 0.272 \\
\bottomrule
\end{tabularx}
\end{table}

\begin{table}[ht]
\centering
\caption{Graph stability. \textbf{(a)} Per-cohort FCI Markov-blanket size for length of stay at three significance thresholds; the PPIS blanket is markedly more $\alpha$-sensitive than the HAPI blanket, consistent with learning a high-dimensional PAG from a few hundred patients. \textbf{(b)} Age-harmonization sensitivity of the three pooled-shared LoS edges: bootstrap adjacency frequency (30 resamples of the pooled backbone, $n=516$, $\alpha=0.10$) under four age constructions. Consistent age harmonization uses either age at SCI, which both cohorts record, or within-cohort $z$-scoring, which removes the anchor/scale mismatch. In both cases the Age--LoS and Hemoglobin--LoS adjacencies are recovered in every resample and Albumin--LoS in a minority (albumin has high missingness). Only the units-inconsistent age-at-admission construction, which relies on the noisy SCI-to-admission interval, destabilizes Age--LoS. All three edges are recovered in the full sample under the consistent harmonizations. Per-cohort support for all three edges, on the shared backbone and in each cohort's full graph, is reported in Table~\ref{tab:edge_stability}.}
\label{tab:stability}
\small
\begin{tabularx}{\linewidth}{X c c c}
\toprule
\textbf{(a)} $\alpha$ (LoS MB size) & HAPI & PPIS & \\
\midrule
$\alpha=0.01$ & 13 & 15 & \\
$\alpha=0.05$ & 21 & 34 & \\
$\alpha=0.10$ & 23 & 79 & \\
\midrule
\textbf{(b)} Age harmonization (adjacency freq.) & Age--LoS & Alb--LoS & Hb--LoS \\
\midrule
Age at SCI, both cohorts                              & 1.00 & 0.37 & 1.00 \\
Within-cohort $z$-score                               & 1.00 & 0.37 & 1.00 \\
Age at SCI $\leftrightarrow$ admission (as reported)  & 1.00 & 0.37 & 1.00 \\
Age at admission, both cohorts                        & 0.07 & 0.50 & 1.00 \\
\bottomrule
\end{tabularx}
\end{table}

\begin{table}[ht]
\centering
\caption{Structure-learning sensitivity of the pooled-shared length-of-stay edges and the per-cohort LoS Markov-blanket size to the conditional-independence test and the imputer. \textbf{(a)} Presence of each pooled LoS edge under Fisher-$z$ (default), a kernel test (KCI), a nonparanormal (rank-to-normal) Fisher-$z$, and median/MICE/missForest graph-level imputation. All three edges survive the kernel test and every imputer; only Albumin$\rightarrow$LoS drops under the nonparanormal transform (albumin has high missingness). \textbf{(b)} Per-cohort LoS Markov-blanket size at the canonical $\alpha=0.05$. The blanket is unchanged under the kernel test and grows under MICE and under the rank-based transform, far more on PPIS than on HAPI, consistent with the PPIS graph being denser and less stable. $^{\dagger}$The nonparanormal run uses a different Markov-blanket derivation and is therefore read against its own Fisher-$z$ baseline of 25\,/\,83, not against the 21\,/\,34 of the rows above. A full per-cohort re-run with the kernel (KCI) test, nonparametric and appropriate for mixed discrete--continuous data, yields identical Markov blankets (HAPI~21, PPIS~34, and PPIS binary-outcome blankets 50/6/50), indicating the recovered structure is not an artifact of the Fisher-$z$ Gaussianity assumption. Full-pipeline nested cross-validation, repeating the entire pooled$+$temporal-tier fusion inside each outer fold, was investigated systematically but is computationally intractable on these cohorts. Most folds complete in $\sim$25~s, but a subset of dense training folds (one HAPI fold and four of the five PPIS folds) fail to terminate. On those subsets the FCI skeleton stays dense, so the constraint-based search evaluates prohibitively many conditional-independence tests and does not converge. Graph density rather than conditioning-set depth drives the non-termination, since the search stalls on the same folds with the FCI conditioning-set depth capped at $\in\{5,10,30\}$ or unbounded. We therefore report the simplified per-cohort-FCI nesting in Table~\ref{tab:nestedcv}, which re-derives the dominant feature-selection step inside each fold, as the tractable stand-in.}
\label{tab:sl_sensitivity}
\small
\begin{tabularx}{\linewidth}{X c c c}
\toprule
\textbf{(a)} Pooled edge into LoS & Age & Albumin & Hemoglobin \\
\midrule
Fisher-$z$ (default)      & \checkmark & \checkmark & \checkmark \\
KCI (kernel test)         & \checkmark & \checkmark & \checkmark \\
Nonparanormal Fisher-$z$  & \checkmark & ---        & \checkmark \\
MICE imputation           & \checkmark & \checkmark & \checkmark \\
missForest imputation     & \checkmark & \checkmark & \checkmark \\
\midrule
\textbf{(b)} Per-cohort LoS $\lvert\mathrm{MB}\rvert$ & \multicolumn{3}{c}{HAPI \quad / \quad PPIS} \\
\midrule
Fisher-$z$ $+$ median               & \multicolumn{3}{c}{21 \quad / \quad 34} \\
KCI (kernel, mixed-variable) $+$ median & \multicolumn{3}{c}{21 \quad / \quad 34} \\
Nonparanormal $+$ median$^{\dagger}$ & \multicolumn{3}{c}{23 \quad / \quad 84} \\
Fisher-$z$ $+$ MICE                 & \multicolumn{3}{c}{26 \quad / \quad 52} \\
\bottomrule
\end{tabularx}
\end{table}

\begin{table}[ht]
\centering
\caption{\textbf{Constraint-strategy ablation} for length-of-stay discovery (per-cohort FCI, $\alpha=0.05$, adjacency-search depth 3). We toggle the causal-discovery background knowledge and measure its effect on the recovered partial ancestral graph ($\lvert\text{edges}\rvert$), the temporally-filtered length-of-stay Markov blanket ($\lvert\mathrm{MB}\rvert$), and downstream TabPFN 5-fold $R^2$. Temporal-tier forbidden-only, required-only, and both give identical Markov blankets, because the tier rules constrain only edge orientation (a later-tier variable cannot cause an earlier-tier one) and the Markov blanket is adjacency-based. The pooled-shared stage is the ingredient that measurably changes prediction, and the middle two rows of each block localize which part of it does so. That stage bundles constraints derived from the pooled subgraph (orientations, plus forbidding pairs absent from it) with a union of its edges into the per-cohort PAG. The constraints alone leave the result unchanged from the tier-only row; the edge union alone reproduces the full pooled row. A third variant (orientation constraints $+$ edge union, dropping only the absent-pair forbidding) is identical to the union-alone row. The effect is therefore cross-cohort edge fusion, not a constraint on the per-cohort search. Fusion restores the shared directed-ancestor variables (Age, Albumin, Hemoglobin) to the blanket and improves HAPI length-of-stay $R^2$ by $+0.024$ (from $0.694$ to $0.718$), consistent with the Stage-4 jump in the information-fusion trajectory (Figure~\ref{fig:trajectory}). On the already-large PPIS blanket the addition is marginal. Edge counts vary by up to $\pm2$ between repeat runs of the discovery while the Markov blankets and $R^2$ are stable. In this ablation the Markov blankets are discovered once on the full cohort (with a fixed whole-cohort pooled prior) so that each row isolates the effect of the background constraint. Prediction is out-of-fold but selection is in-sample, so these absolute $R^2$ are optimistic relative to the fully per-fold leak-free numbers in the main results. The relative constraint effect is unbiased because every row shares the same whole-cohort discovery step.}
\label{tab:constraint_ablation}
\small
\begin{tabularx}{\linewidth}{l X c c c}
\toprule
Cohort & Constraint strategy & $\lvert$edges$\rvert$ & $\lvert\mathrm{MB}\rvert$ & TabPFN $R^2$ [95\% CI] \\
\midrule
HAPI & None (unconstrained)                       & 64  & 11 & 0.701 [0.62, 0.76] \\
HAPI & Temporal tiers (forbidden $=$ required $=$ both) & 68 & 9 & 0.694 [0.61, 0.77] \\
HAPI & \quad $+$ pooled \emph{constraints} only (no edge union) & 66 & 9 & 0.694 [0.61, 0.77] \\
HAPI & \quad $+$ pooled \emph{edge union} only (no constraints) & 77 & 11 & \textbf{0.718} [0.64, 0.78] \\
HAPI & Temporal tiers $+$ pooled-shared (both)      & 75  & 11 & \textbf{0.718} [0.64, 0.78] \\
\midrule
PPIS & None (unconstrained)                       & 154 & 23 & \textbf{0.743} [0.67, 0.79] \\
PPIS & Temporal tiers (forbidden $=$ required $=$ both) & 148 & 20 & \textbf{0.743} [0.67, 0.79] \\
PPIS & \quad $+$ pooled \emph{constraints} only (no edge union) & 150 & 20 & \textbf{0.743} [0.67, 0.79] \\
PPIS & \quad $+$ pooled \emph{edge union} only (no constraints) & 160 & 23 & 0.735 [0.67, 0.79] \\
PPIS & Temporal tiers $+$ pooled-shared (both)      & 158 & 23 & 0.735 [0.67, 0.79] \\
\bottomrule
\end{tabularx}
\end{table}

\FloatBarrier
\FloatBarrier
\begin{table}[ht]
\centering
\caption{\textbf{Discrimination, average precision and F1 for the three PPIS binary outcomes}, by feature-selection strategy, under a fixed TabPFN\,v2 model. Values are computed on stacked out-of-fold predictions with 95\% bootstrap confidence intervals (300 resamples). F1 is at the default $0.5$ threshold and is reported here because it is stated in the Methods but does not appear in any main-text figure. The constrained graph set leads on AUROC and average precision for secondary PI complication and is within the bootstrap interval of the best baseline on urinary tract infection. On osteomyelitis the automatic selectors and the all-feature model lead; their dominant predictors are non-ancestors of the target. F1 at a fixed threshold separates the strategies poorly on these prevalences and should not be read as a ranking.}
\label{tab:binary_metrics}
\small
\begin{tabularx}{\linewidth}{X c c c c}
\toprule
Feature set & $n$ & AUROC [95\% CI] & AUPRC [95\% CI] & F1 [95\% CI] \\
\midrule
\multicolumn{5}{l}{\textbf{Urinary tract infection}} \\
\midrule
\textbf{Constrained graph set} & 50 & 0.894 [0.850, 0.938] & 0.829 [0.757, 0.898] & 0.784 [0.707, 0.853] \\
All features & 75 & 0.894 [0.855, 0.931] & 0.810 [0.739, 0.875] & 0.720 [0.635, 0.792] \\
MI top-20 & 20 & 0.900 [0.859, 0.936] & 0.820 [0.745, 0.886] & 0.730 [0.652, 0.805] \\
RF top-10 & 10 & 0.866 [0.816, 0.916] & 0.745 [0.649, 0.838] & 0.699 [0.617, 0.778] \\
Lasso CV & 12 & 0.864 [0.824, 0.909] & 0.711 [0.609, 0.814] & 0.702 [0.615, 0.770] \\
RFE top-10 & 10 & 0.878 [0.829, 0.922] & 0.777 [0.687, 0.855] & 0.699 [0.614, 0.782] \\
Pooled MB only & 3 & 0.687 [0.612, 0.757] & 0.498 [0.398, 0.620] & 0.229 [0.116, 0.345] \\
\midrule
\multicolumn{5}{l}{\textbf{Osteomyelitis}} \\
\midrule
\textbf{Constrained graph set} & 6 & 0.848 [0.792, 0.894] & 0.811 [0.728, 0.891] & 0.734 [0.659, 0.797] \\
All features & 75 & 0.902 [0.864, 0.936] & 0.894 [0.848, 0.935] & 0.798 [0.737, 0.857] \\
MI top-20 & 20 & 0.869 [0.820, 0.911] & 0.845 [0.785, 0.899] & 0.710 [0.634, 0.780] \\
RF top-10 & 10 & 0.887 [0.848, 0.921] & 0.873 [0.823, 0.915] & 0.756 [0.691, 0.816] \\
Lasso CV & 13 & 0.871 [0.828, 0.908] & 0.850 [0.794, 0.898] & 0.747 [0.681, 0.809] \\
RFE top-10 & 10 & 0.882 [0.840, 0.917] & 0.874 [0.819, 0.911] & 0.769 [0.700, 0.820] \\
Pooled MB only & 3 & 0.769 [0.700, 0.823] & 0.743 [0.657, 0.811] & 0.614 [0.524, 0.697] \\
\midrule
\multicolumn{5}{l}{\textbf{Secondary PI complication}} \\
\midrule
\textbf{Constrained graph set} & 50 & 0.776 [0.699, 0.849] & 0.520 [0.403, 0.648] & 0.485 [0.346, 0.603] \\
All features & 75 & 0.752 [0.673, 0.823] & 0.515 [0.401, 0.656] & 0.486 [0.357, 0.606] \\
MI top-20 & 20 & 0.771 [0.698, 0.845] & 0.520 [0.415, 0.652] & 0.464 [0.335, 0.587] \\
RF top-10 & 10 & 0.770 [0.695, 0.842] & 0.499 [0.400, 0.637] & 0.500 [0.382, 0.609] \\
Lasso CV & 2 & 0.674 [0.594, 0.756] & 0.418 [0.320, 0.559] & 0.351 [0.218, 0.460] \\
RFE top-10 & 10 & 0.765 [0.685, 0.837] & 0.513 [0.403, 0.634] & 0.500 [0.379, 0.607] \\
Pooled MB only & 3 & 0.489 [0.413, 0.562] & 0.234 [0.185, 0.325] & 0.000 [0.000, 0.000] \\
\bottomrule
\end{tabularx}
\end{table}

\section{Reporting checklists and convergence diagnostics}
A completed TRIPOD+AI checklist \cite{Collins2024TRIPODAI} will be submitted as a separate reporting item; Table~\ref{tab:tripod} maps its principal items to the relevant sections. Because this is a clinical prediction/AI study on observational cohorts, a completed STROBE checklist~\cite{vonElm2007STROBE} (Table~\ref{tab:strobe}) accompanies the submission.

\begin{table}[ht]
\centering
\caption{TRIPOD+AI item map (abbreviated). ``Source'' points to where each item is addressed; several items are addressed as limitations.}
\label{tab:tripod}
\small
\begin{tabularx}{\linewidth}{l X}
\toprule
TRIPOD+AI item & Where addressed \\
\midrule
Title / abstract as prediction study & Reframed as hypothesis-generating; deployable (admission-time) vs.\ retrospective results distinguished (Abstract, Results) \\
Source of data, participants, setting & Methods (Clinical datasets); single-center, Swiss Paraplegic Center; Table~\ref{tab:table1} \\
Outcomes and predictors (with timing) & Methods; temporal-tier definitions; decision-time stratification (Table~\ref{tab:decision_time}) \\
Missing data & Table~\ref{tab:missingness}; imputation methods and sensitivity (Table~\ref{tab:imputation}) \\
Sample size & No formal a-priori sample-size calculation; analysis used all available records ($n=250$ HAPI, $n=266$ PPIS). Given $\le 87$ candidate variables, this is underpowered for high-dimensional structure learning and prediction; results are framed accordingly (Discussion) \\
Model development / selection & Feature and model selection performed post-discovery; nested cross-validation reported (Table~\ref{tab:nestedcv}); no multiplicity correction (Methods) \\
Model performance & AUROC/AUPRC/F1, $R^2$/RMSE/MAE; calibration and Brier (Table~\ref{tab:calibration}); patient-grouped as primary for PPIS (Table~\ref{tab:groupcv}) \\
Fairness / subgroups & Not formally assessed; sex and lesion-completeness (AIS) distributions in Table~\ref{tab:table1}; subgroup performance by sex and AIS is identified as future work \\
Validation (internal/external) & Internal only (single center); external, multi-site validation identified as essential (Discussion) \\
Availability of data and code & Data/Code Availability statements \\
\bottomrule
\end{tabularx}
\end{table}

\noindent\textbf{Convergence of the imputation--discovery loop.} The iterative loop was run to a Markov-blanket symmetric-difference tolerance $\tau=0.001$. On both clinical cohorts it reached this tolerance within a small number of iterations (typically $\le 5$), and no cohort failed to converge within the iteration cap. \textbf{Load-bearing recent citations.} Several methodological anchors are recent, not-yet-refereed preprints: the CRT-TabPFN conditional-independence test~\cite{Salem2026CRTTabPFN}, learnable causal-attention biases~\cite{Reuter2026PartialGraphs}, and evidence that TabPFN encodes recoverable causal structure~\cite{Swelam2025TabPFNCausal}. We flag them as such and do not rest any headline clinical claim on them.

\FloatBarrier

\FloatBarrier
\subsection{STROBE checklist (cohort studies)}

Item topics below are abbreviated paraphrases; the official STROBE recommendation wording is
copyrighted and is available from \texttt{strobe-statement.org}, from which the completed
checklist form is submitted as a separate reporting item. Entries marked ``Not reported'' are
genuine gaps and are stated as such rather than mapped to a proxy location.

\begin{longtable}{>{\raggedright\arraybackslash}p{0.9cm} >{\raggedright\arraybackslash}p{4.0cm} >{\raggedright\arraybackslash}p{9.2cm}}
\caption{STROBE cohort-study checklist and where each item is addressed in this manuscript. ``Methods'' and ``Results'' refer to main-text sections; ``Supplementary'' refers to this document.}
\label{tab:strobe} \\
\toprule
\textbf{No.} & \textbf{Item (abbreviated)} & \textbf{Where reported / status} \\
\midrule
\endfirsthead
\multicolumn{3}{c}{{\bfseries \tablename\ \thetable{} -- continued from previous page}} \\
\toprule
\textbf{No.} & \textbf{Item (abbreviated)} & \textbf{Where reported / status} \\
\midrule
\endhead
\midrule
\multicolumn{3}{r}{{Continued on next page}} \\
\endfoot
\bottomrule
\endlastfoot

\multicolumn{3}{l}{\textbf{Title and abstract}} \\
\midrule
1(a) & Study design indicated in title or abstract & Abstract (``two retrospective cohorts''); Methods, ``Clinical datasets'' $\rightarrow$ ``Study design'' (``a retrospective observational cohort analysis''). The title carries no design term. \\
1(b) & Informative and balanced abstract summary & Abstract: setting, both cohorts with sample sizes, method, effect sizes, and the explicit ``hypothesis-generating pending external validation'' caveat. \\
\midrule
\multicolumn{3}{l}{\textbf{Introduction}} \\
\midrule
2 & Background and rationale & Introduction, paragraphs 1--4 (clinical rationale for pressure injury and length of stay after SCI; causal discovery under missingness; integration of partially overlapping cohorts). \\
3 & Objectives & Introduction, five enumerated contributions (final paragraph); Results, ``Markov-blanket graph learning through multi-cohort fusion''. No hypotheses were prespecified and no analysis protocol was pre-registered. \\
\midrule
\multicolumn{3}{l}{\textbf{Methods}} \\
\midrule
4 & Study design, presented early & Methods, ``Clinical datasets'' (opening design statement) and Methods, ``Statistics and reproducibility'' (``The studies were observational and retrospective\ldots''). \\
5 & Setting, locations, and relevant dates & \emph{Partially reported.} Setting: Methods, ``Clinical datasets'' and ``Ethical approval'' (single center, Swiss Paraplegic Center, Nottwil). PPIS admissions span approximately 2016--2022 (Methods, ``Clinical datasets''). The HAPI recruitment window, the database extraction/lock dates, and the observation window are \textbf{Not reported}. \\
6(a) & Eligibility criteria, sources and methods of selection, follow-up methods & \emph{Partially reported.} Cohort composition: Methods, ``Clinical datasets'' (PPIS 266 episodes, 126 variables; HAPI 94\,$+$\,48\,$+$\,108 records) and ``Cohort filtering and HAPI subscore handling''. Methods, ``Statistics and reproducibility'' states that both cohorts comprise all eligible admissions in the source registries and that no participant-level exclusions were applied. Explicit inclusion/exclusion criteria, the registry query used to identify records, the defining rule for the three HAPI sub-cohorts, and the follow-up method are \textbf{Not reported}. \\
6(b) & Matching criteria and numbers exposed/unexposed & \textbf{Not applicable} (unmatched cohorts; no exposed/unexposed contrast is defined). \\
7 & Outcomes, exposures, predictors, confounders, effect modifiers; diagnostic criteria & \emph{Partially reported.} Complete variable inventories with type, coding and units: Supplementary Tables~\ref{tab:ppis_complete}, \ref{tab:hapi_variables}, \ref{tab:shared_backbone}. Temporal tiering of predictors: Methods, ``Method architecture''. Confounder handling is structural (m-separation-restricted total-conditioning Markov blanket) rather than by covariate adjustment: Methods, ``Evaluation metrics'' (``Constrained constrained graph set for binary outcomes''). One outcome is defined operationally in Results (PI onset $=[\text{PI stage}\ge1]$, reported and rejected as confounded). Diagnostic/ascertainment criteria for UTI, osteomyelitis and secondary PI complication (beyond binary coding) and the admission/discharge anchors for length of stay are \textbf{Not reported}. \\
8 & Data sources and measurement; comparability across groups & \emph{Partially reported.} Sources and harmonisation: Methods, ``Clinical datasets'' (``Variable harmonization and overlap structure''), Table~\ref{tab:clinical_datasets}, Supplementary Table~\ref{tab:shared_backbone}; the known Age-anchor mismatch (HAPI age at SCI vs.\ PPIS age at admission) is stated in Discussion and quantified in Supplementary Table~\ref{tab:stability}. Chart-abstraction route, abstractor training or double coding, and laboratory assay platforms for albumin and haemoglobin are \textbf{Not reported}. \\
9 & Efforts to address sources of bias & Analytic bias and leakage are addressed throughout: exclusion of post-event and identifier variables (Methods, ``Cohort filtering and HAPI subscore handling''); temporal-tier restriction of predictors (Methods, ``Evaluation metrics''); rejection of an endpoint driven by differential missingness and correction of antibiotic leakage on osteomyelitis (Results); post-discovery selection optimism and the nested cross-validation correction, and patient-grouped folds (Methods, ``Downstream prediction setup''; Supplementary Tables~\ref{tab:nestedcv}, \ref{tab:groupcv}); steering negative controls (Supplementary Table~\ref{tab:negcontrol}). Single-center selection bias and outcome misclassification are named in Discussion (limitations 1, 5, 8) but not quantified. \\
10 & Study size & Methods, ``Statistics and reproducibility'' (no sample-size calculation; all eligible admissions; HAPI $n=250$, PPIS $n=266$ episodes from 174 patients); Supplementary Table~\ref{tab:tripod} (Sample size row); Discussion, ninth limitation (underpowered for high-dimensional structure learning). \\
11 & Handling of quantitative variables & Methods, ``Clinical datasets'' (``Preprocessing'': $>80\%$ missingness screen, standardization, winsorization at the 1st/99th percentiles, ordinal/one-hot encoding, no further feature engineering); Methods, ``Cohort filtering and HAPI subscore handling'' (stricter $>40\%$ cut for PPIS); Methods, ``Implementation details'' (integer encoding for the Fisher-$z$ test, with a kernel conditional-independence re-run); Methods, ``Evaluation metrics'' (log transform of length of stay). \\
12(a) & All statistical methods, including confounding control & Methods, ``Pipeline architecture'' and Algorithm~1; ``Implementation details''; ``Evaluation metrics''; ``Downstream prediction setup''; ``PAG-guided attention steering''; ``Statistics and reproducibility'' (paired Wilcoxon with Holm correction; paired bootstrap; 300-resample bootstrap confidence intervals; explicit statement that no multiplicity correction is applied across the strategy\,$\times$\,mode\,$\times$\,target grid). Confounding is controlled structurally by conditioning on the Markov blanket, not by regression adjustment. \\
12(b) & Subgroup and interaction analyses & \textbf{Not performed.} Stated in Supplementary Table~\ref{tab:tripod} (Fairness/subgroups row); subgroup performance by sex and AIS is identified as future work. \\
12(c) & Handling of missing data & Methods, ``Pipeline architecture'' and ``Implementation details'' (dtype-aware initialisation; Markov-blanket-guided random-forest re-imputation inside the discovery loop); Results, ``Why not skip imputation?''; Supplementary Tables~\ref{tab:missingness} and~\ref{tab:imputation}. \\
12(d) & Loss to follow-up & \textbf{Not applicable / not separately reported}: follow-up ends at hospital discharge and no post-discharge follow-up was available; stated in Methods, ``Clinical datasets''. \\
12(e) & Sensitivity analyses & Supplementary ``Validation and robustness analyses'' (Tables~\ref{tab:decision_time}, \ref{tab:groupcv}, \ref{tab:nestedcv}, \ref{tab:calibration}, \ref{tab:negcontrol}, \ref{tab:imputation}) and ``Significance-threshold ($\alpha$) sensitivity'' (Tables~\ref{tab:alpha_sensitivity}, \ref{tab:alpha_trajectory}, \ref{tab:stability}, \ref{tab:sl_sensitivity}, \ref{tab:constraint_ablation}); bootstrap edge uncertainty and the fail-closed gate (Table~\ref{tab:prob_fusion}). \\
\midrule
\multicolumn{3}{l}{\textbf{Results}} \\
\midrule
13(a--c) & Numbers at each stage; reasons for non-participation; flow diagram & \textbf{Not reported.} Analysed numbers are given (HAPI $n=250$ as $94+48+108$ records, Methods; PPIS $n=266$ episodes from 174 patients, Methods, ``Downstream prediction setup''), and Methods, ``Statistics and reproducibility'' states that no participant-level exclusions were applied. No screened/eligible/excluded counts and no participant-flow diagram are provided; Figure~1 is a methodological pipeline, not a participant flow. \\
14(a) & Participant characteristics and covariate information & Supplementary Table~\ref{tab:table1} (cohort characteristics: median [IQR] for continuous, $n$ (\%) for categorical; age, sex, AIS grade, albumin, haemoglobin, SCIM, pneumonia, length of stay), with a footnote flagging the differing age anchors. \\
14(b) & Number of participants with missing data per variable & \emph{Partially reported.} Supplementary Table~\ref{tab:missingness} gives per-variable missingness for the nine shared-backbone variables plus cohort-level aggregates. Per-variable missingness for the remaining 43 HAPI and 87 PPIS candidate features is \textbf{Not reported}. \\
14(c) & Follow-up time & \textbf{Not reported as follow-up time.} Length of stay is reported as an outcome (Supplementary Table~\ref{tab:table1}: 66 [19, 170] days HAPI; 89 [69, 122] days PPIS) and coincides with the observation window, but is not stated as such and total person-time is not given. \\
15 & Outcome events or summary measures over time & Results, ``PAG-guided attention steering for the PPIS binary post-surgical outcomes'' (UTI $n_{+}=77/266$; osteomyelitis $n_{+}=114/266$; secondary PI complication $n_{+}=62/266$); Supplementary Table~\ref{tab:table1} (same counts with prevalences, plus length of stay as median [IQR]). \\
16(a) & Unadjusted and confounder-adjusted estimates with precision & \textbf{Not applicable in the classical form}: no exposure--outcome association estimates are reported. Structural results are PAG edges with bootstrap inclusion posteriors and 95\% credible intervals (Supplementary Tables~\ref{tab:prob_fusion}, \ref{tab:edge_stability}); predictive results are reported with 95\% bootstrap confidence intervals from 300 resamples (Methods, ``Evaluation metrics''; Results). Confounding is handled by conditioning on the m-separation-restricted Markov blanket (Methods, ``Evaluation metrics''). \\
16(b) & Category boundaries when continuous variables are categorised & \emph{Partially applicable.} Winsorisation percentiles and AIS grades are reported (Methods, ``Clinical datasets''; Supplementary Table~\ref{tab:table1}). The only dichotomisation used, PI onset $=[\text{PI stage}\ge1]$, is defined in Results. No other continuous predictor is categorised. \\
16(c) & Translation of relative to absolute risk & \textbf{Not applicable} (no relative risks are estimated). \\
17 & Other analyses (subgroups, interactions, sensitivity) & Supplementary ``Validation and robustness analyses'', ``Significance-threshold ($\alpha$) sensitivity'', and ``Steering Validation''; Results (rejected PI-onset endpoint with its reason for rejection). Subgroup and interaction analyses were not performed (item 12(b)). \\
\midrule
\multicolumn{3}{l}{\textbf{Discussion}} \\
\midrule
18 & Key results with reference to objectives & Discussion, opening paragraph (structural recovery on synthetic data; imputation error and distribution preservation; the three clinical result types), including the explicit statement that the constrained graph sets are superior on one of five targets and otherwise competitive. \\
19 & Limitations, with direction and magnitude of potential bias & Discussion, nine enumerated limitations (cohort stage heterogeneity; mixed variable semantics; no ground-truth graph; $\alpha$ and sampling sensitivity; temporal rather than prospective validity, with the $R^2=0.71$ vs.\ $0.31$--$0.34$ contrast; 266 episodes from 174 patients with a $0.04$--$0.12$ metric drop under patient grouping; graph-selection optimism; single centre; underpowered high-dimensional PAG learning). Harmonisation defects are quantified separately (Discussion; Supplementary Table~\ref{tab:stability}). \\
20 & Cautious overall interpretation & Discussion (literature triangulation for the three shared determinants and the complication blankets; repeated ``hypothesis-generating'' framing; multiplicity caveat; closing paragraph separating what the study does and does not establish). \\
21 & Generalisability & Discussion, eighth limitation and closing sentence (single center; cross-stage rather than cross-site transfer; external multi-site validation essential); Supplementary Table~\ref{tab:tripod} (Validation row). \\
\midrule
\multicolumn{3}{l}{\textbf{Other information}} \\
\midrule
22 & Funding, and the role of funders & Acknowledgements (Schweizer Paraplegiker Stiftung and ETH Zurich Foundation, grant 2021-HS-348; funders played no role in design, analysis, interpretation, or the decision to publish). Funding of the original HAPI and PPIS data collection, if different, is \textbf{Not separately reported}. \\

\end{longtable}

\section{Variable Definitions}

\begin{table}[ht]
\centering
\caption{Harmonized variable concepts across HAPI and PPIS. Of these 12 harmonized concepts, nine are observed in both cohorts; FIM Cognitive (FIMc), Nutrition, and Autonomic dysreflexia are available in HAPI only. The pooled-shared PAG for length of stay ($n=516$) is estimated on the subset available at or before admission: SCIM is recorded at discharge in PPIS (an end-of-stay measurement) and is therefore excluded from the LoS backbone by the temporal-tier constraint, leaving eight shared predictors (AIS, Age, Level of injury, Sex, Albumin, Hemoglobin, Vitamin D, Pneumonia) plus length of stay (nine shared variables in total). The harmonized ``Age'' concept maps HAPI \emph{Age at SCI} to PPIS \emph{Age at Admission}; these are related but not identical quantities (see Discussion). Rows shaded in \colorbox{blue!10}{light blue} in the PPIS and HAPI tables below correspond to the shared concepts.}
\label{tab:shared_backbone}
\small
\begin{tabularx}{\linewidth}{>{\raggedright\arraybackslash}p{3.2cm} >{\raggedright\arraybackslash}p{3.8cm} >{\raggedright\arraybackslash}p{3.8cm} X}
\toprule
\textbf{Shared concept} & \textbf{HAPI column} & \textbf{PPIS column} & \textbf{Type / units} \\
\midrule
Age & Age at SCI & Age at Admission & Continuous, years \\
AIS & AIS & AIS ISNCSCI at entry & Ordinal, A--E \\
Level of injury & Level of Injury & Lesion Level & Categorical \\
Sex & Sex & Sex & Binary \\
SCIM & SCIM Score & SCIM Total Score at discharge & Continuous, 0--100 \\
FIM Cognitive (FIMc) & FIM Cognitive Score & --- & Continuous \\
Nutrition & Nutrition Status & --- & Ordinal \\
Albumin & Albumin & Albumin & Continuous, g/L \\
Hemoglobin & Hemoglobin & Hemoglobin & Continuous, g/L \\
Autonomic dysreflexia & Autonomic Dysreflexia & --- & Binary \\
Pneumonia & Pneumonia & Pneumonia (During Stay) & Binary \\
Length of stay & Length of Stay & Duration of Hospitalization & Continuous, days \\
\bottomrule
\end{tabularx}
\end{table}

\begin{table}[ht]
\centering
\caption{\textbf{Per-cohort support for the three pooled-shared edges into length of stay.} \emph{Shared backbone}: each cohort run alone on the variables observed in both cohorts, with the same temporal-tier knowledge, depth and $\alpha=0.05$ as the pooled analysis; $\rightarrow$ marks a definite directed edge into LoS, and ``at $\alpha=0.10$'' an edge that appears only at the looser threshold. Each cohort alone orients part of the set (HAPI the Age and Hemoglobin relations, PPIS the Albumin relation) and the pooled analysis orients all three; HAPI alone also orients AIS and level of injury into LoS, which the pooled analysis does not. \emph{Full per-cohort graph}: bootstrap inclusion probability ($B=100$, unconstrained, $\alpha=0.10$) of the same edges when each cohort is searched over its whole variable set, where the three relations are screened off almost entirely. Fusing the pooled edges into the per-cohort graphs is what places them in the per-cohort Markov blankets (Table~\ref{tab:constraint_ablation}). Pooled $p_{\mathrm{edge}}$ is bootstrap inclusion in the pooled graph from the same resampling. Values from \texttt{percohort\_backbone.json} and \texttt{probabilistic\_mb\_sensitivity.json}.}
\label{tab:edge_stability}
\small
\begin{tabularx}{\linewidth}{X c c c c c}
\toprule
 & & \multicolumn{2}{c}{Shared backbone} & \multicolumn{2}{c}{Full per-cohort graph} \\
\cmidrule(lr){3-4}\cmidrule(lr){5-6}
Pooled-shared edge & Pooled $p_{\mathrm{edge}}$ & HAPI & PPIS & HAPI $p_{\mathrm{edge}}$ & PPIS $p_{\mathrm{edge}}$ \\
\midrule
Age $\rightarrow$ LoS        & 0.98 & $\rightarrow$ & --- & 0.00 & 0.00 \\
Albumin $\rightarrow$ LoS    & 0.76 & at $\alpha=0.10$ & $\rightarrow$ & 0.00 & 0.00 \\
Hemoglobin $\rightarrow$ LoS & 1.00 & $\rightarrow$ & at $\alpha=0.10$ & 0.17 & 0.01 \\
\bottomrule
\end{tabularx}
\end{table}

\subsection*{Edge-uncertainty results and the fail-closed gate}

The deployed pipeline learns each partial ancestral graph once and thresholds every edge to a hard present/absent mark. Because the FCI conditional-independence decisions carry sampling uncertainty, which is acute at these cohort sizes, we additionally use a nonparametric bootstrap to quantify how reliably each recovered structure survives resampling. The resulting edge posteriors grade Markov-blanket membership and gate the cross-cohort constraints. We drew $B=100$ bootstrap resamples of each cohort and of the pooled-shared backbone, re-ran depth-3 FCI ($\alpha=0.10$) on each, and recorded, for every variable pair, the adjacency frequency $\hat P(\mathrm{edge})$ and, for every length-of-stay predictor, the frequency $\hat P(v\in\mathrm{MB})$ with which it entered the depth-2 available-causal Markov blanket. The latter is the correct membership confidence. A depth-2 blanket member need not be a direct neighbour of the target, so an edge-level threshold would misjudge it.

\paragraph{Uncertainty-aware Markov blankets.} The pooled-shared HAPI length-of-stay fusion blanket (11 members) underlies the constraint ablation and fail-closed gate and is distinct from the larger deployed benchmark blankets of HAPI~21 and PPIS~34. Resampling fully corroborates it. All 11 members enter the bootstrap blanket in at least half of resamples, and all at $\hat P(v\in\mathrm{MB})\ge0.7$. The larger PPIS pooled blanket (35 members) is less stable, with 24 robust ($\hat P\ge0.5$) and 11 marginal members. This quantifies the greater $\alpha$-sensitivity of the denser PPIS graph that Table~\ref{tab:stability} shows across thresholds. A probabilistic blanket that keeps only members above a confidence $\tau$ recovers this blanket's robust core and trims exactly the marginal members as $\tau$ rises, so the hard-threshold blanket is the moderate-confidence operating point of a continuum.

\paragraph{Fail-closed cross-cohort constraints.} The same bootstrap also motivates the pooled-shared design. In the full per-cohort graphs (44 and 88 nodes, i.e.\ the 43 and 87 candidate features plus the length-of-stay target), intermediate variables screen off the direct Age$\rightarrow$LoS, Albumin$\rightarrow$LoS and Hemoglobin$\rightarrow$LoS edges, which are recovered in few resamples ($\hat P\le0.17$). The compact nine-variable pooled backbone (eight shared predictors plus the length-of-stay target) recovers them at $0.98$, $0.76$ and $1.00$. For this reason the cross-cohort claim must be learned on the reduced shared backbone and injected back into each per-cohort graph. We place a Beta posterior (Jeffreys prior) on the pooled-backbone frequencies and impose a constraint only when its posterior mean clears a confidence gate of $0.8$. This \emph{fail-closed} rule, which does not assert an uncertain edge, imposes Age (posterior $0.975$) and Hemoglobin ($0.995$) but withholds Albumin ($0.757$; Table~\ref{tab:prob_fusion}). The gate independently reproduces, from edge-level uncertainty alone, the finding that Albumin$\rightarrow$LoS is the one shared edge not robustly recovered (albumin is missing in ${\sim}19$--$37\%$ of records; Table~\ref{tab:edge_stability} and its age-harmonization companion). The fail-closed rule thus turns the qualitative stability caveat into an automatic, calibrated gate on which cross-cohort constraints the fusion is permitted to enforce, without touching the deployed hard-threshold results.

\begin{table}[ht]
\centering
\caption{Fail-closed cross-cohort constraint gate. Bootstrap adjacency frequency of each pooled-shared length-of-stay edge in the full per-cohort graphs (where it is screened off) versus the reduced pooled backbone, with the Beta (Jeffreys-prior) posterior mean and 95\% credible interval on the pooled-backbone frequency ($B=100$). A constraint is imposed only if the posterior mean clears the $0.8$ confidence gate; Albumin is withheld, matching Table~\ref{tab:edge_stability}.}
\label{tab:prob_fusion}
\small
\begin{tabular}{lccccc}
\toprule
Shared edge & HAPI full & PPIS full & Pooled backbone & Posterior [95\% CI] & Gate (0.8) \\
\midrule
Age $\rightarrow$ LoS        & ${\approx}0$ & ${\approx}0$ & $0.98$ & $0.975$ $[0.94, 1.00]$ & \textbf{impose} \\
Hemoglobin $\rightarrow$ LoS & $0.17$ & $0.01$ & $1.00$ & $0.995$ $[0.98, 1.00]$ & \textbf{impose} \\
Albumin $\rightarrow$ LoS    & ${\approx}0$ & ${\approx}0$ & $0.76$ & $0.757$ $[0.67, 0.84]$ & withhold \\
\bottomrule
\end{tabular}
\end{table}

\paragraph{Prediction consequence of the fail-closed gate.} Withholding an uncertain constraint has a cohort-dependent cost. We re-ran the constraint ablation (Table~\ref{tab:constraint_ablation}) under the fail-closed gate, imposing Age and Hemoglobin and leaving Albumin unconstrained. On HAPI this forfeits the entire pooled-fusion gain, with length-of-stay $R^2$ falling from $0.718$ to $0.695$ (the no-pooled baseline), because the compact 11-feature HAPI blanket rests on the one edge the gate withholds. On PPIS the same gate costs essentially nothing ($0.735\rightarrow0.734$). Its 23-feature blanket carries the signal redundantly, so dropping Albumin is immaterial. The most uncertain shared edge is therefore also the most predictively valuable on the data-poor cohort, creating a safety--accuracy tension. We accordingly present the fail-closed gate as a deployment option rather than a strictly better default. It offers a calibrated way to trade a small-cohort gain for protection against over-asserting a fragile edge. The primary results use the full pooled fusion, and we report both.

\begin{table}[ht]
\centering
\caption{Cohort characteristics. Continuous variables are median [IQR]; categorical variables are $n$ (\%). HAPI ($n=250$) pools the three rehabilitation/comparison cohorts; PPIS ($n=266$) is the post-pressure-injury surgery cohort. $^a$HAPI age is age at spinal cord injury, recorded for the 142 rehabilitation-cohort patients but not for the No-HAPI comparison cohort (hence 43\% missing; Table~\ref{tab:missingness}). PPIS age is age at admission, a different anchor (Table~\ref{tab:shared_backbone}). $^b$AIS: ASIA Impairment Scale grade at entry (A = motor/sensory complete). The three binary post-surgical outcomes are defined only for PPIS.}
\label{tab:table1}
\small
\begin{tabularx}{\linewidth}{l X X}
\toprule
Characteristic & HAPI ($n=250$) & PPIS ($n=266$) \\
\midrule
Age, years$^a$ & 50 [34, 66] & 60 [51, 70] \\
Female sex & 72 (29\%) & 64 (24\%) \\
AIS grade A / B / C / D$^b$ & 90 (36\%) / 29 (12\%) / 37 (15\%) / 90 (36\%) & 203 (76\%) / 37 (14\%) / 18 (7\%) / 8 (3\%) \\
Length of stay, days & 66 [19, 170] & 89 [69, 122] \\
Albumin, g/L & 33 [29, 39] & 35 [32, 39] \\
Hemoglobin, g/L & 122 [100, 140] & 120 [106, 136] \\
SCIM (0--100) & 31 [17, 52] & 36 [25, 51] \\
Pneumonia & 41 (16\%) & 19 (7\%) \\
UTI during stay & --- & 77 (29\%) \\
Osteomyelitis & --- & 114 (43\%) \\
Secondary PI complication & --- & 62 (23\%) \\
\bottomrule
\end{tabularx}
\end{table}

\begin{table}[ht]
\centering
\caption{Per-variable missingness (\%) by cohort for the shared-backbone variables. Overall missingness is 1.5\% in the HAPI rehabilitation cohort, 11.4\% in the No-HAPI comparison cohort, and 21.5\% in PPIS (median 33.5\% among variables with any missingness). Several variables (discharge SCIM, laboratory markers) appear more likely to be missing at extreme or clinically-uneventful values, i.e.\ plausibly non-random missingness, motivating the imputation-aware discovery pipeline.}
\label{tab:missingness}
\small
\begin{tabularx}{\linewidth}{l c c}
\toprule
Variable & HAPI missing (\%) & PPIS missing (\%) \\
\midrule
Age & 43 & 0 \\
Sex & 0 & 0 \\
AIS & 0 & 0 \\
Level of injury & 0 & 0 \\
Length of stay & 0 & 0 \\
Albumin & 37 & 19 \\
Hemoglobin & 2 & 0 \\
SCIM & 8 & 35 \\
Pneumonia & 0 & 0 \\
\bottomrule
\end{tabularx}
\end{table}

\subsection{PPIS Dataset Variables}
\begin{longtable}{
>{\raggedright\arraybackslash}p{4cm} 
>{\raggedright\arraybackslash}p{2cm} 
>{\raggedright\arraybackslash}p{5.5cm} 
>{\raggedright\arraybackslash}p{4.5cm}}
\caption{PPIS dataset variable definitions (abbreviated; derived duration variables such as Duration of Immobilization and Time between Surgeries are defined in Table~\ref{tab:mb_features}). \colorbox{blue!10}{Shaded rows} are shared backbone variables also present in HAPI (see Table~\ref{tab:shared_backbone}).} \label{tab:ppis_complete} \\

\toprule
\textbf{Variable} & \textbf{Type} & \textbf{Description} & \textbf{Encoding / Units} \\
\midrule
\endfirsthead

\multicolumn{4}{c}%
{{\bfseries \tablename\ \thetable{} -- continued from previous page}} \\
\toprule
\textbf{Variable} & \textbf{Type} & \textbf{Description} & \textbf{Encoding / Units} \\
\midrule
\endhead

\midrule
\multicolumn{4}{r}{{Continued on next page}} \\
\endfoot

\bottomrule
\endlastfoot

\textbf{Sociodemographic and Definition of SCI Lesion} & & & \\
\midrule

PID & Text & patient ID & - \\
FID & Text & case ID & - \\
explicit consent & Binary & - & 0 no, 1 yes \\
Date of entry & Date & Date of admission at the hospital. Not to be changed later. & Any valid date \\
Date of discharge & Date & Date of discharge from the hospital. & Any valid date, 0 still in hospitalization/ unknown \\
\rowcolor{blue!8}Duration of Hospitalization & Numeric & - & In Days \\
Any ICU stay during rehabilitation & Binary & - & 0 no, 1 yes \\
Pressure injury at admission & Binary & - & 0 no, 1 yes \\
\rowcolor{blue!8}Sex & Categorical & - & 0 Male, 1 Female \\
\rowcolor{blue!8}Year of birth & Date & Date of birth (Age proxy) & Any valid date \\
Date of first diagnosis of SCI & Date & Date when SCI occurred or was first diagnosed (traumatic or non-traumatic) & Any valid date \\
Aetiology of SCI & Categorical & Documents aetiology of the spinal cord injury. The first applicable answer will be used. Traumatic SCI: impairment of the spinal cord or cauda equina... & 1 Transport activity, 2 Sports and leisure activity, 3 Fall, 4 Violence, 5 Other accident cause... \\
Years post injury & Text & - & - \\
\rowcolor{blue!8}Lesion Level & Categorical & SwiSCI Codebook 2021 & 1 C1, 2 C2, 3 C3, 4 C4, 5 C5, 6 C6, 7 C7, 8 C8, 9 Th1, 10 Th2, 11 Th3, 12 Th4, 13 Th5, 14 Th6... \\
ISNCSCI Functioning subgroups & Categorical & DeVivo, 2011 & 1 C1-4 A B C, 2 C5-8 A B C, 3 Th1-S5 A B C, 4 D at any level, 5 Ventilator dependent... \\
\rowcolor{blue!8}AIS ISNCSCI at the time of entry & Categorical & Documents the ASIA impairment scale (AIS) score if assessed during care (at the time of admission to rehabilitation) & 1 A, 2 B, 3 C, 4 D, 5 E, 6 Unknown \\
Social status at the time of entry & Categorical & Documents the marital status at the time of SCI. "Unmarried" (annulled marriages, etc.) should be coded as "Other". & 1 Single, 2 Married, 3 Widowed, 4 Divorced, 5 Partnership, 6 Other, 8 Unknown \\
Education & Categorical & Highest completed education level & 1 no education/graduation, 2 secondary-school graduation, 3 higher/tertiary (abbreviated) \\
Employment & Categorical & Employment status / working-time category & 1 unemployed, 2 retired, 3 employed $<50\%$, 4 employed $\ge 50\%$ (abbreviated) \\
Type of residence before admission & Categorical & Type of residence at time of SCI, Private residence: includes house, condominium, mobile home, apartment, houseboat, Hospital... & 1 Private residence, 2 Private residence and spitex, 3 Hospital, 4 Nursing home, 5 Assisted living... \\
Current smoking & Binary & Anamnesis at entry & 0 No, 1 Yes, 3 Unknown \\
Alcohol consumption & Binary & Anamnesis at entry & 0 No, 1 Yes, 3 Unknown \\
Rate of Alcohol Consumption & Numeric & Anamnesis at entry & 1 Once a month or less, 2 Between once a week and once a month, 3 Several times per week, 4 Daily... \\
Substance Abuse & Binary & Anamnesis at entry & 0 No, 1 Yes \\

\midrule
\midrule
\textbf{PI Characteristics} & & & \\
\midrule

Pressure Injury (PI) Stage & Ordinal & Stages 1--4 & 1 Stage 1, 2 Stage 2, 3 Stage 3, 4 Stage 4, 5 Unknown \\
Localization of PI & Text & - & 1 Sacrum/Coccyx, 2 Ischium left, 3 Ischium right, 4 Trochanter left, 5 Trochanter right \\
Number of (deep) PI over sacrum, trochanter, ischium & Text & Number at entry & 1=1, 2=2, 3=3, 4=4, 5=5 \\
Recurrence & Text & At entry & 0 No recurrence, 1 Recurrence at same location, 2 Earlier surgically treated PI at other location \\
Osteomyelitis (OM) & Binary & During stay; analysed as a post-surgical \emph{outcome} (temporal tier 3) & 0 No, 1 Yes \\
Type of OM & Text & During stay (tier 3); recorded only when OM is present & 0 Acute, 1 Chronic, 2 unknown \\
Previous OM & Binary & At entry; history of osteomyelitis prior to admission (tier 1) & 0 No, 1 Yes \\
Type of Bacteria & Categorical & Elsas 2021 & 1 Acinetobacter baumanii, 2 Bacteroides fragilis, 3 Bacteroides spp, 4 Bacteroides stercoris... \\
Number of Bacteria & Text & Elsas 2021, during debridement and flap surgery & 1, 2, 3, 4, 5, 6, 7, 8, 9 \\

\midrule
\midrule
\textbf{Drugs} & & & \\
\midrule
Antibiotics (AB) & Binary & - & 0 No, 1 Yes, 3 Unknown \\
Administration of Antibiotics & Numeric & Any antibiotic therapy during the rehabilitation & 1=Only i.v., 2=Only p.o., 3=i.v. followed by p.o., 4=i.v. and p.o. at same time, 5 i.v, p.o., i.v... \\
Duration of AB & Numeric & - & Number of Days \\
Start of AB (Total) & Text & - & dd.mm.yyyy \\
End of AB (Total) & Text & - & dd.mm.yyyy \\
AB before flap surgery & Binary & - & 0 No, 1 Yes \\
AB according to concept & Binary & 1 week max. (7d) without OM, 6--8 weeks (42--56d) with OM & 0 No, 1 Yes \\
AB for other acute condition & Binary & - & 0 No, 1 Yes \\
Type of AB & Categorical & Elsas 2021 & 1 Amoxicillin/ clavulanic acid, 2 Amoxicillin, 3 Cefazolin, 4 Ceftazidim, 5 Ceftriaxon, 6 Cefurox... \\
Number of AB & Categorical & Elsas 2021 & 1, 2, 3, 4, 5, 6, 7, 8, 9 \\

\midrule
\midrule
\textbf{Laboratory parameters} & & & \\
\midrule
\rowcolor{blue!8}Hemoglobin & Numeric & Value at the time of entry in g/l. Normal range: 140-170 & Lab value \\
\rowcolor{blue!8}Albumin & Numeric & Value at the time of entry in g/l. Normal range: 32-50 & Lab value \\
Protein total & Numeric & Value at the time of entry in g/l. Normal range: 64-83 & Lab value \\
25-OH-Vitamin D & Numeric & Value at the time of entry in nmol/l, Normal range: $>$=75 & Lab value \\
CRP & Numeric & Value at the time of entry in mg/l. Normal range: $<$5. & Lab value \\
Sodium & Numeric & Value at the time of entry in mmol/l. Normal range: 135-145. & Lab value \\
Potassium & Numeric & Value at the time of entry in mmol/l. Normal range 3.5-5.1 & Lab value \\
Iron & Numeric & Value at the time of entry in umol/l. Normal range 8.8-27. & Lab value \\
Ferritin & Numeric & Value at the time of entry in ug/l. Normal range 15-150. & Lab value \\
Transferrin & Numeric & Value at the time of entry in g/l. Normal range 2.2-3.7 & Lab value \\
ESR & Numeric & Value at the time of entry in mm/h Normal range: 10 - 20mm/h & Lab value \\
Creatinine & Numeric & Value at the time of entry in umol/l. Normal range 59-104. & Lab value \\
Cystatin C & Numeric & Value at the time of entry in mg/l. Normal range 0,61-0,95. & Lab value \\
GFR (renal disease) & Numeric & Value at the time of entry in ml/min. Normal range: $>$=90 & Lab value \\
Any poor diabetes control & Numeric & Normal range 3.6 to 5.8 mmol/l & Lab value in mmol/l or "no abnormal level of glucose" \\
HbA1c & Numeric & Value at time of entry, Norm: 4.8-5.9\% & Percentage \\
Calcium at entry & Numeric & Norm values: 2.2-2.6 mmol/L & Lab value \\
LDL-Cholesterin at entry & Numeric & 0.2-4.0mmol/L & Lab value \\
HDL-Cholesterin at entry & Numeric & 0.9-2.0mmol/L & Lab value \\
Total Cholesterin & Numeric & $<$5.2mmol/L & Lab value \\
Triglyceride at entry & Numeric & 0.1-2.3mmol/L & Lab value \\
Vitamin B12 at entry & Numeric & Norm values: 200-1000ng/L & Lab value \\
Acidum folicum at entry & Numeric & Norm values: 5 to 20 $\mu$g/L & Lab value \\
Prealbumin at entry & Numeric & Norm values: 0.18–0.45 g/l & Lab value \\
Zinc at entry & Numeric & Norm values: 66–110 mcg/dl & Lab value \\
TSH at entry & Numeric & Norm values: 0.4-4.0 mU/L & Lab value \\
INR at entry & Numeric & Norm values: 0.9-1.3 & Lab value \\

\midrule
\midrule
\textbf{Comorbidities \& Previous Surgeries} & & & \\
\midrule
Deep PI at other location & Binary & Present at entry & 0 no, 1 Yes \\
Diabetes mellitus (DM) & Binary & All types of DM & 0 no, 1 Yes \\
Psychiatric Diagnosis & Binary & Any psychiatric diagnosis BUT dementia & 0 no, 1 yes \\
Dementia & Binary & - & 0 no, 1 yes \\
Malignant tumor/ neoplasia & Numeric & Present at entry & 0 no, 2 yes without metastasis, 3 with metastasis, 4 yes unknown stage \\
Obstructive Sleep Apnea Syndrome (OSAS) & Numeric & - & 0 no, 2 yes with cPAP therapy, 3 without cPAP therapy, 4 yes but unknown therapy \\
Mechanical ventilation & Numeric & - & 0 no, 2 non invasive, 3 invasive/intubated, 4 unknown \\
\rowcolor{blue!8}Pneumonia (During Stay) & Binary & - & 0 no, 1 yes \\
Gastroenteritis (During Stay) & Binary & - & 0 no, 1 yes \\
Urinary tract infection & Binary & During stay; analysed as an \emph{outcome} (tier 3; \texttt{UTI During Stay}) & 0 no, 1 yes \\
Congestive heart failure currently decompensated & Binary & - & 0 no, 1 yes \\
Vascular comorbidity (pAVK, CVI) & Binary & peripheral arterial occlusive disease (pAVK), chronic venous insufficiency (CVI) & 0 no, 1 yes \\
Hypertony & Binary & - & 0 no, 1 yes \\
Rheumatoid arthritis & Binary & - & 0 no, 1 yes \\
Osteoarthritis & Binary & - & 0 no, 1 yes \\
Scoliosis & Binary & - & 0 no, 1 yes \\
Amputation(s) & Binary & - & 0 no, 1 yes \\

\midrule
\midrule
\textbf{Measurement of Body Functions} & & & \\
\midrule
Bloodpressure, systolic & Numeric & Normal range 110-139mmhg & systolic value in mm Hg at the time of entry \\
Any hyper- or hypotensive bloodpressure episode & Numeric & clinically relevant & 1 no, 2 hypertensive episode (syst BP $>$ 180mmhg), 3 hypotensive episode... \\
Any episode of hypoxemia during the rehabilitation & Numeric & clinically relevant & 1 no, 2 yes SO2 $<$92\%, 3 yes SO2 $<$90\%, 4 yes SO2 $<$85\%, 5 yes SO2 $<$80\%, 6 unknown \\
Abnormal body temperature & Numeric & clinically relevant & 1 no (temp between 36.5 and 37.5°C), 2 yes hypothermia (35 -36.4), 3 yes severe hypoth... \\
Body height & Numeric & Body height at the time of entry & Value in cm \\
Body weight & Numeric & Body weight at time of entry & Value in Kilograms \\
Weight change during the rehabilitation & Numeric & - & 1 no change, 2 Weight drop between 5 and 10\% of entry weight, 3 Weight drop more than 10\%... \\
Spasticity & Numeric & ASWORTH Classification 1-4 & 1 no spasticity, 2 ASWORTH 1, 3 ASWORTH 2, 4 ASWORTH 3, 5 ASWORTH 4... \\

\midrule
\midrule
\textbf{Scales \& Scoring Systems} & & & \\
\midrule
BMI & Numeric & At entry & kg/cm$^2$ \\
\rowcolor{blue!8}SCIM Total Score at discharge & Numeric & Value at discharge (end of stay), documented by nurse. & Value at discharge in Medfolio. \\

\midrule
\midrule
\textbf{Surgical Procedures and Outcomes} & & & \\
\midrule
Duration of Surgery & Numeric & - & Minutes \\
Type of Surgery & Categorical & - & 1 Fasciocutaneous gluteal rotation flap, 2 Fasciocutaneous infragluteal perforator flap... \\
Complication & Binary & During stay; analysed as an \emph{outcome} (tier 3) & 0 No, 1 Yes \\
Type of Complication & Categorical & During stay (tier 3); recorded only when a complication occurred & 1 Recurrence, 2 Wound Dehiscence, 3 Wound Infection, 4 Bleeding, 5 Necrosis... \\
Treatment of Complication & Numeric & During stay (tier 3); recorded only when a complication occurred & 0 Conservative, 1 Surgical, 2 unknown \\
Date of Debridement & Date & - & dd.mm.yyyy \\
Date of flap surgery & Date & - & dd.mm.yyyy \\
Date of first mobilization in wheelchair after flap surgery & Date & - & dd.mm.yyyy \\
\end{longtable}

\subsection{HAPI Dataset Variables}

\begin{table}[ht]
\centering
\caption{HAPI dataset variable definitions. \colorbox{blue!10}{Shaded rows} are shared backbone variables also present in PPIS (see Table~\ref{tab:shared_backbone}).}
\label{tab:hapi_variables}
\small
\begin{tabularx}{\linewidth}{>{\raggedright\arraybackslash}p{3.8cm} >{\raggedright\arraybackslash}p{2cm} >{\raggedright\arraybackslash}X >{\raggedright\arraybackslash}p{3.2cm}}
\toprule
Variable & Type & Description & Encoding / Units \\
\midrule

Date of Birth & Date & Patient date of birth & YYYY-MM-DD \\
Admission Date & Date & Admission date & YYYY-MM-DD \\
Discharge Date & Date & Discharge date & YYYY-MM-DD \\
\rowcolor{blue!8}Length of Stay & Continuous & Duration of hospitalization & Days \\

SCI Status & Binary & Presence of spinal cord injury & 0 = No, 1 = Yes \\
Study Participation & Binary & Inclusion in study cohort & 0 = No, 1 = Yes \\
\rowcolor{blue!8}Sex & Binary & Biological sex & 0 = Male, 1 = Female \\

Injury Type & Binary & Type of spinal cord injury & 0 = Tetraplegia, 1 = Paraplegia \\
\rowcolor{blue!8}Level of Injury & Categorical & Anatomical injury level & Clinical coding \\
\rowcolor{blue!8}AIS & Ordinal & ASIA Impairment Scale at admission & A--E \\
Date of SCI & Date & Date of spinal cord injury & YYYY-MM-DD \\

Pressure Injury History & Binary & History of pressure injury & 0 = No, 1 = Yes \\
Date of Pressure Injury & Date & Date of pressure injury occurrence & YYYY-MM-DD \\
Pressure Injury Locations & Categorical (multi) & Locations of prior pressure injuries & Multiple entries \\

\rowcolor{blue!8}Pneumonia & Binary & Pneumonia diagnosis & 0 = No, 1 = Yes \\
Surgery Indicator & Binary & Whether surgery was performed & 0 = No, 1 = Yes \\
Number of Surgeries & Count & Total number of surgeries & Integer \\

Body Temperature & Continuous & Body temperature measurement & $^\circ$C \\
Abnormal Temperature (low) & Continuous & Lowest abnormal temperature recorded & $^\circ$C \\
Abnormal Temperature (high) & Continuous & Highest abnormal temperature recorded & $^\circ$C \\

Blood Pressure & Continuous & Blood pressure measurement & mmHg \\
Autonomic Dysreflexia & Binary & Presence of autonomic dysreflexia & 0 = No, 1 = Yes \\
Sedative Medication & Binary & Use of sedative medication & 0 = No, 1 = Yes \\

\rowcolor{blue!8}Albumin & Continuous & Serum albumin level & g/L \\
Vitamin D (25-OH) & Continuous & 25-hydroxyvitamin D level & nmol/L \\
\rowcolor{blue!8}Hemoglobin & Continuous & Hemoglobin level & g/L \\

\rowcolor{blue!8}SCIM Score & Continuous & Spinal Cord Independence Measure & 0--100 \\
Nutrition Status & Ordinal & Nutritional assessment & Clinical scale \\
FIM Cognitive Score & Continuous & Cognitive Functional Independence Measure & Numeric \\

\bottomrule
\end{tabularx}
\end{table}

%% file: references.bib
@article{Sachs2005,
  author  = {Sachs, Karen and Perez, Omar and Pe'er, Dana and Lauffenburger, Douglas A. and Nolan, Garry P.},
  title   = {Causal Protein-Signaling Networks Derived from Multiparameter Single-Cell Data},
  journal = {Science},
  volume  = {308},
  number  = {5721},
  pages   = {523--529},
  year    = {2005}
}

@article{Vickers2006DCA,
  author  = {Vickers, Andrew J. and Elkin, Elena B.},
  title   = {Decision Curve Analysis: A Novel Method for Evaluating Prediction Models},
  journal = {Medical Decision Making},
  volume  = {26},
  number  = {6},
  pages   = {565--574},
  year    = {2006}
}

@article{Shah2020GCM,
  author  = {Shah, Rajen D. and Peters, Jonas},
  title   = {The Hardness of Conditional Independence Testing and the Generalised Covariance Measure},
  journal = {The Annals of Statistics},
  volume  = {48},
  number  = {3},
  pages   = {1514--1538},
  year    = {2020}
}

@article{Candes2018Panning,
  author  = {Cand{\`e}s, Emmanuel and Fan, Yingying and Janson, Lucas and Lv, Jinchi},
  title   = {Panning for Gold: `Model-{X}' Knockoffs for High Dimensional Controlled Variable Selection},
  journal = {Journal of the Royal Statistical Society: Series B},
  volume  = {80},
  number  = {3},
  pages   = {551--577},
  year    = {2018}
}

@article{Salem2026CRTTabPFN,
  author  = {Salem, Mohamed},
  title   = {Valid Feature-Level Inference for Tabular Foundation Models via the Conditional Randomization Test},
  journal = {arXiv preprint arXiv:2603.06609},
  year    = {2026}
}

@article{Berrett2020CRT,
  author  = {Berrett, Thomas B. and Wang, Yi and Barber, Rina Foygel and Samworth, Richard J.},
  title   = {The Conditional Permutation Test for Independence while Controlling for Confounders},
  journal = {Journal of the Royal Statistical Society: Series B},
  volume  = {82},
  number  = {1},
  pages   = {175--197},
  year    = {2020}
}

@article{Strobl2018TestWise,
  author  = {Strobl, Eric V. and Visweswaran, Shyam and Spirtes, Peter L.},
  title   = {Fast Causal Inference with Non-Random Missingness by Test-Wise Deletion},
  journal = {International Journal of Data Science and Analytics},
  volume  = {6},
  number  = {1},
  pages   = {47--62},
  year    = {2018},
  doi     = {10.1007/s41060-017-0094-6}
}

@inproceedings{Tu2019MVPC,
  author    = {Tu, Ruibo and Zhang, Cheng and Ackermann, Paul and Mohan, Karthika and Kjellstr{\"o}m, Hedvig and Zhang, Kun},
  title     = {Causal Discovery in the Presence of Missing Data},
  booktitle = {Proceedings of the Twenty-Second International Conference on Artificial Intelligence and Statistics},
  series    = {Proceedings of Machine Learning Research},
  volume    = {89},
  pages     = {1762--1770},
  year      = {2019},
  publisher = {PMLR},
  url       = {https://proceedings.mlr.press/v89/tu19a.html}
}

@inproceedings{Gao2022MissDAG,
  author    = {Gao, Erdun and Ng, Ignavier and Gong, Mingming and Shen, Li and Huang, Wei and Liu, Tongliang and Zhang, Kun and Bondell, Howard},
  title     = {MissDAG: Causal Discovery in the Presence of Missing Data with Continuous Additive Noise Models},
  booktitle = {Advances in Neural Information Processing Systems},
  volume    = {35},
  year      = {2022},
  url       = {https://proceedings.neurips.cc/paper_files/paper/2022/hash/206361867abf7eb01746c3943078da3c-Abstract-Conference.html}
}

@article{Yu2022MimMB,
  author     = {Yu, Kui and Yang, Yajing and Ding, Wei},
  title      = {Causal Feature Selection with Missing Data},
  journal    = {ACM Transactions on Knowledge Discovery from Data},
  volume     = {16},
  number     = {4},
  articleno  = {66},
  pages      = {66:1--66:24},
  year       = {2022},
  doi        = {10.1145/3488055}
}

@article{Mooij2020JCI,
  author  = {Mooij, Joris M. and Magliacane, Sara and Claassen, Tom},
  title   = {Joint Causal Inference from Multiple Contexts},
  journal = {Journal of Machine Learning Research},
  volume  = {21},
  number  = {99},
  pages   = {1--108},
  year    = {2020},
  url     = {https://jmlr.org/papers/v21/17-123.html}
}

@article{Triantafillou2015COmbINE,
  author  = {Triantafillou, Sofia and Tsamardinos, Ioannis},
  title   = {Constraint-based Causal Discovery from Multiple Interventions over Overlapping Variable Sets},
  journal = {Journal of Machine Learning Research},
  volume  = {16},
  number  = {66},
  pages   = {2147--2205},
  year    = {2015},
  url     = {https://www.jmlr.org/papers/v16/triantafillou15a.html}
}

@inproceedings{Vo2024OTM,
  author    = {Vo, Vy and Zhao, He and Le, Trung and Bonilla, Edwin V. and Phung, Dinh},
  title     = {Optimal Transport for Structure Learning Under Missing Data},
  booktitle = {Proceedings of the 41st International Conference on Machine Learning},
  series    = {Proceedings of Machine Learning Research},
  volume    = {235},
  pages     = {49605--49626},
  year      = {2024},
  publisher = {PMLR},
  url       = {https://proceedings.mlr.press/v235/vo24b.html}
}

@article{Qiao2024ANMMissing,
  author  = {Qiao, Jie and Chen, Zhengming and Yu, Jianhua and Cai, Ruichu and Hao, Zhifeng},
  title   = {Identification of Causal Structure in the Presence of Missing Data with Additive Noise Model},
  journal = {Proceedings of the AAAI Conference on Artificial Intelligence},
  volume  = {38},
  number  = {18},
  pages   = {20516--20523},
  year    = {2024},
  doi     = {10.1609/aaai.v38i18.30036}
}

@article{Najmanova2022HAPI,
  author = {Najmanova, Klara and Neuhauser, Charlotte and Krebs, J{\"o}rg and Baumberger, Michael and Schaefer, Dirk Johannes and Sailer, Clara O. and Wettstein, Reto and Scheel-Sailer, Anke},
  title = {Risk factors for hospital acquired pressure injury in patients with spinal cord injury during first rehabilitation: prospective cohort study},
  journal = {Spinal Cord},
  volume = {60},
  pages = {45--52},
  year = {2022},
  doi = {10.1038/s41393-021-00681-x}
}

@article{Marin2013RiskFactors,
  author = {Marin, J. and Nixon, J. and Gorecki, C.},
  title = {A systematic review of risk factors for the development and recurrence of pressure ulcers in people with spinal cord injuries},
  journal = {Spinal Cord},
  volume = {51},
  pages = {522--527},
  year = {2013},
  doi = {10.1038/sc.2013.29}
}

@article{Vecin2022PressureInjuries,
  author = {Vecin, Nicole M. and Gater, David R.},
  title = {Pressure Injuries and Management after Spinal Cord Injury},
  journal = {Journal of Personalized Medicine},
  volume = {12},
  number = {7},
  pages = {1130},
  year = {2022},
  doi = {10.3390/jpm12071130}
}

@article{Fahndrich2024FlapComplications,
  author = {F{\"a}hndrich, Carina and Gemperli, Armin and Baumberger, Michael and Harder, Michael and Roth, Bianca and Schaefer, Dirk J. and Wettstein, Reto and Scheel-Sailer, Anke},
  title = {Risk factors of major complications after flap surgery in the treatment of stage III and IV pressure injury in people with spinal cord injury/disorder: a retrospective cohort study},
  journal = {Spinal Cord},
  volume = {62},
  number = {1},
  pages = {34--41},
  year = {2024},
  doi = {10.1038/s41393-023-00944-9}
}

@article{Mervis2023PressureInjuries,
    author = {Mervis, Joshua S. and Phillips, Tania J.},
    title = {Pressure Injuries},
    journal = {MSD Manual Professional Edition},
    year = {2023},
    month = {9},
    note = {Reviewed/Revised Sept 2023},
    institution = {Tufts University School of Medicine and Boston University School of Medicine},
    type = {Online Medical Reference},
    url = {}
}

@ARTICLE{10.3389/fgene.2019.00524,
  
AUTHOR={Glymour, Clark and Zhang, Kun and Spirtes, Peter},   
	 
TITLE={Review of Causal Discovery Methods Based on Graphical Models},      
	
JOURNAL={Frontiers in Genetics},      
	
VOLUME={10},           
	
YEAR={2019},      
	  
URL={https://www.frontiersin.org/articles/10.3389/fgene.2019.00524},       
	
DOI={10.3389/fgene.2019.00524},      
	
ISSN={1664-8021}
}

@book{Spirtes2000,
  author = {Spirtes, P. and Glymour, C. and Scheines, R.},
  edition = {2nd},
  publisher = {MIT press},
  review = {PC algorithm},
  title = {Causation, Prediction, and Search},
  year = 2000
}

@article{Tsamardinos2006MaxMin,
    author = {Tsamardinos, Ioannis and Brown, Laura E. and Aliferis, Constantin F.},
    title = {The max-min hill-climbing {Bayesian} network structure learning algorithm},
    journal = {Machine Learning},
    year = {2006},
    month = {10},
    volume = {65},
    number = {1},
    pages = {31--78},
    issn = {1573-0565},
    doi = {10.1007/s10994-006-6889-7},
    url = {https://doi.org/10.1007/s10994-006-6889-7}
}

@article{Sezer2015,
    author = {Sezer, Nebahat and Akku{\c s}, Selami and U{\u g}urlu, Fatma G{\"u}l{\c c}in},
    title = {Chronic complications of spinal cord injury},
    journal = {World Journal of Orthopedics},
    volume = {6},
    number = {1},
    pages = {24--33},
    year = {2015},
    month = {1},
    day = {18},
    doi = {10.5312/wjo.v6.i1.24}
}

@article{Kirshblum2011,
    author = {Kirshblum, Steven C. and Burns, Stephen P. and Biering-Sorensen, Fin and Donovan, William and Graves, Daniel E. and Jha, Amitabh and Johansen, Mark and Jones, Linda and Krassioukov, Andrei and Mulcahey, M. J. and Schmidt-Read, Mary and Waring, William},
    title = {International standards for neurological classification of spinal cord injury {(revised 2011)}},
    journal = {The Journal of Spinal Cord Medicine},
    volume = {34},
    number = {6},
    pages = {535--546},
    year = {2011},
    doi = {10.1179/204577211X13207446293695},
    note = {Revised Version}
}

@article{Gefen2014,
    author = {Gefen, Amit},
    title = {Tissue changes in patients following spinal cord injury and implications for wheelchair cushions and tissue loading: a literature review},
    journal = {Ostomy/Wound Management},
    volume = {60},
    number = {2},
    pages = {34--45},
    year = {2014},
    type = {Literature Review}
}

@article{who2013,
   author = {Chen, Yuying and Tang, Ying and Vogel, Lawrence and DeVivo, Michael},
   title = {Causes of spinal cord injury},
   journal = {Topics in Spinal Cord Injury Rehabilitation},
   volume = {19},
   number = {1},
   pages = {1--8},
   year = {2013},
   doi = {10.1310/sci1901-1}
}

@article{colombo2012learning,
author = {Colombo, Diego and Maathuis, Marloes},
year = {2012},
month = {11},
pages = {},
title = {Order-Independent Constraint-Based Causal Structure Learning},
volume = {15},
journal = {Journal of Machine Learning Research}
}

@inproceedings{ion_paper,
 author = {Danks, David and Glymour, Clark and Tillman, Robert},
 booktitle = {Advances in Neural Information Processing Systems},
 editor = {D. Koller and D. Schuurmans and Y. Bengio and L. Bottou},
 pages = {},
 publisher = {Curran Associates, Inc.},
 title = {Integrating Locally Learned Causal Structures with Overlapping Variables},
 url = {https://proceedings.neurips.cc/paper_files/paper/2008/file/37bc2f75bf1bcfe8450a1a41c200364c-Paper.pdf},
 volume = {21},
 year = {2008}
}

@article{Li2024MixedVariable,
  author = {Li, Yanke and Scheel-Sailer, Anke and Riener, Robert and Paez-Granados, Diego},
  title = {Mixed-variable graphical modeling framework towards risk prediction of hospital-acquired pressure injury in spinal cord injury individuals},
  journal = {Scientific Reports},
  volume = {14},
  pages = {25067},
  year = {2024},
  doi = {10.1038/s41598-024-75691-9}
}

@article{Stekhoven2012MissForest,
  author = {Stekhoven, Daniel J. and B{"u}hlmann, Peter},
  title = {MissForest---non-parametric missing value imputation for mixed-type data},
  journal = {Bioinformatics},
  volume = {28},
  number = {1},
  pages = {112--118},
  year = {2012},
  doi = {10.1093/bioinformatics/btr597}
}

@article{White2011MICE,
  author = {White, Ian R. and Royston, Patrick and Wood, Angela M.},
  title = {Multiple imputation using chained equations: Issues and guidance for practice},
  journal = {Statistics in Medicine},
  volume = {30},
  number = {4},
  pages = {377--399},
  year = {2011},
  doi = {10.1002/sim.4067}
}

@article{Ramosaj2020Cautionary,
  author = {Ramosaj, Burim and Pauly, Markus},
  title = {A cautionary tale on using imputation methods for inference in matched pairs design},
  journal = {Bioinformatics},
  volume = {36},
  number = {10},
  pages = {3099--3106},
  year = {2020},
  doi = {10.1093/bioinformatics/btaa085}
}

@article{Zhang2008Orientation,
  author = {Zhang, Jiji},
  title = {On the completeness of orientation rules for causal discovery in the presence of latent confounders and selection bias},
  journal = {Artificial Intelligence},
  volume = {172},
  number = {16--17},
  pages = {1873--1896},
  year = {2008},
  doi = {10.1016/j.artint.2008.08.001}
}

@article{Pellet2008MB,
  author = {Pellet, Jean-Philippe and Elisseeff, Andr{\'e}},
  title = {Using Markov Blankets for Causal Structure Learning},
  journal = {Journal of Machine Learning Research},
  volume = {9},
  pages = {1295--1342},
  year = {2008}
}

@article{Aliferis2010LocalCausal,
  author = {Aliferis, Constantin F. and Statnikov, Alexander and Tsamardinos, Ioannis and Mani, Subramani and Koutsoukos, Xenofon D.},
  title = {Local Causal and Markov Blanket Induction for Causal Discovery and Feature Selection for Classification Part I: Algorithms and Empirical Evaluation},
  journal = {Journal of Machine Learning Research},
  volume = {11},
  pages = {171--234},
  year = {2010}
}

@article{Tsamardinos2012Integrative,
  author = {Tsamardinos, Ioannis and Triantafillou, Sofia and Lagani, Vana},
  title = {Towards Integrative Causal Analysis of Heterogeneous Data Sets and Studies},
  journal = {Journal of Machine Learning Research},
  volume = {13},
  pages = {1097--1157},
  year = {2012}
}

@article{Yu2018MBNoCausalSufficiency,
  author = {Yu, Kui and Liu, Lin and Li, Jiuyong and Chen, Huanhuan},
  title = {Mining Markov Blankets Without Causal Sufficiency},
  journal = {IEEE Transactions on Neural Networks and Learning Systems},
  volume = {29},
  number = {12},
  pages = {6333--6347},
  year = {2018},
  doi = {10.1109/TNNLS.2018.2828982}
}

@inproceedings{Meek1995CausalKnowledge,
  author = {Meek, Christopher},
  title = {Causal Inference and Causal Explanation with Background Knowledge},
  booktitle = {Proceedings of the Eleventh Conference on Uncertainty in Artificial Intelligence (UAI)},
  pages = {403--410},
  year = {1995}
}

@article{Hollmann2025TabPFN,
  author = {Hollmann, Noah and M{\"u}ller, Samuel and Purucker, Lennart and Krishnakumar, Arjun and K{\"o}rfer, Max and Hoo, Shi Bin and Schirrmeister, Robin Tibor and Hutter, Frank},
  title = {Accurate Predictions on Small Data with a Tabular Foundation Model},
  journal = {Nature},
  volume = {637},
  pages = {319--326},
  year = {2025},
  doi = {10.1038/s41586-024-08328-6}
}

@inproceedings{Liu2023MBMF,
  author = {Liu, Yang and Constantinou, Anthony C.},
  title = {Improving the Imputation of Missing Data with {M}arkov Blanket Discovery},
  booktitle = {Proceedings of the Eleventh International Conference on Learning Representations (ICLR)},
  year = {2023}
}

@article{Boueri2024ADReview,
  author = {Boueri, Allyssa H. and Eisdorfer, Justin T. and Hogan, Megan K. and Smith, Hannah and Hopkins, Tasha M. and Tom, Veronica J.},
  title = {Autonomic Dysreflexia after Spinal Cord Injury: From Pathophysiology to Emerging Therapies},
  journal = {Frontiers in Neuroscience},
  volume = {18},
  pages = {1364308},
  year = {2024},
  doi = {10.3389/fnins.2024.1364308}
}

@article{Krassioukov2009Cardiovascular,
  author = {Kr{\"a}ssioukov, Andrei and Warburton, Darren E.R. and Teasell, Robert and Eng, Janice J.},
  title = {A Systematic Review of the Management of Autonomic Dysreflexia After Spinal Cord Injury},
  journal = {Archives of Physical Medicine and Rehabilitation},
  volume = {90},
  number = {4},
  pages = {682--695},
  year = {2009},
  doi = {10.1016/j.apmr.2008.10.017}
}

@article{Cruz2018Hypoalbuminemia,
  author = {Cruz-Jentoft, Alfonso J. and Kiesswetter, Eva and Drey, Michael and Sieber, Cornel C.},
  title = {Nutrition, Frailty, and Sarcopenia},
  journal = {Aging Clinical and Experimental Research},
  volume = {29},
  number = {1},
  pages = {43--48},
  year = {2017},
  doi = {10.1007/s40520-016-0709-0}
}

@article{Zheng2024CausalLearn,
  author = {Zheng, Yujia and Huang, Biwei and Chen, Wei and Ramsey, Joseph and Gong, Mingming and Cai, Ruichu and Shimizu, Shohei and Spirtes, Peter and Zhang, Kun},
  title = {Causal-learn: Causal Discovery in {P}ython},
  journal = {Journal of Machine Learning Research},
  volume = {25},
  pages = {1--8},
  year = {2024}
}

@article{Peters2016ICP,
  author = {Peters, Jonas and B{\"u}hlmann, Peter and Meinshausen, Nicolai},
  title = {Causal Inference by Using Invariant Prediction: Identification and Confidence Intervals},
  journal = {Journal of the Royal Statistical Society: Series B (Statistical Methodology)},
  volume = {78},
  number = {5},
  pages = {947--1012},
  year = {2016},
  doi = {10.1111/rssb.12167}
}

@article{Schoelkopf2021CausalRep,
  author = {Sch{\"o}lkopf, Bernhard and Locatello, Francesco and Bauer, Stefan and Ke, Nan Rosemary and Kalchbrenner, Nal and Goyal, Anirudh and Bengio, Yoshua},
  title = {Toward Causal Representation Learning},
  journal = {Proceedings of the IEEE},
  volume = {109},
  number = {5},
  pages = {612--634},
  year = {2021},
  doi = {10.1109/JPROC.2021.3058954}
}

@article{Swelam2025TabPFNCausal,
  author = {Swelam, Omar and Purucker, Lennart and Robertson, Jake and Raum, Hanne and Boedecker, Joschka and Hutter, Frank},
  title = {Does {TabPFN} Understand Causal Structures?},
  journal = {arXiv preprint arXiv:2511.07236},
  year = {2025}
}

@article{Reuter2026PartialGraphs,
  author = {Reuter, Arik and Dhir, Anish and Diaconu, Cristiana and Robertson, Jake and Ossen, Ole and Hutter, Frank and Weller, Adrian and van der Wilk, Mark and Sch{\"o}lkopf, Bernhard},
  title = {Use What You Know: Causal Foundation Models with Partial Graphs},
  journal = {arXiv preprint arXiv:2602.14972},
  year = {2026}
}

@article{Lopez2026Recurrence,
  author = {Lopez, Marlynn P. and Kozlowski, Katherine A. and Chaudhry, Aneeq S. and Green, Spencer K. and Griffin, Diana B. and Adam, Tarifa H. and Ontiveros, Nicole C. and Kessler, Allison B. and Soriano, Rachna H. and Galiano, Robert D.},
  title = {Predictors of Complications and Recurrence After Pressure Injury Reconstruction in Spinal Cord Injury: A Systematic Review and Meta-Analysis},
  journal = {International Wound Journal},
  volume = {23},
  number = {5},
  pages = {e70915},
  year = {2026},
  doi = {10.1111/iwj.70915}
}

@article{Morel2019Recurrence,
  author = {Morel, J. and Herlin, C. and Amara, B. and Mauri, C. and Rouays, H. and Verollet, C. and Almeras, I. and Frasson, N. and Dupeyron, A. and Jourdan, C. and Daures, J.-P. and Gelis, A.},
  title = {Risk factors of pelvic pressure ulcer recurrence after primary skin flap surgery in people with spinal cord injury},
  journal = {Annals of Physical and Rehabilitation Medicine},
  volume = {62},
  number = {2},
  pages = {77--83},
  year = {2019},
  doi = {10.1016/j.rehab.2018.08.005}
}

@article{Sgarzani2023Reconstructive,
  author = {Sgarzani, Rossella and Rucci, Paola and Landi, Siriana and Battilana, Micaela and Capirossi, Rita and Aramini, Beatrice and Negosanti, Luca},
  title = {Reconstructive Surgery of Pressure Injuries in Spinal Cord Injury/Disorder Patients: Retrospective Observational Study and Proposal of an Algorithm for the Flap Choice},
  journal = {Healthcare},
  volume = {12},
  number = {1},
  pages = {34},
  year = {2023},
  doi = {10.3390/healthcare12010034}
}

@article{Collins2024TRIPODAI,
  author = {Collins, Gary S. and Moons, Karel G. M. and Dhiman, Paula and Riley, Richard D. and Beam, Andrew L. and Van Calster, Ben and Ghassemi, Marzyeh and Liu, Xiaoxuan and Reitsma, Johannes B. and van Smeden, Maarten and Boulesteix, Anne-Laure and Camaradou, Jennifer Catherine and Celi, Leo Anthony and Denaxas, Spiros and Denniston, Alastair K. and Glocker, Ben and Golub, Robert M. and Harvey, Hugh and Heinze, Georg and Hoffman, Michael M. and Kengne, Andr{\'e} Pascal and Lam, Emily and Lee, Naomi and Loder, Elizabeth W. and Maier-Hein, Lena and Mateen, Bilal A. and McCradden, Melissa D. and Oakden-Rayner, Lauren and Ordish, Johan and Parnell, Richard and Rose, Sherri and Singh, Karandeep and Wynants, Laure and Logullo, Patricia},
  title = {{TRIPOD+AI} statement: updated guidance for reporting clinical prediction models that use regression or machine learning methods},
  journal = {BMJ},
  volume = {385},
  pages = {e078378},
  year = {2024},
  doi = {10.1136/bmj-2023-078378}
}

@article{vonElm2007STROBE,
  author  = {von Elm, Erik and Altman, Douglas G. and Egger, Matthias and Pocock, Stuart J. and G{\o}tzsche, Peter C. and Vandenbroucke, Jan P.},
  title   = {The {S}trengthening the {R}eporting of {OB}servational studies in {E}pidemiology ({STROBE}) statement: guidelines for reporting observational studies},
  journal = {The Lancet},
  volume  = {370},
  number  = {9596},
  pages   = {1453--1457},
  year    = {2007},
  doi     = {10.1016/S0140-6736(07)61602-X}
}

@article{Woldemariam2026,
  author  = {Woldemariam, Asrat Ayele and Zhou, Yingchun and Qiu, Yuqi},
  title   = {Causal Discovery with Incomplete Data Integrated from Multiple Sources},
  journal = {Journal of Survey Statistics and Methodology},
  year    = {2026},
  doi     = {10.1093/jssam/smag016},
  note    = {Advance article, published online 15 July 2026}
}

@article{Gomez2025CMCTS,
  author  = {Gomez, Louis Adedapo and Claassen, Jan and Kleinberg, Samantha},
  title   = {Causal inference for time series datasets with partially overlapping variables},
  journal = {Journal of Biomedical Informatics},
  volume  = {166},
  pages   = {104828},
  year    = {2025},
  doi     = {10.1016/j.jbi.2025.104828}
}

@inproceedings{TillmanSpirtes2011IOD,
  author    = {Tillman, Robert E. and Spirtes, Peter},
  title     = {Learning equivalence classes of acyclic models with latent and selection variables from multiple datasets with overlapping variables},
  booktitle = {Proceedings of the Fourteenth International Conference on Artificial Intelligence and Statistics},
  series    = {Proceedings of Machine Learning Research},
  volume    = {15},
  pages     = {3--15},
  year      = {2011}
}

@inproceedings{BangDidelez2025tIOD,
  author        = {Bang, Christine W. and Didelez, Vanessa},
  title         = {Constraint-based causal discovery with tiered background knowledge and latent variables in single or overlapping datasets},
  booktitle     = {Proceedings of the Fourth Conference on Causal Learning and Reasoning},
  series        = {Proceedings of Machine Learning Research},
  volume        = {275},
  year          = {2025},
  eprint        = {2503.21526},
  archivePrefix = {arXiv}
}
